\documentclass[conference]{IEEEtran}
\IEEEoverridecommandlockouts
\usepackage{cite}
\usepackage{amsmath,amssymb,amsfonts}
\usepackage{tabularx}
\usepackage{subfig}
\usepackage{url}
\usepackage{textcomp}
\usepackage{xcolor}
\usepackage{nicefrac}       
\usepackage[pdftex]{graphicx}
\usepackage{multirow}
\usepackage{xspace}
\usepackage{enumitem}
\usepackage{booktabs}
\usepackage{tabu}
\usepackage{siunitx}
\usepackage{algorithm}
\usepackage{algpseudocode}

\newcolumntype{L}[1]{>{\raggedright\let\newline\\\arraybackslash\hspace{0pt}}m{#1}}
\newcolumntype{C}[1]{>{\centering\let\newline\\\arraybackslash\hspace{0pt}}m{#1}}
\newcolumntype{R}[1]{>{\raggedleft\let\newline\\\arraybackslash\hspace{0pt}}m{#1}}

\def\BibTeX{{\rm B\kern-.05em{\sc i\kern-.025em b}\kern-.08em
    T\kern-.1667em\lower.7ex\hbox{E}\kern-.125emX}}

\newcommand{\zilongout}[1]{}

\newcommand{\system}{RAD\xspace}
\newcommand{\systemVoting}{RAD Voting\xspace}
\newcommand{\systemActive}{RAD Active Learning\xspace}
\newcommand{\systemActiveLimit}{RAD Active Learning Limited\xspace}
\newcommand{\systemSlimmed}{RAD Slim\xspace}
\newcommand{\systemSlimmedLimit}{RAD Slim Limited\xspace}
\newcommand{\systemIDS}{IDS\xspace}
\newcommand{\systemCoteaching}{Co-Teaching\xspace}
\newcommand{\systemPreselect}{Pre-Select Oracle\xspace}
\newcommand{\highestActiveLimit}{Highest Uncertainty Method\xspace}

\author{

   \IEEEauthorblockN{Zilong Zhao\IEEEauthorrefmark{1}
   ,  Robert Birke\IEEEauthorrefmark{2}, Rui Han\IEEEauthorrefmark{3}, Bogdan Robu\IEEEauthorrefmark{1}, Sara Bouchenak\IEEEauthorrefmark{4}, Sonia Ben Mokhtar\IEEEauthorrefmark{4}, Lydia Y. Chen
    \IEEEauthorrefmark{5} }
    \IEEEauthorblockA{\IEEEauthorrefmark{1}Univ. Grenoble Alpes, France
    \{zilong.zhao, bogdan.robu\}@gipsa-lab.fr}    
    \IEEEauthorblockA{\IEEEauthorrefmark{2}ABB Research, Switzerland
    \{robert.birke\}@ch.abb.com}
     \IEEEauthorblockA{\IEEEauthorrefmark{3} Corresponding author: Beijing Institute of Technology, China 
    \{hanrui\}@bit.edu.cn}
    \IEEEauthorblockA{\IEEEauthorrefmark{4}INSA Lyon, France
    \{sara.bouchenak, sonia.benmokhtar\}@insa-lyon.fr}
    \IEEEauthorblockA{\IEEEauthorrefmark{5}TU Delft, Netherlands
    \{y.chen-10\}@tudelft.nl}
}

\title{Enhancing Robustness of On-line Learning Models on Highly Noisy Data}

\begin{document}

\maketitle

\begin{abstract}

Classification algorithms have been widely adopted to detect anomalies for various systems, e.g.,~IoT, cloud and face recognition, under the common assumption that the data source is clean, i.e.,~features and labels are correctly set.
However, data collected from the wild can be unreliable due to careless annotations or malicious data transformation for incorrect anomaly detection.
In this paper, 
we extend a two-layer on-line data selection framework: Robust Anomaly Detector (\system) with a newly designed ensemble prediction where both layers contribute to the final anomaly detection decision. To adapt to the on-line nature of anomaly detection, we consider additional features of conflicting opinions of classifiers, repetitive cleaning, and oracle knowledge. We on-line learn from incoming data streams and continuously cleanse the data, so as to adapt to the increasing learning capacity from the larger accumulated data set. Moreover, we explore the concept of oracle learning that provides additional information of true labels for difficult data points. 
We specifically focus on three use cases, (i) detecting 10 classes of IoT attacks, (ii) predicting 4 classes of task failures of big data jobs, and (iii) recognising 100 celebrities faces.
Our evaluation results show that \system can robustly improve the accuracy of anomaly detection, to reach up to 98.95\% for IoT device attacks (i.e.,~+7\%), up to 85.03\% for cloud task failures (i.e.,~+14\%) under 40\% label noise, and for its extension, it can reach up to 77.51\% for face recognition (i.e.,~+39\%) under 30\% label noise. 
The proposed \system and its extensions are general and can be applied to different anomaly detection algorithms. 
 \end{abstract}

\begin{IEEEkeywords}
Unreliable Data; Anomaly Detection; Failures; Attacks; Machine Learning
\end{IEEEkeywords}


\section{Introduction}
\label{sec:Introduction}

Anomaly detection is one of the core operations for enforcing dependability and performance in modern distributed systems~\cite{Xue:TNSM18:Ticket,DBLP:journals/tpds/PhamWTBTKI17}. Anomalies can take various forms including erroneous data produced by a corrupted IoT device or the failure of a job executed in a datacenter~\cite{Birke:TNSM16:Cloud,BirkeDSN2014,Zhao:DSN19}.

Dealing with this issue has often been done in recent art by relying on machine learning-based classification algorithms over system logs~\cite{Fang:2010,Giantamidis:2016} or backend collected data~\cite{DBLP:conf/ijcai/ZhangLSLH19,Huang:2017}. These systems often rely on the assumption of clean datasets from which the classifier learns to distinguish between data corresponding to a correct execution of the system from data corresponding to an abnormal execution of the latter (i.e., anomaly detection). As workloads at real systems are highly dynamic over time, it is even more challenging to predict anomalies that can not be easily distinguished from the system dynamics, compared to the systems with static workloads.

In this context, a rising concern when applying classification algorithms is the accessibility to a reliable ground truth for anomalies~\cite{Cerf:NIPsWS18:Duao}. Typically, anomaly data is manually annotated by human experts and hence the generation of anomaly labels is subject to quality variation, so-called noisy labels. For instance, annotating service failure types for data centers is done by operators of varying expertise.

However, standard machine learning algorithms typically assume clean labels and overlook the risk of noisy labels.  Moreover, recent studies point out the increase in dirty data attacks that can maliciously alter the anomaly labels to mislead the machine learning models~\cite{Kang:2018,Fan:2018,He:2017,tps2019}. As a result, anomaly detection algorithms need to capture not only anomalies that are entangled with system dynamics but also the unreliable nature of anomaly labels.

Indeed, a strong anomaly classification model can be learned by incorporating a larger amount of data. However, learning from data with noisy labels can significantly degrade the classification accuracy, even for deep neural networks~\cite{Vagin:2011, DBLP:conf/iclr/ZhangBHRV17, tps2019}. 
Such concerns lead us to ask the following question:
how to build an anomaly detection framework that can robustly differentiate between true and noisy anomalies and efficiently learn anomaly classification models from a succinct amount of clean data. The immediate challenge of capturing the data quality lies at the fact that label qualities are not directly observable but only via anomaly classification outcomes that in turn are coupled with the noise level in data labels.

We extend Robust Anomaly Detector (\system)~\cite{Zhao:DSN19}, a generic framework that continuously learns an anomaly classification model from streams of event logs or images that are subject to label noise.  The original design of \system is composed of two layers of learning models, i.e., a data label model and an anomaly classifier. The label model aims at differentiating the label quality, i.e., noisy v.s. true labels, for each batch of new data and only ``clean'' data points are fed in the anomaly classifier. The anomaly classifier predicts the event outcome that can be divided in multiple classes of (non)anomalies, depending on the specific use case. In this extension, we derive three alternatives of \system, namely, voting, active learning and slim. These use additional information, e.g., opinions of conflicting classifiers and queries of oracles. We iteratively update the prediction of historical windows such that weak predictions can be continuously improved by the latest model. Moreover, we propose an ensemble prediction strategy to reconcile the prediction outcomes of the two models, namely label model and anomaly classification model, instead of only relying on classification model as~\cite{Zhao:DSN19}

To demonstrate the effectiveness of \system, we consider three use cases using open datasets: detecting 10 classes of attacks on IoT devices~\cite{meidan2018n}, predicting 4 types of task failures for big data processing cluster~\cite{reiss2011google,Rosa-DSN15}, and recognising 100 most abundant celebrity faces~\cite{facescrub:2014}.
Our results show that \system can effectively cleanse the data, i.e., selecting data with clean labels, and result in better anomaly detection accuracy per additional included streamed data, compared to classifiers without continuous data cleansing. Specifically, under 40\% noise, \system achieves up to 98.95\%, 85.03\% (comparing to 92.27\% and 71.02\% by anomaly classification model of no selection on dataset) for detecting IoT device attacks and predicting cluster task failures, respectively. If we implement \systemActive on cluster dataset with the same noise level, the final accuracy could improve from 85.03\% to 90.77\%. For face image dataset, final accuracy of \systemSlimmed under 30\% noise achieves to 77.51\% (comparing to 38.89\% of no selection on dataset). Furthermore, our study shows that \system is stable even when the noise is very strong. And if we do not have many clean data at beginning to pre-train the model, \systemActive and \systemActiveLimit can still perform very well from a very bad starting model. 

The main contributions of this study can be summarized as follows:
\begin{itemize}
 \item  We design an effective on-line anomaly detection framework, \system, consisting of a data selection and prediction module that cater to a wide range of implementation choices from regular machine learning models to deep neural networks. 
 \item We explore three novel data selection schemes: namely voting, active learning, and active learning limit. These can filter out the suspicious data and call upon experts to cleanse the data based on the predicted uncertainty from the quality model and classification model. We combine the novel ideas of model disagreement and active learning. 
\item We leverage the power of ensemble model prediction to enhance the robustness of trained anomaly classifier by incorporating the predictions of the label model used in the data selection.
\item \system can be applied on multiple types of anomaly inputs, i.e., server failure, IoT devices, and images. Specially, \systemActive can achieve remarkable accuracy similar to the performance under no label noise.
\end{itemize}

The remainder of the paper is organized as follows.
Section~\ref{sec:ProblemStatement} describes the motivating case studies.
Sections~\ref{sec:Framework} and \ref{sec:Evaluation} present the proposed \system framework and the results of its experimental evaluation, respectively.
Section~\ref{sec:RelatedWork} describes the related work. Finally, Section~\ref{sec:Conclusion} draws our conclusions.

\section{Related Work}
\label{sec:RelatedWork}



Machine learning has been extensively used for failure detection~\cite{Pitakrat:2013,Pellegrini:2015,Reuter:2018,Campos:2018},
attack prediction~\cite{Agarwal:2016,Banescu:2017,Anbar:2018,Kozik:2018,Karagiannis:2018,Zhou:2018}, and face recognition~\cite{DBLP:conf/cvpr/TaigmanYRW14,DBLP:conf/cvpr/SchroffKP15,DBLP:conf/cvpr/WangWZJGZL018}.
Considering noisy labels in classification algorithms is also a problem that has been explored in the machine learning community as discussed in~\cite{Frenay:TNNLS14:survey,biggio2011support,natarajan2013learning}.

The problem of classification in presence of noisy labels can be organized into various categories according to, on the one hand, the type of classification algorithm subject to noise, and on the other hand, the techniques used to remove the noise.

Noisy labels have been studied for binary classification where noisy labels are considered as symmetric (e.g., \cite{larsen1998design}) and for classification with multiple classes where noisy labels are considered as asymmetric, e.g.,~\cite{patrini2017making,sukhbaatar2014training}. For this paper, we consider the problem of classification with multiple classes.
Furthermore, noisy labels have been considered in various types of classifiers KNN~\cite{Wilson:ML2000:MLselection}, SVM~\cite{An:Neurocomput.13:SVM}, and deep neural networks~\cite{Vahdat:NIPS17:NoisyLabelDNN}. In the context of this paper, our proposed approach is agnostic to the underlying classifier type as noise removal is performed ahead of the classification.

Various techniques explore countering label noise following two main strategies. The first strategy trains a single model as filter for noisy label data. \cite{Agarwal:2016,veit2017learning,li2017learning,Dan:nips2018:Using} train a separate filter from clean data for distinguishing noisy labels. Instead \cite{DBLP:conf/apweb/ThongkamXZH08} trains on the original data (with noise). Training the filter with clean data is better, but the assumption of large quantity of clean data does not always hold, especially in our on-line learning scenario. Using noisy data to train a filter raises a \textit{chicken-and-egg} dilemma~\cite{Frenay:TNNLS14:survey}, since: 1) good classifiers are necessary for filtering but 2) learning from noisy label data may precisely produce poor classifiers. 

\begin{figure*}[h]
	\begin{center}
		\subfloat[Use case of IoT thermostat device attacks]{
			\includegraphics[width=0.60\columnwidth]{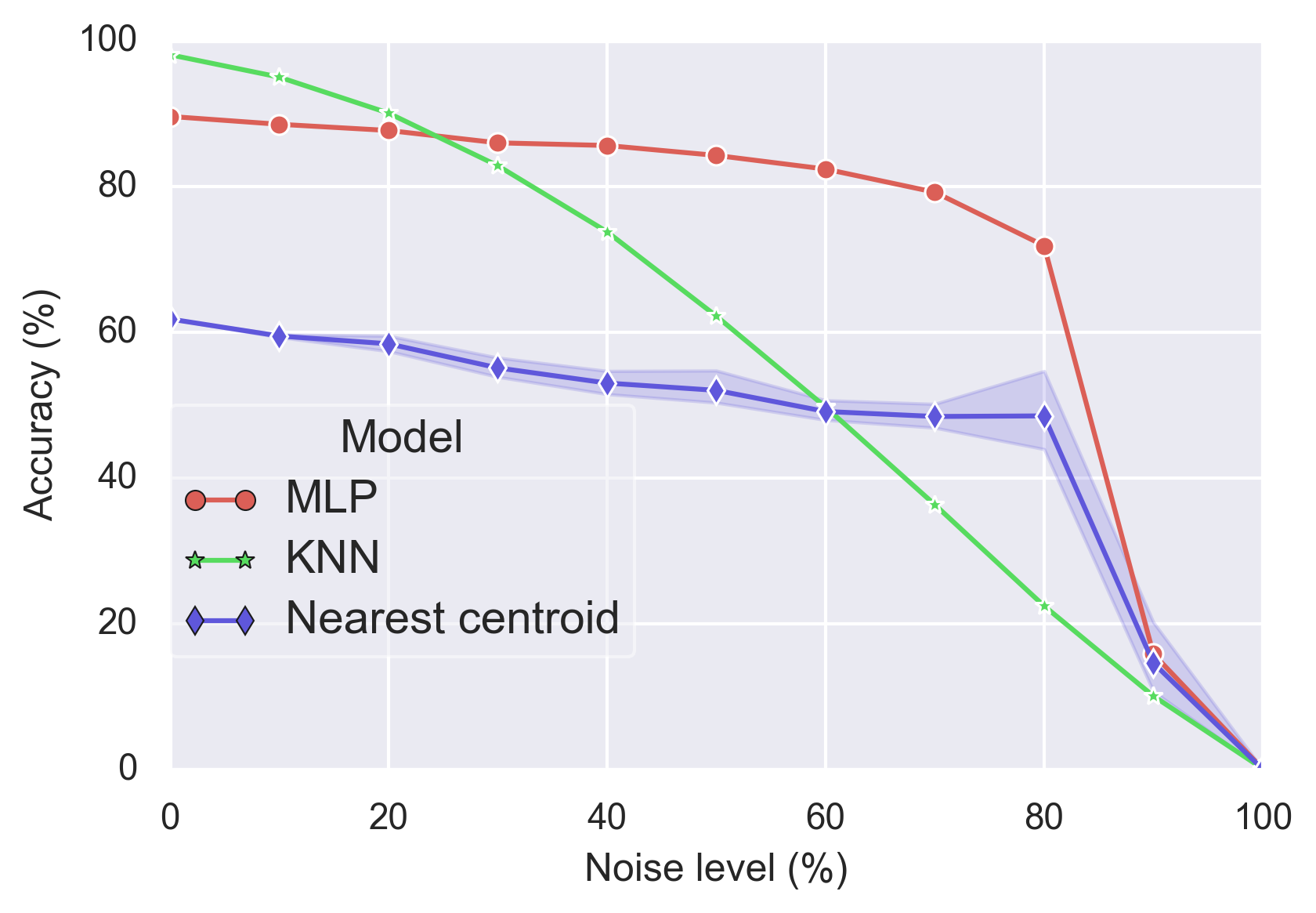}
			\label{fig:NoisyData-Classif-IoT}
		}		
		\hspace{\fill}
		\subfloat[Use case of Cluster task failures]{
			\includegraphics[width=0.60\columnwidth]{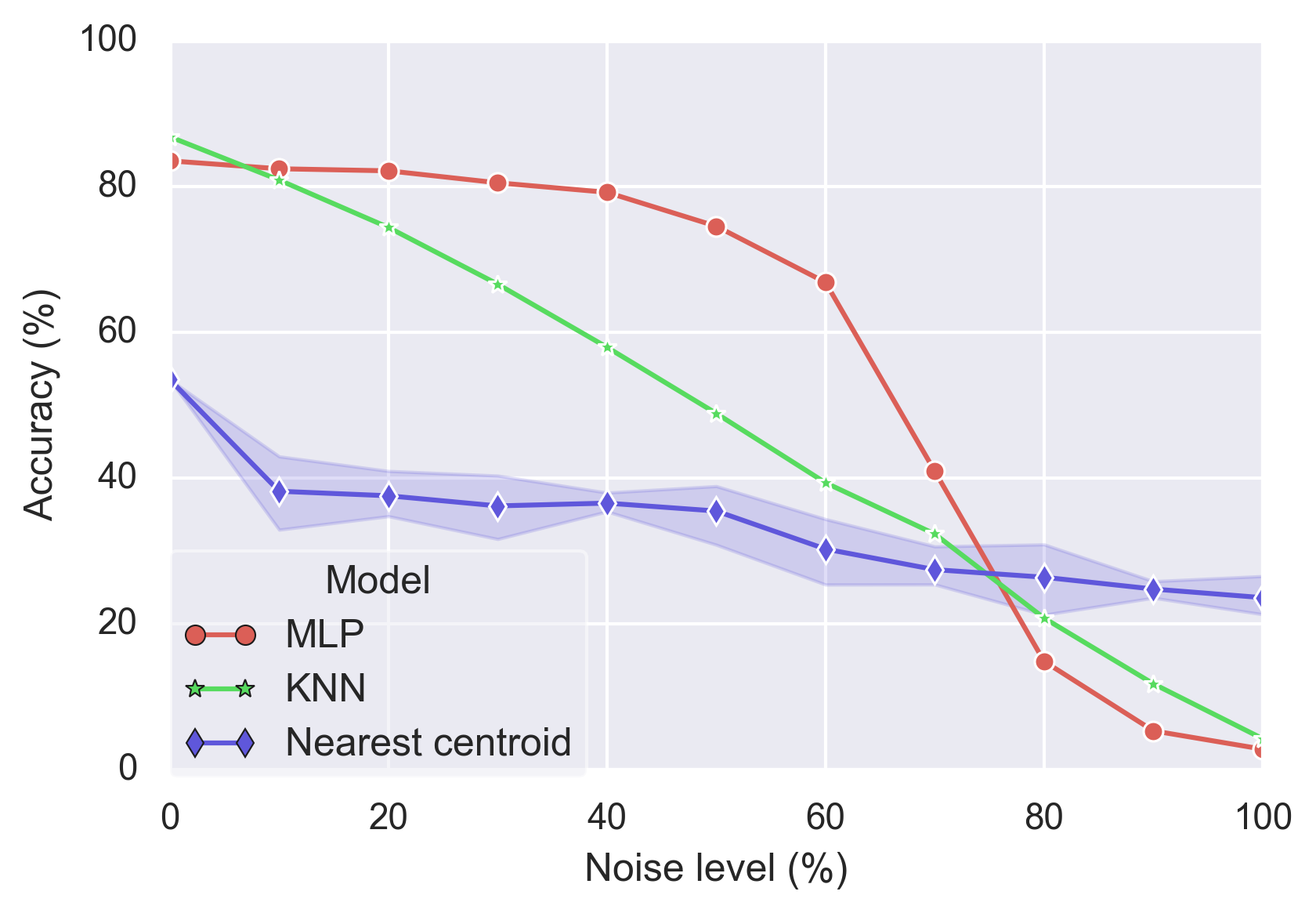}
			\label{fig:NoisyData-Classif-ClusterTasks}
		}
		\hspace{\fill}
		\subfloat[Use case of Face Recognition]{
			\includegraphics[width=0.60\columnwidth]{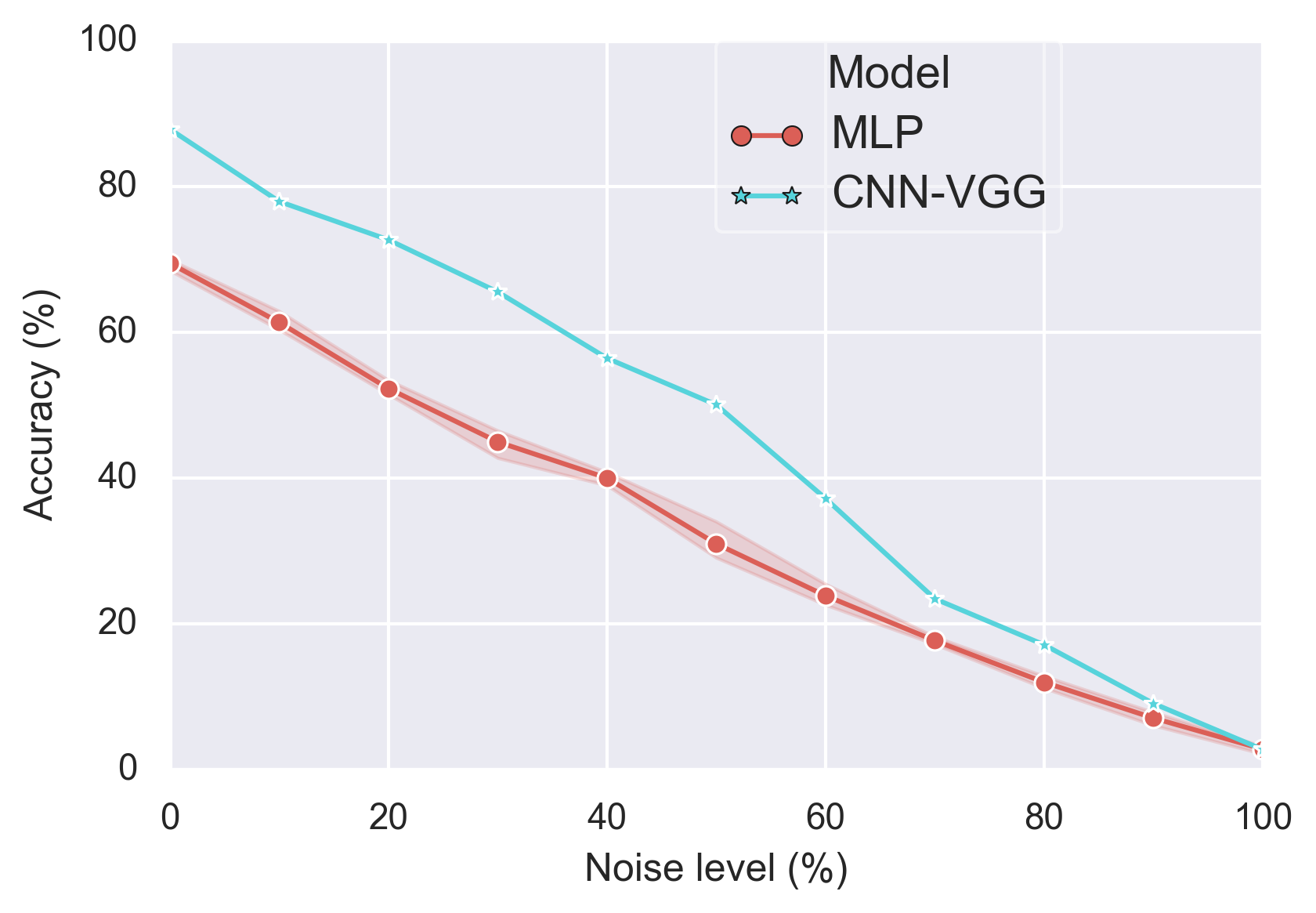}
			\label{fig:NoisyData-Classif-Facescrub}
		}
		\caption{Impact of noisy data on anomaly classification}
		\label{fig:NoisyData-Classif}
	\end{center}
	\vspace{-1.5em}
\end{figure*}

The second strategy relies on voting based algorithms to mitigate possible biases stemming from a chosen single filter.
\cite{brodley1996,DBLP:conf/hais/MirandaGCL09,piyasak2010} simultaneously train several classifiers directly on the original data. Afterwards, they use either majority vote (i.e. classify a sample as mislabeled if a majority of classifiers misclassified it) or consensus vote (i.e. classify a sample as mislabeled if all classifiers misclassified the sample) to filter noisy data. There are similarities between these algorithms and our \systemVoting and \systemActive. However, these solutions focus on static datasets and the off-line setting. They do not consider the learning efficiency and training limitation for on-line scenarios. Since the interval between data batches can be short, we need to ensure the training and inference times per batch are as short as possible. The two models in our \system are connected in cascade. Only data instances deemed uncertain by the first model get to be predicted twice. Off-line voting methods instead need to process each data record multiple times, once for each classifier trained by the algorithm. \cite{piyasak2010} trains two neural networks on top of the classification model. In image classification, training three big CNNs can be hardware-impossible on many devices or very slow and not suitable for on-line learning.

Using active learning in data cleaning has been explored in pattern recognition research. \cite{guyon1994} proposes to define an information criteria function for patterns (data instances). If the information value is below a given threshold, the pattern can be used by the learning algorithm. Otherwise, the pattern is sent to a human expert for checking. The idea is similar to our active learning method, but we go one step beyond by limiting the number of expert queries and proposing an uncertainty-based ranking in \systemActiveLimit and \systemSlimmedLimit. 
Only the most valuable instances are thus selected to expert cleansing.

\section{Motivating case studies}
\label{sec:ProblemStatement}


To qualitatively demonstrate the impact of noisy data on anomaly detection, we use three case studies.

\begin{itemize}
\item Detecting \textbf{IoT device attacks} from inspecting network traffic data collected from commercial IoT devices~\cite{meidan2018n}. This dataset contains nine types of IoT devices which are subject to 10 types of attacks. Specifically, we focus on the Ecobee thermostat device that may be infected by Mirai malware and BASHLITE malware. 
Here we focus on the scenario of detecting and differentiating between 10 attacks. It is important to detect those attacks with high accuracy against all load conditions and data qualities.

\item Predicting \textbf{task execution failures} for big data jobs running at a Google cluster~\cite{reiss2011google,Rosa:TSC17:failurePrediction}. This trace contains a month-long job execution records from Google clusters. Each job contains multiple tasks, which can be terminated into four different states: \textit{finish}, \textit{fail}, \textit{evict}, or \textit{kill}. The last three states are considered as anomalous states. To minimise the computational resource waste due to anomalous states, it is imperative to predict the final execution state of task upon their arrivals.

\item Recognizing \textbf{celebrity faces} from photos of the FaceScrub dataset~\cite{facescrub:2014}. The set is a collection of photos of celebrities roughly half female and half male. The task is to recognize faces by matching each photo to the identity of the celebrity shown on it. Here we focus on the face recognition of the 100 celebrities with the highest number of photos in the dataset totalling to 12K images. 
Faces are widely used in biometric identification systems in many security applications, e.g., access control. This makes the robustness of such systems critical. Furthermore, this image dataset is studied because we want to show the broad applicability of our proposed framework.
\end{itemize}

The details about data definition, and statistics, e.g., number of feature and number of data points, can be found in Section~\ref{ssec:Datasets}. To recognize anomalies/faces in each use case, related studies have applied different machine learning classification algorithms, from simple ones, e.g., k-nearest neighbour (KNN), to complex ones, e.g., deep neural networks (DNN), under scenarios with different levels of symmetric label noise.
Noisy labels are corrupted with equal probability to all classes except the true one.
Here, we evaluate how the detection accuracy changes relative to different levels of noises. We focus on off-line scenarios where we split the data in a training set affected by label noise and a clean evaluation set. Due to our focus on resistance to noisy labels, we repeat experiments by regenerating the noise while keeping the model initialisation constant.

\subsection{Anomaly Detection}

Classification models are learned from 14,000 training records and evaluated on a clean testing set of 6,000 records. We specifically apply KNN, nearest centroid and multilayer perceptron (MLP) (a.k.a feed-forward deep neural networks) on both the IoT device attacks and the cluster task failures.
\zilongout{For MLP, we fix the seed of the random number generator to ensure the model initialization is identical for each run. The randomness across runs only stems from the label noise injected into the training data.}
We repeat all experiments 10 times\footnote{\label{note1} To verify and reproduce the results the code is available at \url{https://github.com/zhao-zilong/MotivationCaseStudies}}. Fig.~\ref{fig:NoisyData-Classif-IoT} and Fig.~\ref{fig:NoisyData-Classif-ClusterTasks} summarize the accuracy results.

One can see that noisy labels clearly deteriorate the detection results for both IoT attacks and task failures, across all three classification algorithms. For standard classifiers, like KNN and nearest centroid, the detection accuracy decays faster than MLP which is more robust to noisy labels. Such an observation holds for both use cases. For IoT attacks, MLP can even achieve a similar accuracy as the no label noise case, when 40\% of label classes are altered. Moreover, the impact of noise depends also heavily on the specific sequence of label noise. Corrupting the labels of some samples has a higher impact than corrupting others. As a consequence the curve is highly unstable with large variances and leads sometimes to counter intuitive results of non monotonic impact of noise level  on accuracy. An example is given by the nearest centroid results on the cluster task dataset. Even across 100 runs the mean accuracy at 30\% noise is slightly lower than the mean accuracy at 40\% noise.



\subsection{Face Recognition}
For face recognition we use a subset of our complete dataset (which contains 100 celebrities). The subset contains 2,639 images from 20 celebrities with varying degrees of label noise as training set and 665 clean images as testing set. Due to the complex features of image data, we use a MLP and convolutional neural networks (CNN). Specifically we use a small VGG~\cite{Simonyan15} with 6 convolutional layers. \zilongout{Similar to before, for MLP and VGG we fix the seed of the random generator, so that the model initialization is the same for each run. The randomness across experiments only stems from the non deterministic label noises in the training data.} We repeat MLP experiments 10 times, and VGG experiments 3 times due to the higher training complexity.
Fig.~\ref{fig:NoisyData-Classif-Facescrub} shows the accuracy results under different label noise levels. Similar to previous use cases, one can observe that label noise strongly affects the performance of both classifiers. The accuracy degradation is approximately linear with the noise level. VGG outperforms MLP in this dataset under all noise levels except 100\% corrupted labels.

\begin{figure}[t]
    \vspace{-1.0em}
	\centering
	\subfloat[Original data]{
		\includegraphics[width=0.47\columnwidth]{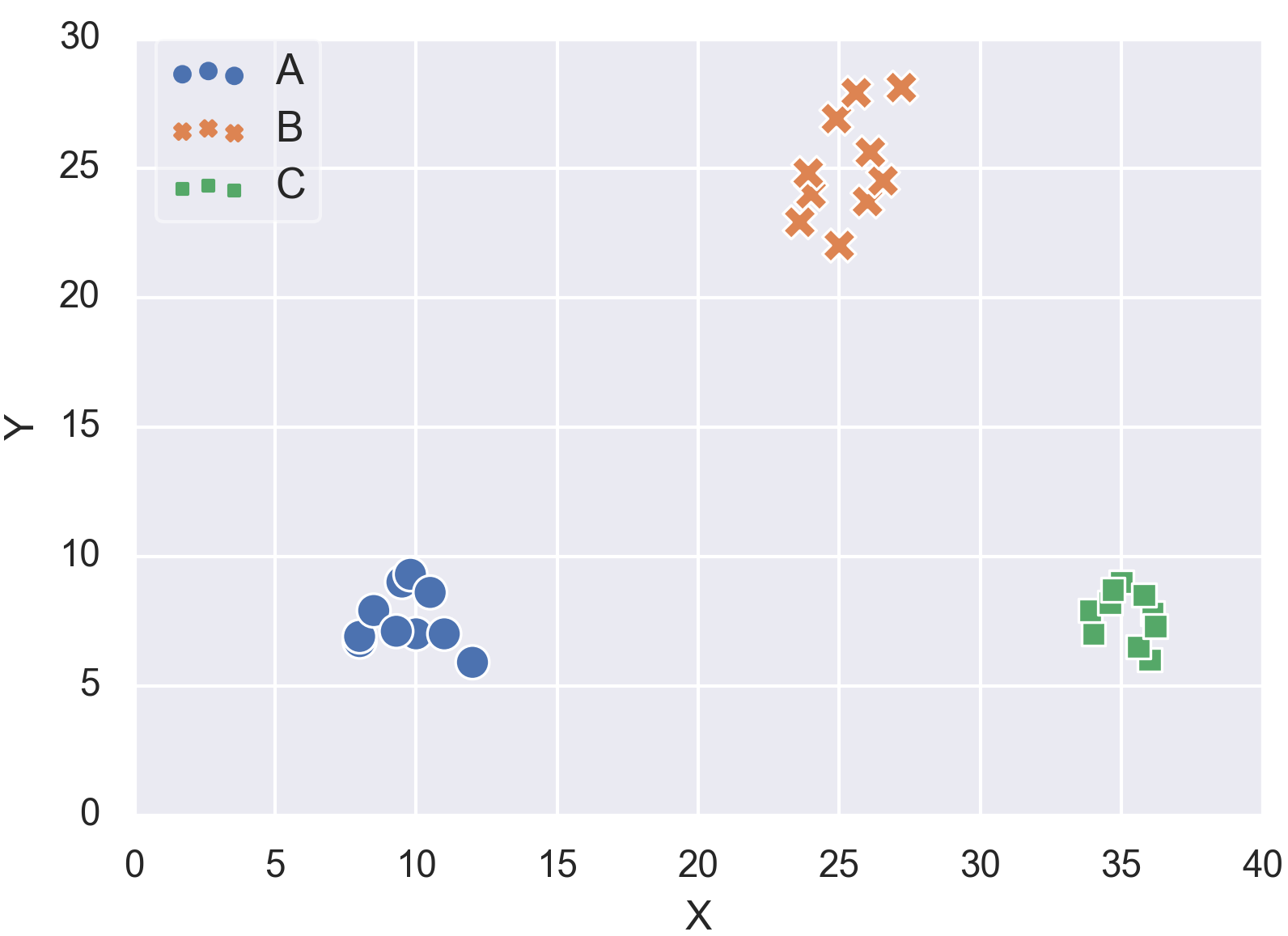}
		\label{fig:original_points_scatter}
	}
	\hfil
	\subfloat[100\% noisy label data]{
		\includegraphics[width=0.47\columnwidth]{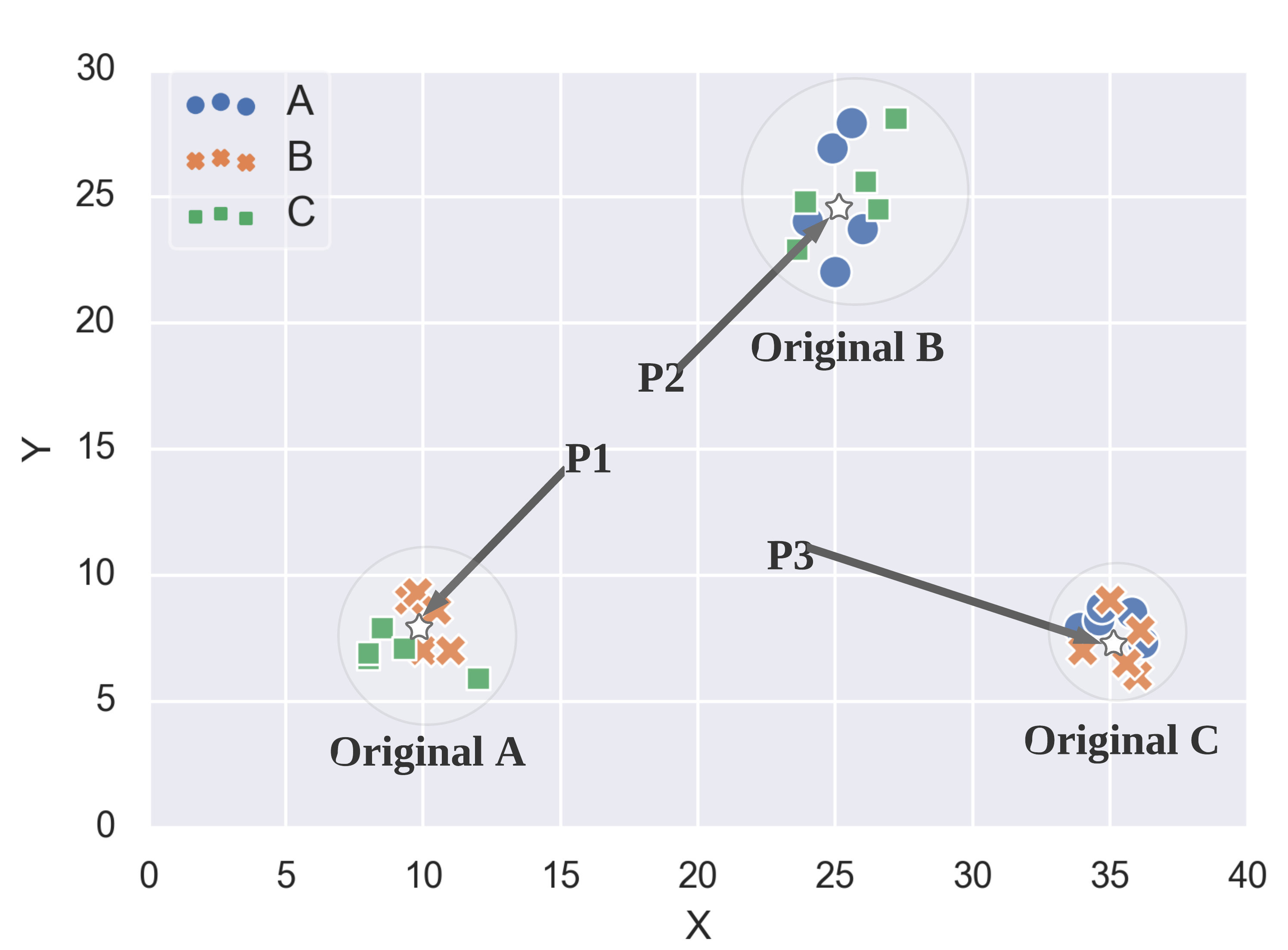}
		\label{fig:noisy_points_scatterplot}
	}
	\caption{Example of training under 100\% label noise.}
	\label{fig:100_percent_noise_points_scatter}
	\vspace{-1.0em}
\end{figure}

Although it is rare to encounter a dataset with 100\% noisy labels, it is still an interesting scenario to study. Almost all accuracy curves in Fig.~\ref{fig:NoisyData-Classif} reach near 0\% under 100\% noise. This can be counter intuitive as illustrated by the following example. If the dataset contains $K$ balanced classes, one might think that it should be possible to obtain an accuracy of $\frac{1}{K}$ just by guessing.
However training on 100\% noisy label data is worse than random guessing. We illustrate this via a simple example with three classes, A, B and C, and 10 samples per class. Fig.~\ref{fig:original_points_scatter} shows the original sample distribution. Fig.~\ref{fig:noisy_points_scatterplot} shows the sample distribution with 100\% label noise. Since all labels are corrupted, each original cluster only contains labels of wrong classes. If we train a machine learning model (e.g., KNN with $k=5$) on this noisy label data, we learn a wrong model which can misclassify any data point. 
See the highlighted points P1, P2 and P3 in Fig.~\ref{fig:noisy_points_scatterplot} as examples. Training on 100\% noisy label data is hence worse than zero-knowledge guessing because fully corrupted data can mislead the learning process.


\begin{figure}[t]
    \vspace{-0.8em}
	\centering
	\subfloat[IoT attack]{
		\includegraphics[width=0.47\columnwidth]{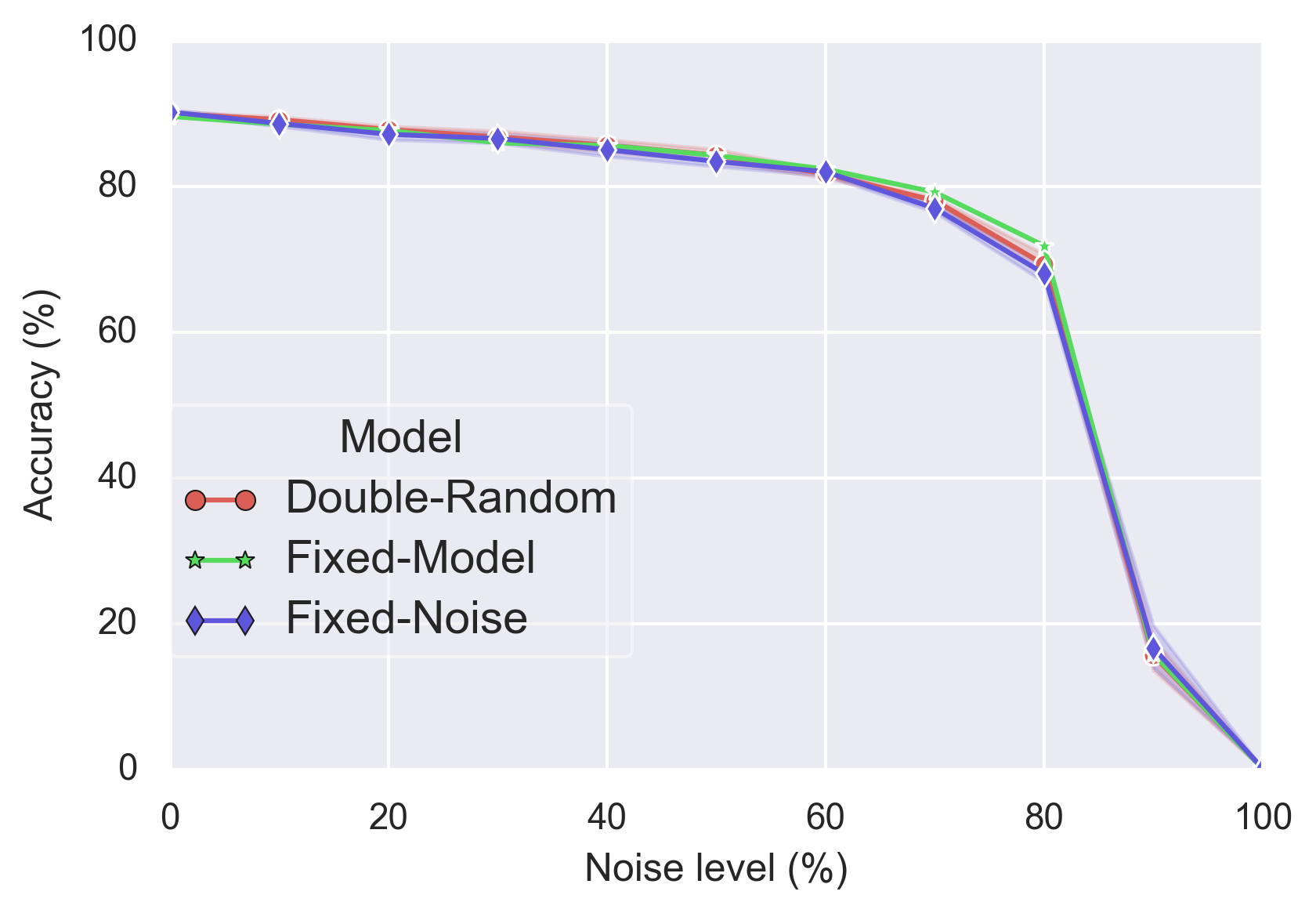}
		\label{fig:thermostat_noise_type}
	}
	\hfil
	\subfloat[Face recognition]{
		\includegraphics[width=0.47\columnwidth]{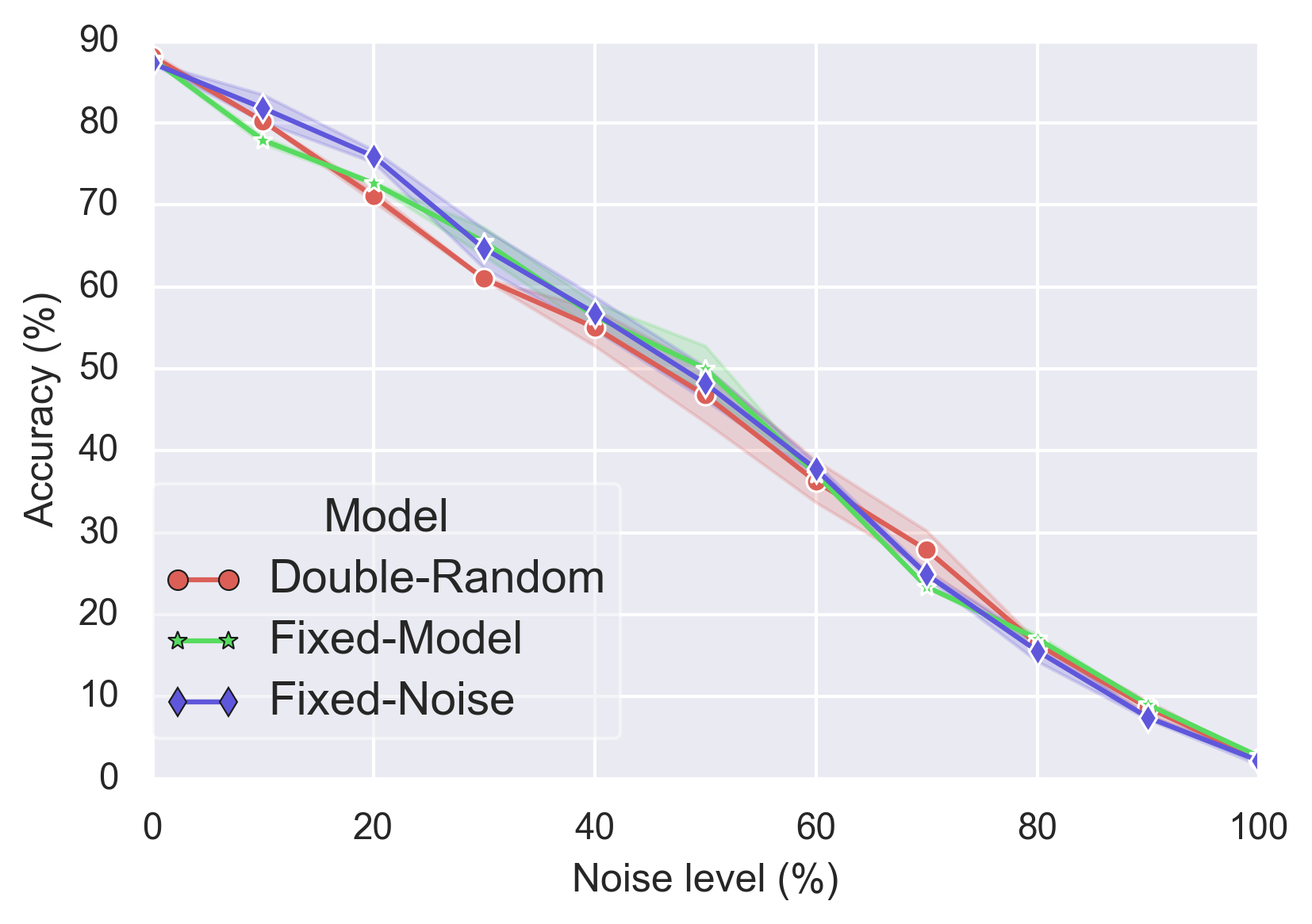}
		\label{fig:task_noise_type}
	}
	\caption{Impact of model vs. noise randomness across runs.}
	\label{fig:noise_type}
	\vspace{-1.0em}
\end{figure}

We fix the model initialization and regenerate the noisy data across experiment runs. The randomness of the results only stems from the noise injected into the data labels used for training.
Before choosing this setting, we run preliminary experiments  with different types of randomness. Fig.\ref{fig:noise_type} shows the results on the IoT attack and Face recognition datasets using MLP and VGG, respectively. It compares: \emph{Fixed-Model} initialization with regenerated noise -- our setting throughout the paper; \emph{Fixed-Noise} with random model initialization; and \emph{Double-Random} with regenerated noise and random model initialization. Both cases show significant overlaps between the three types of randomness, especially for MLP. Due to the lower number of runs (3 against 10) and higher model complexity, the VGG results are slightly more dispersed. Neither case shows significant impact of the randomness type on the results. As our study focuses on the influence of noisy label data, we choose \emph{Fixed-Model} for the remainder of the paper.

\zilongout{One thing to notice is that all the algorithms used in this paper are initialized with a fixed random state or random seed, the randomness of the results only come from the noise injected into the training data labels. Before choosing this setting, we have implemented three experiments for IoT Cluster and Face recognition datasets that: (1) the initialization of model is random, the noise injected into training labels is random, (2) the initialization of model is fixed, the noise injected into training labels is random, (3) the initialization of model is random, the noise injected into training labels is fixed. Results show that neither randomness from initialization of model nor randomness from noise shows particular influence for the results. As our study focus on the influence of noisy label data, we stick to our current setting. Due to page limit, we do not show more details on this.}

The above three experiments clearly show that under the presence of noisy label data, all models are progressively degraded. 
These cases motivate us to design the \system framework and its extension to counter the influence of noisy label data on the learning process.
\section{Design Principles of \system Framework}
\label{sec:Framework}



In this section, we introduce the system model followed by the general structure of \system and its extended features with respect to data selection and model prediction -- ensemble prediction.
All used symbols are summarized in Table~\ref{tab:symboltable}.

\subsection{System Model}
\label{ssec:Definitions}


We consider a dataset that consists of several data instances.
Each data instance has $f$ features.
Each data instance belongs to a class $k$, where $k \in \mathcal{K}=\{1, \dots, \textit{K}\}$.
Data instances are part of a pre-labeled dataset $\mathcal{D}$ with labels $Y$ used for training.
Furthermore, a labeled data instance is either correctly labeled (i.e.,~clean data instance), or incorrectly labeled (i.e.,~noisy data instance). We use the indicator variable $\hat{q}$ to indicate clean $\hat{q}=1$ and dirty $\hat{q}=0$ labels. Wrong labels can stem from several reasons ranging from subjectivity and data-entry errors, to malicious error injection.
The quality of a dataset $\mathcal{D}$ is measured as the percent of clean labeled data instances, denoted here as $\tilde{Q}$.

\begin{table}[t]
	\begin{center}
		\caption{Symbol description}
		\label{tab:symboltable}
		\begin{tabular}{L{1cm} C{6cm} }
			\toprule
			\textbf{Symbol}	& \textbf{Description}\\
			\midrule
			$\mathfrak{L}$ 	& label quality predictor					 \\
			$\mathfrak{C}$ & anomaly detection classifier					\\
			$\mathcal{D}_i$	&  $i^{th}$ training data batch 				\\
			$\mathcal{D}_i^*$	& $i^{th}$  cleansed data batch from 	$\mathfrak{L}$			\\
			$\mathcal{P}_i$	& $i^{th}$ test data batch			\\
			$\hat{Y}_i$	& prediction of $i^{th}$ test data batch from 	$\mathfrak{C}$		\\
			$\tilde{Q}_i$ 	& percent of clean labeled data of $i^{th}$ batch				    \\
			
			$\mathcal{U}_i$	& ``unclean'' data of $i^{th}$ batch determined by $\mathfrak{L}$	 \\
			$\mathcal{U}_i^*$	& $i^{th}$ cleansed data batch from $\mathfrak{C}$			\\
			$\mathcal{S}_i$	& ``unclean'' data  of $i^{th}$ batch determined by $\mathfrak{C}$		\\
			$\mathcal{S}_i^*$	&   data with true label from Expert of $i^{th}$ batch	\\
			$\hat{p}$ & indicator of prediction, 1 for clean, 0 for dirty \\
			$\hat{q}$ & indicator of prediction, 1 for clean, 0 for dirty \\
			$\alpha$ & accuracy on testing set \\
			\bottomrule
		\end{tabular}
	\end{center}
\end{table}

Data instances arrive at the learning system continuously over time in batches.
$\mathcal{D}_i$ denotes the batch of labeled data arriving at time $t_i$ and having labels $Y_i$. In general we denote the time window with the subscript $i$.
We assume that a small initial batch of data instances $\mathcal{D}_0$ has only clean labels, that is $\tilde{Q}_0 = 100\%$.
Subsequent batches, include varying proportions of noisy labels, i.e~$ 0 < \tilde{Q}_i < 100\% , i > 0$.
For simplicity we consider arriving batches of equal size, $\forall \mathcal{D}_i, |\mathcal{D}_i| = N $, but not necessarily at regular times.

A classification request consists of a batch of non-labeled data instances $\mathcal{P}_i$ for which the classifier predicts the class $k$ of each data instance. At each batch arrival, the classification output $\hat{Y}_i$ is thus an array of the predicted classes for each non-labeled data instance. 

\subsection{Design Overview of \system}
\label{ssec:Overview}

We propose the \system learning framework. Its objective is threefold:

\begin{enumerate}

   	\item[(1)] Learn accurate models from noisy data.

	\item[(2)] Continuously update the learned models based on new incoming data.

	\item[(3)] Propose a general approach that fits to different machine learning algorithms and different application use cases.
\end{enumerate}

\begin{figure}[t]
    \centering
    \includegraphics[width=\linewidth]{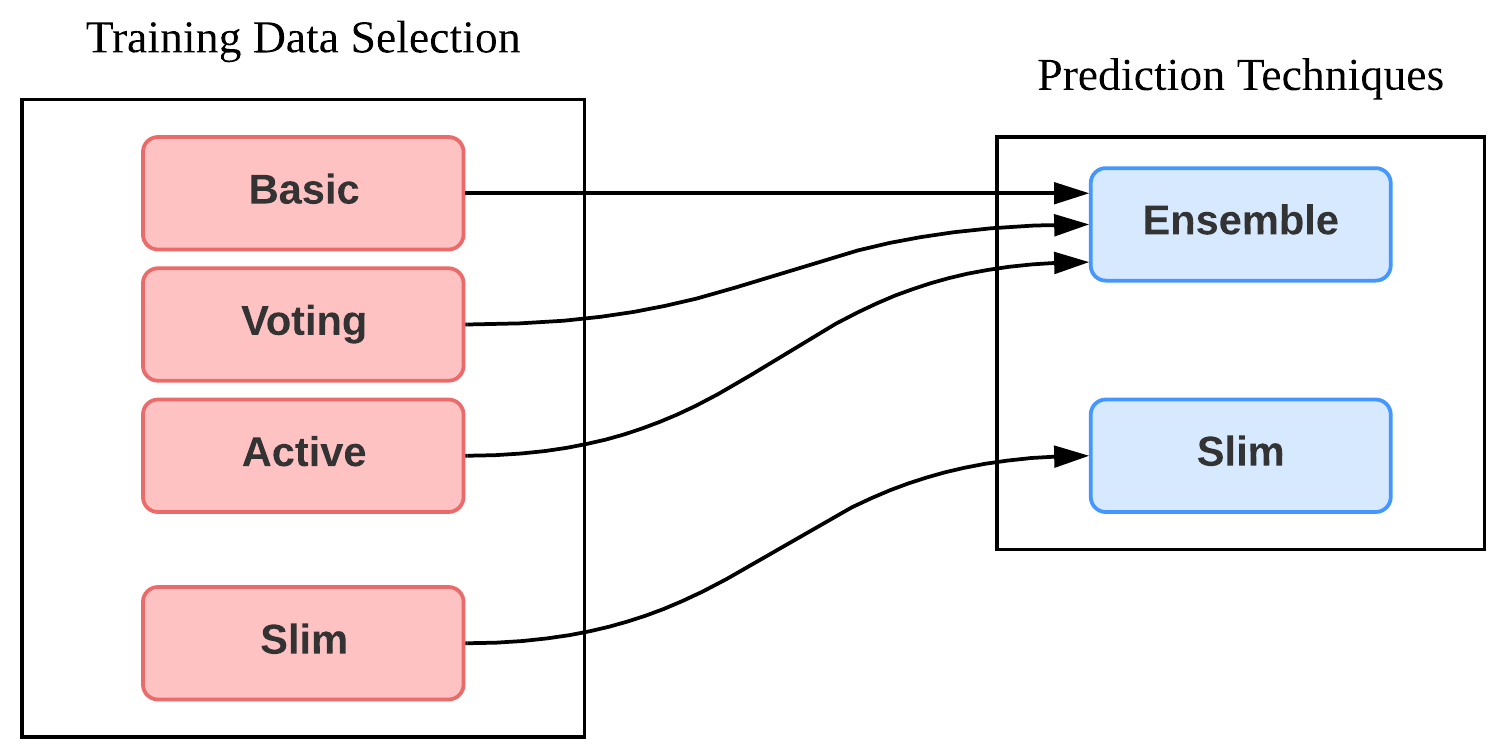}
    \caption{Structures of RAD and its extensions: four choices of data selection and two choices of class prediction.}
    \label{fig:rad_component}
    \vspace{-0.5em}
\end{figure}

 \system is composed of two key steps: training data selection and class prediction, as shown in Fig.~\ref{fig:rad_component}.
Training data selection focuses on how to filter out suspicious noisy data instances and solicit \textit{clean} data to  subsequently train the classification model. It has four options: \textit{basic}, \textit{voting}, \textit{active} and \textit{slim}. The class prediction uses different prediction techniques. Available options are \textit{ensemble}, which combines the prediction outcomes of quality and classification models; and \textit{slim}, which has only one model to filter and classify anomalous images. We consider the following specific combinations: (i) \textit{basic}, \textit{voting}, and \textit{active} are followed by the \textit{ensemble} prediction; (ii) \textit{slim} is followed by the \textit{slim} prediction, which only uses one model to save computation resources.

\begin{figure}[tb]
    \centering
    \includegraphics[width=\linewidth]{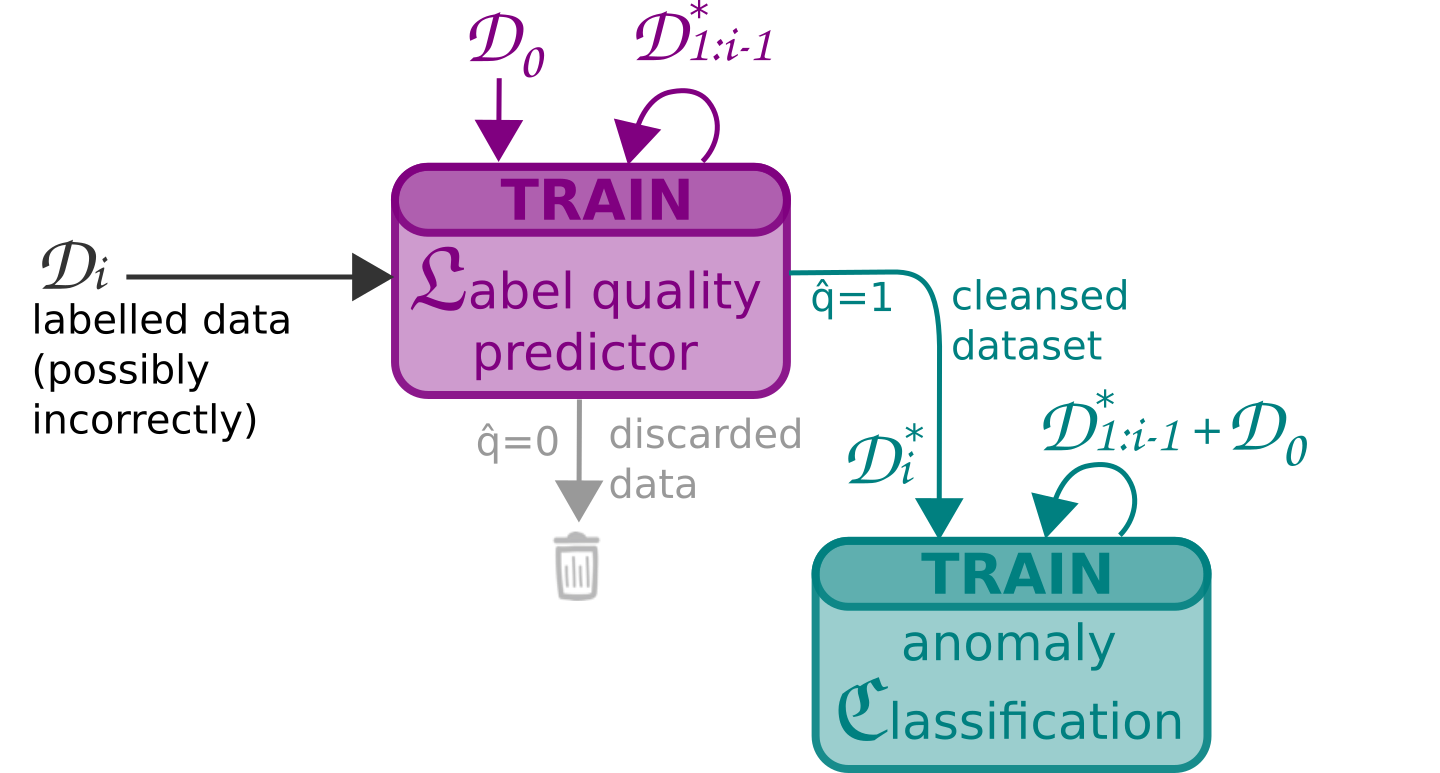}
    \caption{\system training data selection framework. Each block is a machine learning algorithm. 
    Data used to train is represented by colored arrows from the top. The flowchart is iterated at every batch arrival with new labelled and unlabelled data coming in (black arrows on the left).  The labelled training data for $\mathfrak{C}$ is cleansed based on the label quality predicted by $\mathfrak{L}$. 
    }
    \label{fig:architecture}
    \vspace{-0.5em}
\end{figure}

Fig.~\ref{fig:architecture} describes the overall architecture of \system training data selection. it comprises two main components. A label quality model $\mathfrak{L}$ mainly aims at discerning clean labels from dirty labels and a classifier model $\mathfrak{C}$ targets the specific classification task at hand. But both models are used for the ensemble predictions, described in Sec.~\ref{ssec:prediction_technique}.

\system follows a generic approach since the proposed classification framework can be used with any supervised machine learning algorithm, such as SVM, KNN, random forest, nearest centroid, DNN, etc.
Moreover, \system can be applied to a large spectrum of different applications where noisy data are collected and must be cleansed before used to train the classification model. Examples are the failure detection, attack diagnosis and face recognition illustrated in Section~\ref{sec:Evaluation}.

\subsubsection{Data Selection Scheme}
\label{ssec:DataNoise}

The first component of \system aims to select clean data instances from $\mathcal{D}$ through the quality model. 
The objective of the label quality model is to select the most representative data instances to train a strong classifier model.
It solicits data instances with clean labels, avoiding the pitfall that the classifier overfits to the noise.
\system uses supervised-learning algorithms to continuously train the label quality model from accumulated predicted clean data instances, to build a strong classifier. 

We term the following selection procedure as \textit{basic}, that is the default data selection scheme of \system which requires no addition history data lookup nor involvement of human experts.
$\mathfrak{L}_{i-1}$ is the label quality model that is trained with data instances received up to time $t_i-1$, that is $\mathcal{D}_0 \dots \mathcal{D}_{i-1}$.
Upon the arrival of a new batch of data instances $\mathcal{D}_{i}$ at time $t_{i}$, we use the currently learned label quality model $\mathfrak{L}_{i-1}$ to predict the label quality $\hat{q}$ for each data instance in $\mathcal{D}_{i}$ by comparing the given $k$ and predicted class $\hat{k}^{\mathfrak{L}_{i}}$. If they coincide, we consider the label as clean $q = 1$, otherwise as dirty $q = 0$.
Then we build $\mathcal{D}^{*}_{i}$ as the subset of data instances from $\mathcal{D}_{i}$ with $q = 1$ and discard the instances with $q = 0$. This data flow is summarized in Algorithm~\ref{alg:RAD_and_extensions}.


\subsubsection{Generic Approach to Handle Dynamic Data}
\label{ssec:GenericApproach}


The second component of \system is the data classifier $\mathfrak{C}$, whose input data has dynamic noise ratios.
$\mathfrak{C}_{i}$ is trained on all predicted clean data instances $\mathcal{D^{*}}$ received until time $t_i$, that is $\mathcal{D^{*}}_0 \dots \mathcal{D^{*}}_{i}$.
We assume that $\mathcal{D}_0$ contains only clean data instances to kick-start the framework and use the label quality model $\mathfrak{L}_{0} \dots \mathfrak{L}_{i-1}$ to cleanse $\mathcal{D}_1 \dots \mathcal{D}_{i}$ and produce $\mathcal{D^{*}}_1 \dots \mathcal{D^{*}}_i$.
Thus, the \system framework uses the batch-by-batch updated data label quality model to enrich the training data of the classification model.

\subsubsection{Prediction Techniques}
\label{ssec:prediction_technique}
\begin{figure}[t]
	\centering
	\includegraphics[width=\linewidth]{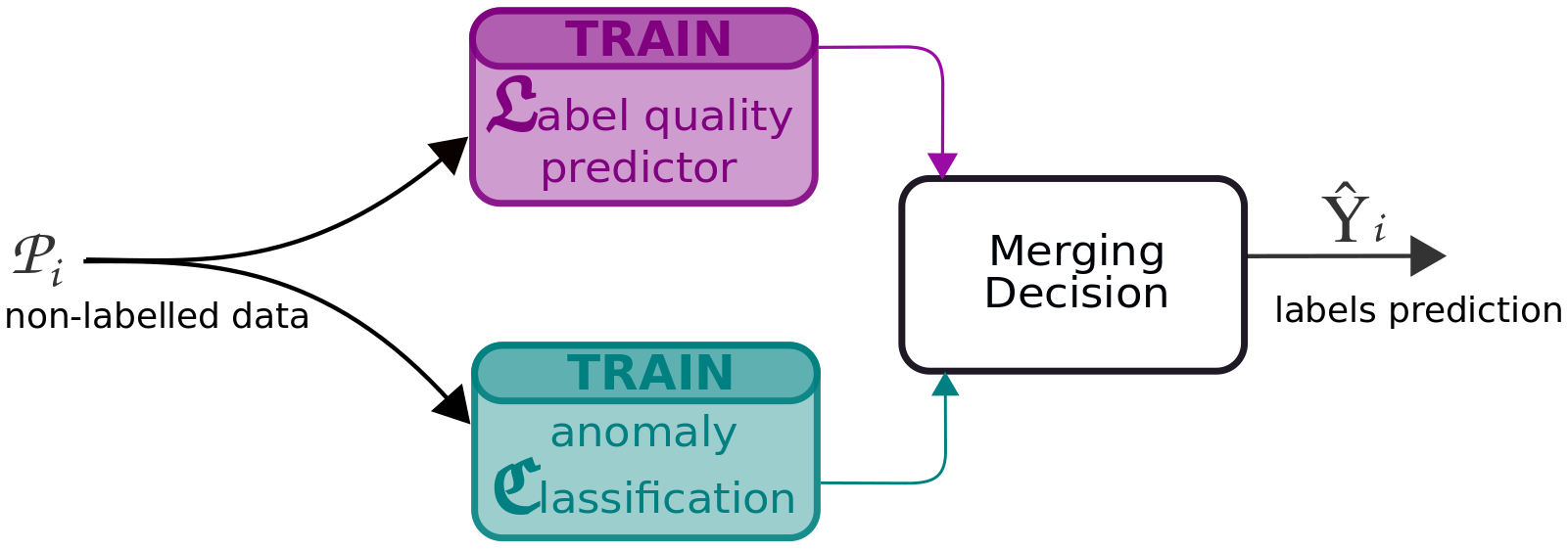}
	\caption{Ensemble Prediction.}
	\label{fig:prediction_ensemble}
	\vspace{-1.0em}
\end{figure}

Fig.~\ref{fig:prediction_ensemble} shows the structure of ensemble prediction, which combines the prediction outcomes of both the quality and classification models. 
The combined decision leverages the confidence from the output probability vectors and the test accuracy of both models from the previous training epoch. If the predictions of the two models coincide, the common prediction is used. If not, we use the prediction of the model having higher confidence. As for the confidence measure, we use the class probability from the output vector multiplied by the test accuracy of the last epoch. We provide the details in Algorithm~\ref{alg:prediction_technique}.

\begin{algorithm}[t]
	\caption{Ensemble Prediction}
	\label{alg:prediction_technique}
	\hspace*{\algorithmicindent} \textbf{Input}: Test data $\mathcal{P}_i$, label quality model $\mathfrak{L}_{i}$, classification model $\mathfrak{C}_{i}$, testing accuracy $\alpha^{\mathfrak{L}_{i-1}}$ of $\mathfrak{L}_{i-1}$ and $\alpha^{\mathfrak{C}_{i-1}}$ of $\mathfrak{C}_{i-1}$.\\ Conv(): convert probability vector to class.
	Max(): return maximum value in a vector.\\
    \hspace*{\algorithmicindent} \textbf{Output}: Predicted labels $\hat{Y}_i$
	\begin{algorithmic}[1]
	\State Get predicted labels for $\mathcal{P}_i$ by $\mathfrak{L}_{i}$ ($Y_{i}^{\mathfrak{L}^{\mathcal{P}}}$) and $\mathfrak{C}_{i}$ ($Y_{i}^{\mathfrak{C}^{\mathcal{P}}}$). Both have length $|\mathcal{P}_i|$ where each element is a vector with probabilities for each of the $K$ classes summing to 1. 
	\State Initialize an empty $\hat{Y}_i$ of length $|\mathcal{P}_i|$
	\For {$j \in \{1,2,\dots, |\mathcal{P}_i|$\}}
	\If{ Conv($Y_{i}^{\mathfrak{L}^{\mathcal{P}}}$[j]) = Conv($Y_{i}^{\mathfrak{C}^{\mathcal{P}}}$[j])}
	\State $\hat{Y}_i$[j] $\leftarrow$ Conv($Y_{i}^{\mathfrak{C}^{\mathcal{P}}}$[j])
	\Else
	\If{$\alpha^{\mathfrak{C}_{i-1}} \times Max(Y_{i}^{\mathfrak{C}^{\mathcal{P}}}[j]) > \alpha^{\mathfrak{L}_{i-1}} \times Max(Y_{i}^{\mathfrak{L}^{\mathcal{P}}}[j])$}
	\State $\hat{Y}_i$[j] $\leftarrow$ Conv($Y_{i}^{\mathfrak{C}^{\mathcal{P}}}$[j])
	\Else
	\State $\hat{Y}_i$[j] $\leftarrow$ Conv($Y_{i}^{\mathfrak{L}^{\mathcal{P}}}$[j])
	\EndIf
    \EndIf
	\EndFor
	\State return $\hat{Y}_i$
	\end{algorithmic}
\end{algorithm}

An alternative prediction technique is \textit{slim} implemented in \systemSlimmed, which relies on one single model for both data selection and classification to save on the computation overhead. We specifically apply \systemSlimmed on image data that demands complex convolutional neural networks.

\subsection{Extended Choices for Data Selection}
\label{ssec:extensions}

In addition to the basic data selection scheme, we provide three additional schemes, namely \systemVoting, \systemActive, and \systemSlimmed. Here, we explain their specific pitfalls and opportunities. 

\subsubsection{\systemVoting}
\label{ssec:systemvoting}

\begin{figure}[t]
	\centering
	\includegraphics[width=\linewidth]{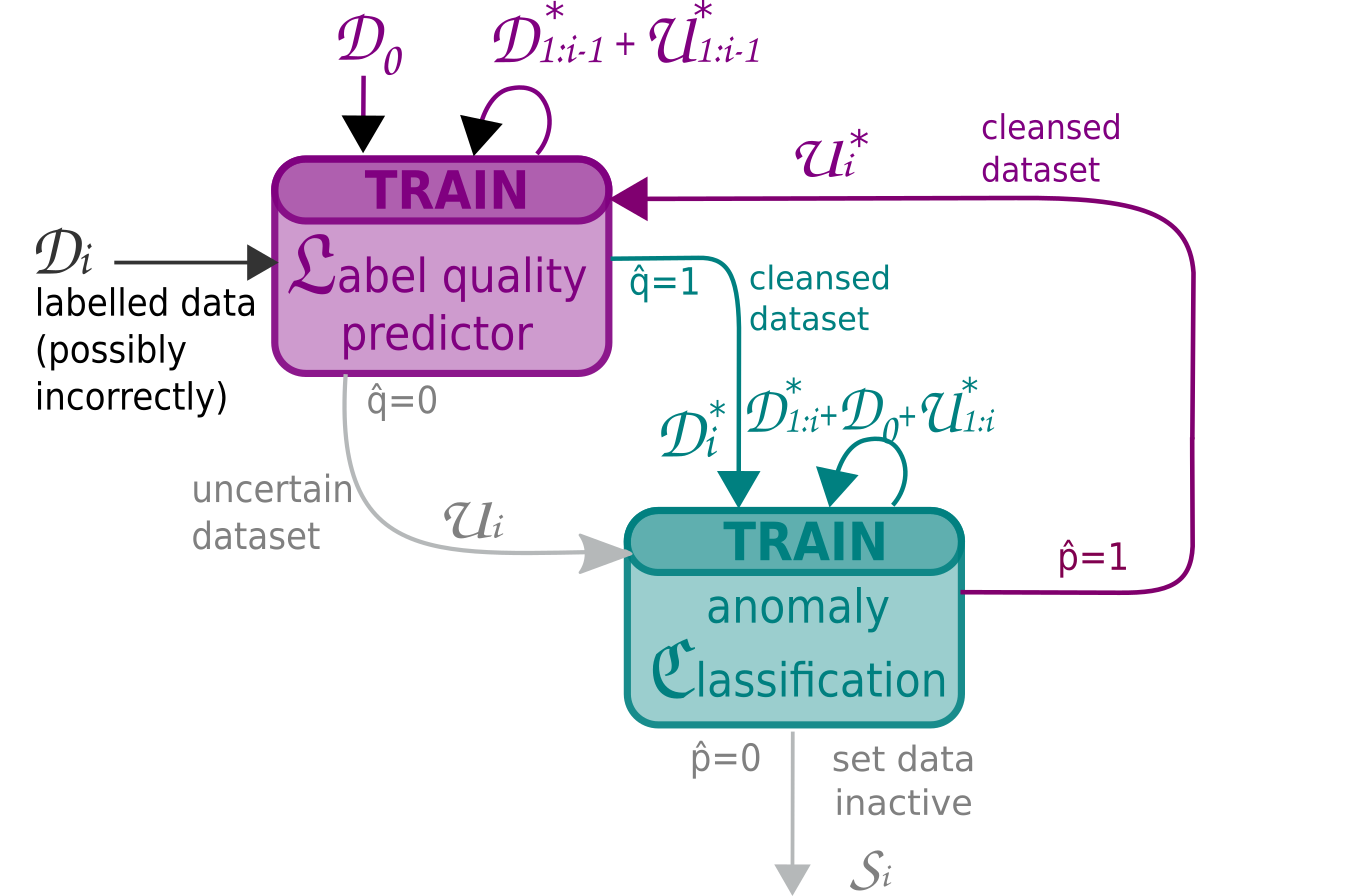}
	\caption{RAD - Voting.}
	\label{fig:architecture_voting}
	\vspace{-1.5em}
\end{figure}

The base \system uses distinctive goals for the two models. However this approach biases the results towards the label quality model $\mathfrak{L}$. We want the classifier model $\mathfrak{C}$ to also play a role in selecting clean data instances. We do this via the voting extension shown in
Fig.~\ref{fig:architecture_voting}.

Comparing to the base \system, predicted dirty labels having $\hat{q} = 0$ are not discarded by $\mathfrak{L}$ but passed to $\mathfrak{C}$ as uncertain data $\mathcal{U}$. Then the classifier $\mathfrak{C}$ further cleanses the uncertain data to produce $\mathcal{U^*}$.
For each data instance in $\mathcal{U}$ we predict its class $\hat{k^{\mathfrak{C}}}$ using $\mathfrak{C}$ and look for agreement with the given class $k$ and the class $\hat{k^{\mathfrak{L}}}$ predicted by $\mathfrak{L}$.
We add data instances to $\mathcal{U^{*}}$ if either $\hat{k^{\mathfrak{C}}}$ equals $k$, or if $\hat{k^{\mathfrak{C}}}$ equals $\hat{k^{\mathfrak{L}}}$. In the latter we replace the given class by the predicted class.

Batches of data instances not added to $\mathcal{U}_i^*$ at time $t_i$ are not immediately discarded but kept in a batch $\mathcal{S}_i$ of inactive data. The idea is that since the accuracy of the classifier improves over time (see Section~\ref{ssec:EvalDynamicData}), we can use the new classifier to re-evaluate old batches of inactive data and further increase the training data.
More in detail we maintain a list $L_{inac}$ of the batches of inactive data $\mathcal{S}_i$. 
After we finish training a new classifier, we select $r$ batches from $L_{inac}$ with the largest number of inactive data and re-process them via the voting system. See more details in Algorithm~\ref{alg:RAD_and_extensions}. The number of batches selected from $L_{inac}$ to re-process is a hyper-parameter. It depends on the time between data batches and the computational efficiency of the training. All training should be finished before the arrival of the next data batch. In our experiments we set $r=2$.

\begin{algorithm}[t]
	\caption{\system, \systemVoting and \systemActive}
	\label{alg:RAD_and_extensions}
	\hspace*{\algorithmicindent} \textbf{Input}: Data batch $\mathcal{D}_i$ with given labels $Y_{i}$, label quality model $\mathfrak{L}_{i-1}$, classification model $\mathfrak{C}_{i-1}$, $r$ reprocessed batches\\
    \hspace*{\algorithmicindent} \textbf{Output: $\mathfrak{L}_{i}$, $\mathfrak{C}_{i}$} 
	\begin{algorithmic}[1]
		\State Predict labels $Y_{i}^{\mathfrak{L}}$ for $\mathcal{D}_i$ using $\mathfrak{L}_{i-1}$.
		\State Create $\mathcal{D}_i^{*}$ as subset of data points in $\mathcal{D}_i$ where $Y_{i}[j] = Y_{i}^{\mathfrak{L}}[j]$ for $j \in 0, ..., |\mathcal{D}_i|$.
		\State $\mathfrak{L}_{i-1}$ sends $\mathcal{D}_i^{*}$ to $\mathfrak{C}_{i-1}$.
		\If {Algorithm is \textbf{\system}}
		\State Retrain $\mathfrak{L}_{i-1}$ and $\mathfrak{C}_{i-1}$ on all accumulated $\mathcal{D}_t^{*}$ t $\in$ [0,i] 
		\hspace*{0.4cm} to get $\mathfrak{L}_{i}$ and
		$\mathfrak{C}_{i}$
		\State \textbf{return $\mathfrak{L}_{i}$, $\mathfrak{C}_{i}$}
		\EndIf
		\If {Algorithm is \textbf{\systemVoting} or \textbf{Active Learning}}
		\State Create $\mathcal{U}_i$ as subset of data points in $\mathcal{D}_i$ where $Y_{i}[j] \neq$  \hspace*{0.4cm} $Y_{i}^{\mathfrak{L}}[j]$ for $j \in 0, ..., |Y_{i}|$.\\
        \hspace*{0.4cm} $\mathfrak{L}_{i-1}$ sends $\mathcal{U}_i$ (with given label $Y_{i}^{U}$) and predictions \hspace*{0.4cm}  $Y_{i}^{\mathfrak{L}^{U}}$ to $\mathfrak{C}_{i-1}$.
		\State Predict labels $Y_{i}^{\mathfrak{C}^{U}}$ for $\mathcal{U}_i$ using $\mathfrak{C}_{i-1}$.
		\If {Algorithm is \textbf{\systemVoting}}
		\State Create $\mathcal{U}_i^*$ as subset of data points in $\mathcal{U}_i$ where: \par
		 \hspace*{0.25cm} for $j \in 0, ..., |U_{i}|$:  \par
		 \hspace*{0.7cm} i) $Y_{i}^U[j] = Y_{i}^{\mathfrak{C}^{U}}[j]$, or \par \hspace*{0.7cm} ii) $Y_{i}^{\mathfrak{C}^{U}}[j] = Y_{i}^{\mathfrak{L}^{U}}[j]$, update $Y_{i}^U[j] \leftarrow Y_{i}^{\mathfrak{C}^{U}}[j]$. 
        \State $\mathfrak{C}_{i-1}$ sends $\mathcal{U}_i^*$ to $\mathfrak{L}_{i-1}$.
        \State Create $\mathcal{S}_i \leftarrow \mathcal{U}_i \setminus \mathcal{U}_i^*$. 
		\State Add $\mathcal{S}_i$ to inactive data list $L_{inac}[i]$.
		\State Select set $r$ batches with highest $|L_{inac}|$. Save\\ \hspace*{0.9cm} indexes in $\mathbf{R}$.
        \State Repeat steps 1-16 for each batch with index $h$ in\\ \hspace*{0.9cm}  $\mathbf{R}$. Use them to update $\mathcal{D}_{h}*$,  $\mathcal{U}_{h}^*$, and $\mathcal{S}_{h}$.
		\State Retrain $\mathfrak{L}_{i-1}$ and $\mathfrak{C}_{i-1}$ on all accumulated $\mathcal{D}_t^{*}, \mathcal{U}_t^{*}$ \hspace*{0.9cm} $t \in [0,i]$ to get $\mathfrak{L}_{i}$ and
		$\mathfrak{C}_{i}$
		\State \textbf{return} $\mathfrak{L}_{i}$, $\mathfrak{C}_{i}$
		\EndIf 
		\If{Algorithm is \textbf{\systemActive}}
		\State Create $\mathcal{U}_i^*$ as subset of data points in $\mathcal{U}_i$ \hspace*{0.95cm} where $Y_{i}^U[j] = Y_{i}^{\mathfrak{C}^{U}}[j]$ for $j \in 0, ..., |U_{i}|$.
        \\ \hspace*{0.95cm} $\mathfrak{C}_{i-1}$ sends $\mathcal{U}_i^*$ to $\mathfrak{L}_{i-1}$.
        \State Create $\mathcal{S}_i \leftarrow \mathcal{U}_i \setminus \mathcal{U}_i^*$. 
		\State Send $S_i$ to Expert. Expert corrects labels and \hspace*{0.9cm} returns $S_i^*$.
		\State Retrain $\mathfrak{L}_{i-1}$ and $\mathfrak{C}_{i-1}$ on all accumulated $\mathcal{D}_t^{*}$, $\mathcal{U}_t^{*}$ \hspace*{0.9cm} and 
	    $S_t^*$ $t \in [0,i]$ to get $\mathfrak{L}_{i}$ and
		$\mathfrak{C}_{i}$.
		\State \textbf{return} $\mathfrak{L}_{i}$, $\mathfrak{C}_{i}$

		\EndIf 
		\EndIf
	\end{algorithmic}
\end{algorithm}

\subsubsection{\systemActive}
In \systemVoting we use $\mathfrak{C}$ and $\mathfrak{L}$ to correct labels and increase the overall amount of data used for training aiming to improve the  framework accuracy. However still not all data is used. To increase further the amount of training data we resort to active learning, i.e., we ask an expert for the true class of the data instances we are least certain.

\begin{figure}[htb]
	\centering
	\includegraphics[width=\linewidth]{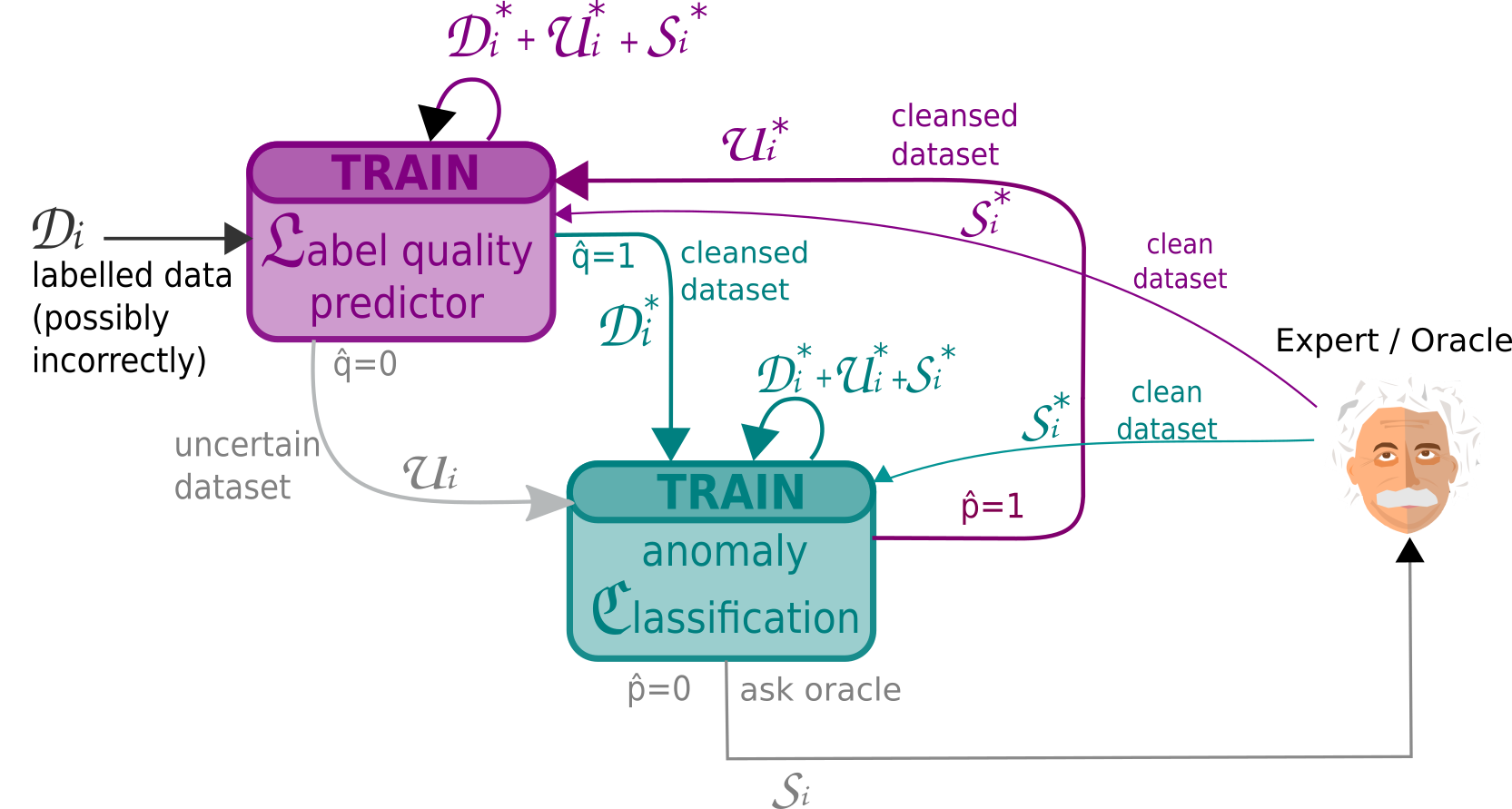}
	\caption{RAD - Active Learning.}
	\label{fig:architecture_oracle}
\end{figure}

Fig.~\ref{fig:architecture_oracle} shows the structure of \systemActive. The difference with \systemVoting is that in \systemActive we do not use the predictions from two models to correct the labels, and we do not send the most uncertain data instances to the inactive list but to an oracle to ask for the true label. In \systemActive, potentially every data instance will be used to train $\mathcal{L}$ and $\mathfrak{C}$ and there is no inactive data anymore.
In practice, consulting an oracle for every single uncertain data instance might be too expensive. 
In \systemActiveLimit we additionally impose a configurable limit $N_{lim}$ on the number of data instances sent to the expert at each batch arrival.
When the number of filtered out data instances exceeds $N_{lim}$, two lists are created:  $RL_{distance}$ and $RL_{std}$. Both measure the uncertainty of data instances. $RL_{distance}$ ranks instances in a decreasing order based on the euclidean distance between the corresponding prediction probability vectors of $\mathfrak{C}$ and $\mathfrak{L}$. $RL_{std}$ ranks instances in an increasing order based on the summed standard deviations of the  corresponding prediction probability vectors of $\mathfrak{C}$ and $\mathfrak{L}$. We alternatively select the top instance from each list until we have $N_{lim}$ instances. Common instances between the two lists are selected only once. We also implement experiments that only sample data from $RL_{distance}$ or $RL_{std}$, but the results are worse. Due to the page limit, we omit the presentation here.
Current method leverages both the different opinions from the two models, and the uncertainty of each model.  We call this the \textbf{\highestActiveLimit}.

\subsubsection{\systemSlimmed}

The \system framework requires two models. Depending on the complexity of the models used, he cost of training might be excessive. Especially in scenarios relying on complex deep neural networks, such as Convolutional Neural Networks (CNNs) for image classification, it might be too expensive and time consuming to train two models. To reduce the computational cost we propose a slimmed version of \systemActive named \systemSlimmed. The idea is to partially delegate the role of the label quality $\mathfrak{L}$ model to the oracle. 

\begin{figure}[htb]
	\centering
	\includegraphics[width=\linewidth]{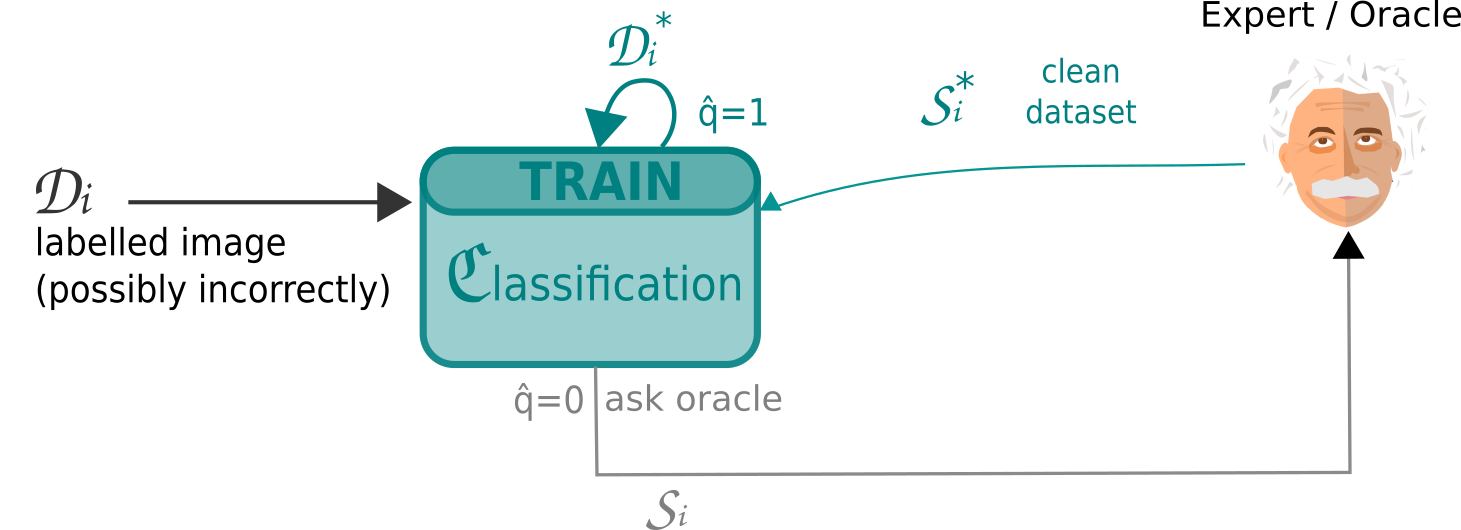}
	\caption{\systemSlimmed.}
	\label{fig:architecture-oracle-image}
	\vspace{-1.0em}
\end{figure}

In \systemSlimmed new data batches arrive directly at the $\mathfrak{C}$ model, see Fig.~\ref{fig:architecture-oracle-image}. For each data instance we compare the given label $k$ to the predicted label $\hat{k^{\mathfrak{C}}}$. If they are the same we add it to $\mathcal{D}^*$. If they differ we ask the oracle for the true label 
and add the answer to $\mathcal{S}^*$. To train the model $\mathfrak{C_{i-1}}$, we use only current $\mathcal{D}_i^*$ plus $\mathcal{S}_i^*$, not all the accumulated cleansed data as before. Considering computational cost, one pair of $\mathcal{D}_i^*$ and $\mathcal{S}_i^*$ will be used to train the model for 60 epochs.
Optionally as before, we can impose a query limit, termed \systemSlimmedLimit. We query experts for the $N_{lim}$ data instances with highest uncertainty ranked by decreasing cross-entropy loss between given label and prediction probability vector made by $\mathfrak{C}$.
We call this the \textbf{Highest Loss Method}.

\section{Experimental Evaluation}
\label{sec:Evaluation}


In this section, we implement \system , \systemVoting and \systemActive on IoT and Cluster datasets and report the evolution of learning accuracy under 30\% and 40\% noise level. For \system , impact of noise level on final accuracy is discussed in Sec.~\ref{ssec:EvalNoiseImpact}. For \systemVoting , analysis on percentage of active and active-truth data changing over time is carried out in Sec.~\ref{ssec:SystemVoting}. \systemActive and its small update \systemActiveLimit are explained in Sec.~\ref{subsec:rad_active_learning}. The impact of the initial data batch size $|\mathcal{D}_0|$ on the above frameworks is studied in Sec.~\ref{ssec:ImpactInitialization}. To demonstrate the applicability of the framework to image dataset, \systemSlimmed and \systemSlimmedLimit are studied in Sec.~\ref{ssec:systemSlimmed}.

\subsection{Use Cases and Datasets}
\label{ssec:Datasets}

In order to demonstrate the general applicability of the proposed \system framework for anomaly detection, we consider the following three use cases:
(i)~Cluster task failures, (ii)~IoT botnet attacks and (iii)~Face recognition.
In our experiments, we use real data collected in cluster and IoT platforms and faces from real celebrity images.

The cluster task traces comprise data instances each corresponding to a task with 27 features capturing information related to static and dynamic system states, e.g. the task start/end times, the task resource utilisations, the hosting machine, etc. Each class is labeled based on its scheduling state. A detailed description of the features and labels can be found in~\cite{reiss2011google}. In particular, we are interested in the four possible termination classes: \textit{finish}, \textit{fail}, \textit{evict}, or \textit{kill}.
We filter out other classes. The resulting class distribution is dominated by successful tasks (\textit{finish}) 77.8\%, followed by \textit{kill} 22.0\%, \textit{fail} 0.2\%, and \textit{evict} $<$0.1\%.
Similar to~\cite{Rosa:TSC17:failurePrediction}, we aim to predict the task outcome to reduce the resource waste and improve the overall scheduling and system performance, e.g., in case of lack of resources and the need to kill a task, help choosing the task with the least probability to succeed. We apply \system to continually train a noise-resistant model for better accuracy. For this dataset we report the F1-score in addition to the accuracy due to the high class unbalance.


The IoT dataset comprises data instances describing 23 network packet-level statistics recursively computed over five different time scales totalling to 115 features. This traffic statistics are collected during normal operation, labeled as benign, or under one of ten different malicious attacks stemming from devices infected by either the {\em BASHLITE} or {\em Mirai} malware. All the classes are evenly distributed in the training and test dataset. Malicious traffic covers mainly scanning for vulnerable devices and various flooding attacks. The dataset provides traces collected at different IoT devices. More details are provided in~\cite{meidan2018n}. We aim to apply \system to build a noise-resistant model to categorize the attacks for post fact analysis, e.g., for threat assessment.

The FaceScrub~\cite{facescrub:2014} dataset is used for face recognition. Original FaceScrub contains more than 100,000 face images of 530 people, with about 200 images per person. Male and Female images are almost equal. We use a subset of 15K FaceScrub images to fit the limits of our compute resources. The 15K images cover the 100 people, 55 males and 45 females, with the highest number of 
images. On average each person (class) has 150 images with a standard deviation of 8.4 images.
We use 12K images as training and 3K as test data.
Training and test datasets have the same data distribution. FaceScrub images were retrieved from the Internet and are taken under real-world situations (uncontrolled conditions). We resize all images to 64x64 pixels. Name is the only annotation we use. Face recognition systems have been widely used in security equipment.
We apply RAD Slimmed to FaceScrub dataset to show that our framework can also help to build robust face recognition models.



The main dataset characteristics are summarized in Table~\ref{tab:datasets}.

\begin{table}[htb!]
	\begin{center}
		\caption{Dataset description}
		\label{tab:datasets}
		\begin{tabular}{L{1.8cm} C{1.8cm} C{1.8cm} C{1.4cm}}
			\toprule
			\textbf{Use case}	& \textbf{Cluster task failures}	& \textbf{IoT device attacks}  & \textbf{FaceScrub}\\
			\midrule
			\#trainig data	& 60,000					& 33,000 & 12,000 \\
			\#test data	& 6,000					& 6,000 & 3,000\\
			\#classes $\textit{K}$ 		& 4 						& 11  &100\\
			\#features $f$ 		& 27 						& 115 & 64*64\\
			data batch size 	& 600					    & 300   & 2400\\
			$|\mathcal{D}_{0}|$ & 6,000 &6,000&2400\\
			\bottomrule
		\end{tabular}
	\end{center}
	\vspace{-1.0em}
\end{table}


\subsection{Experimental Setup}
\label{ssec:ExperimentalSetup}

\begin{figure*}[ht]
	\centering
	\subfloat[With data noise level of 30\%]{
		\includegraphics[width=0.99\columnwidth]{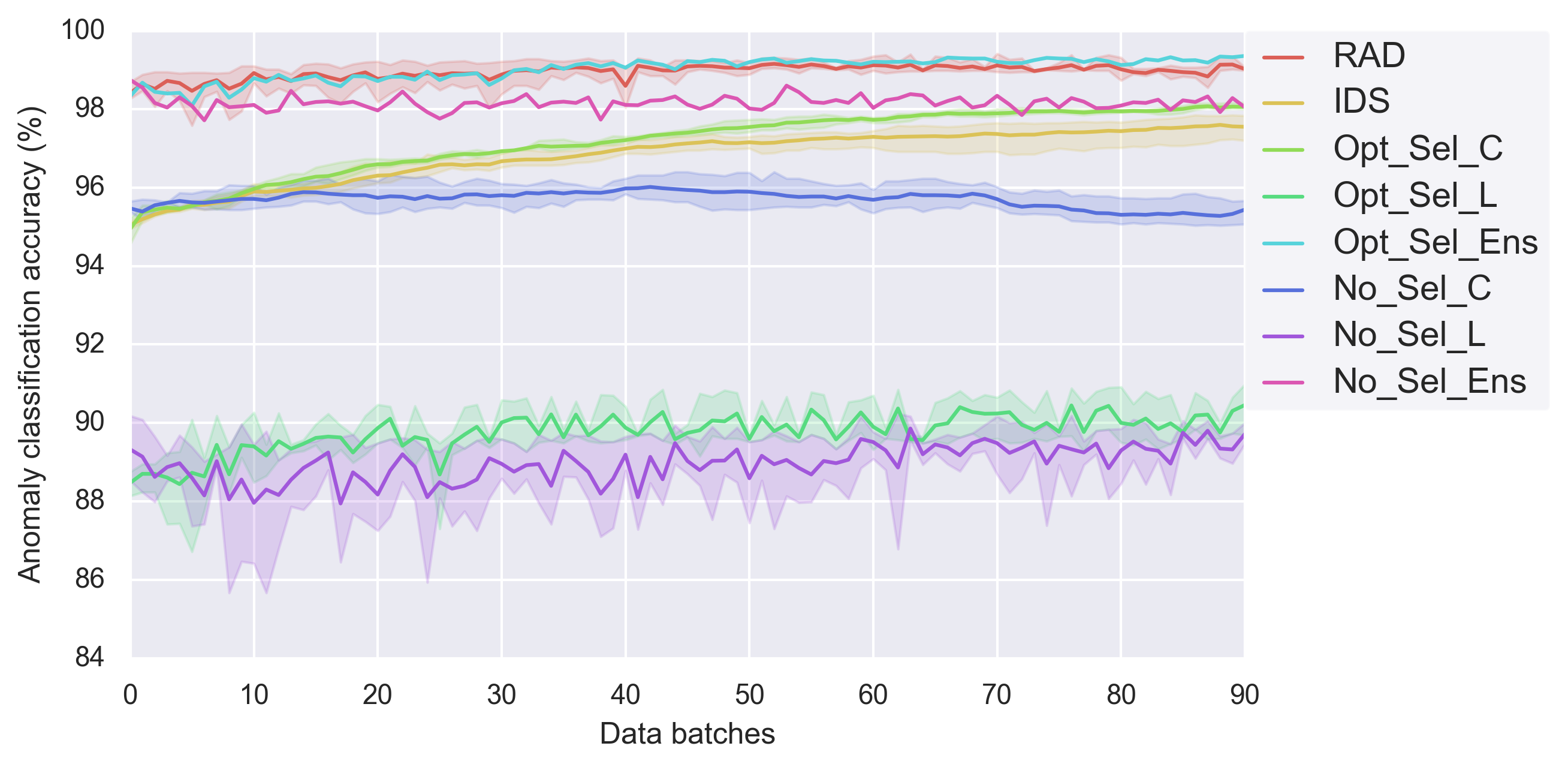}
		\label{fig:EvolOverTime-IoT-30}
	}
	\hfil
	\subfloat[With data noise level of 40\%]{
		\includegraphics[width=0.99\columnwidth]{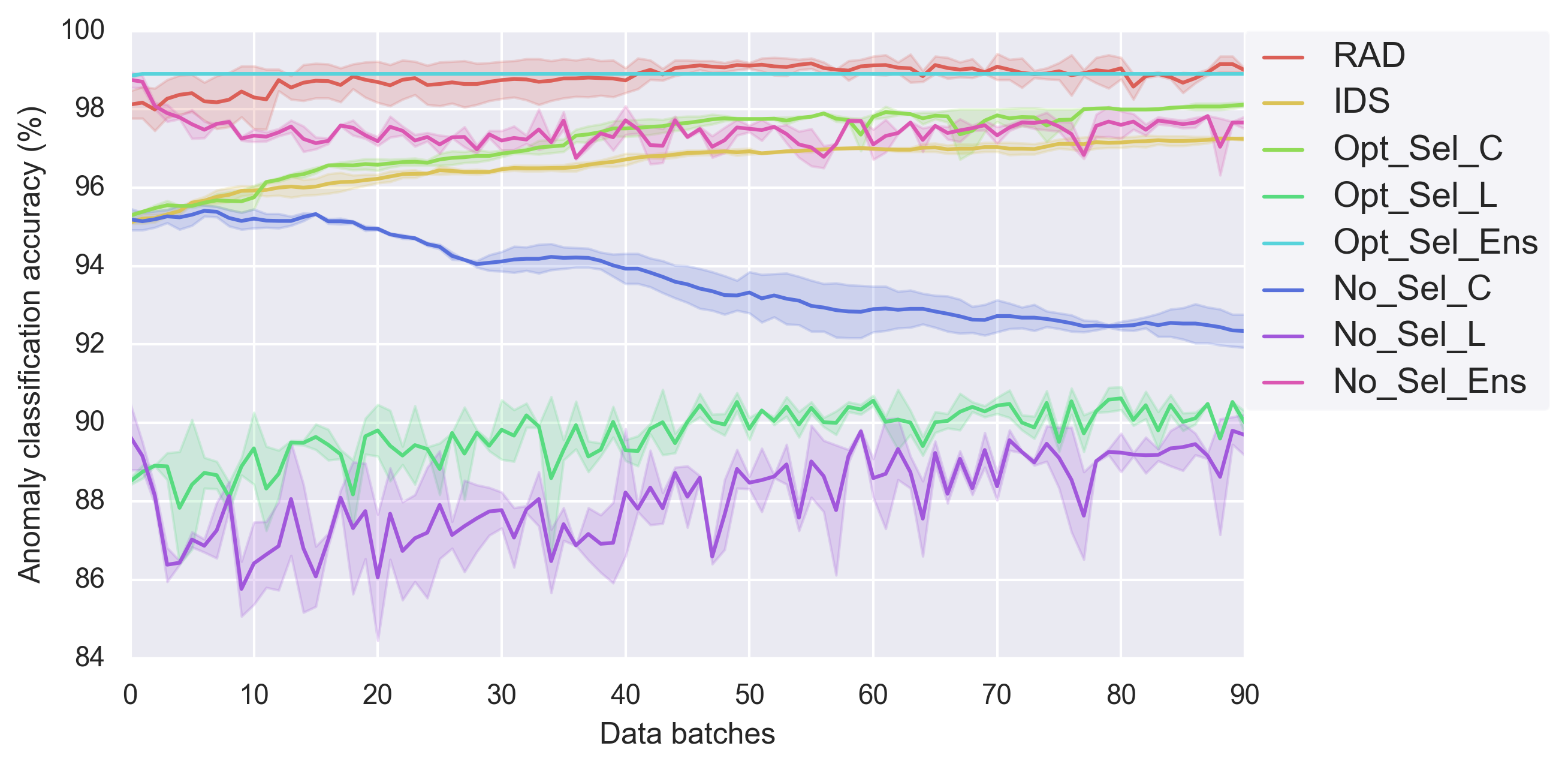}
		\label{fig:EvolOverTime-IoT-40}
	}
	\caption{Evolution of learning over time -- Use case of IoT thermostat device attacks. Opt\_Sel and No\_Sel stand for optimal data selection and no filtering, respectively. \_C, \_L, and \_Ens denote the model or strategy chosen for prediction.}
	\label{fig:EvolOverTime-IoT}
	\vspace{-1.5em}
\end{figure*}

\begin{figure*}[ht]
	\begin{center}
		\subfloat[With data noise level of 30\%]{
			\includegraphics[width=0.99\columnwidth]{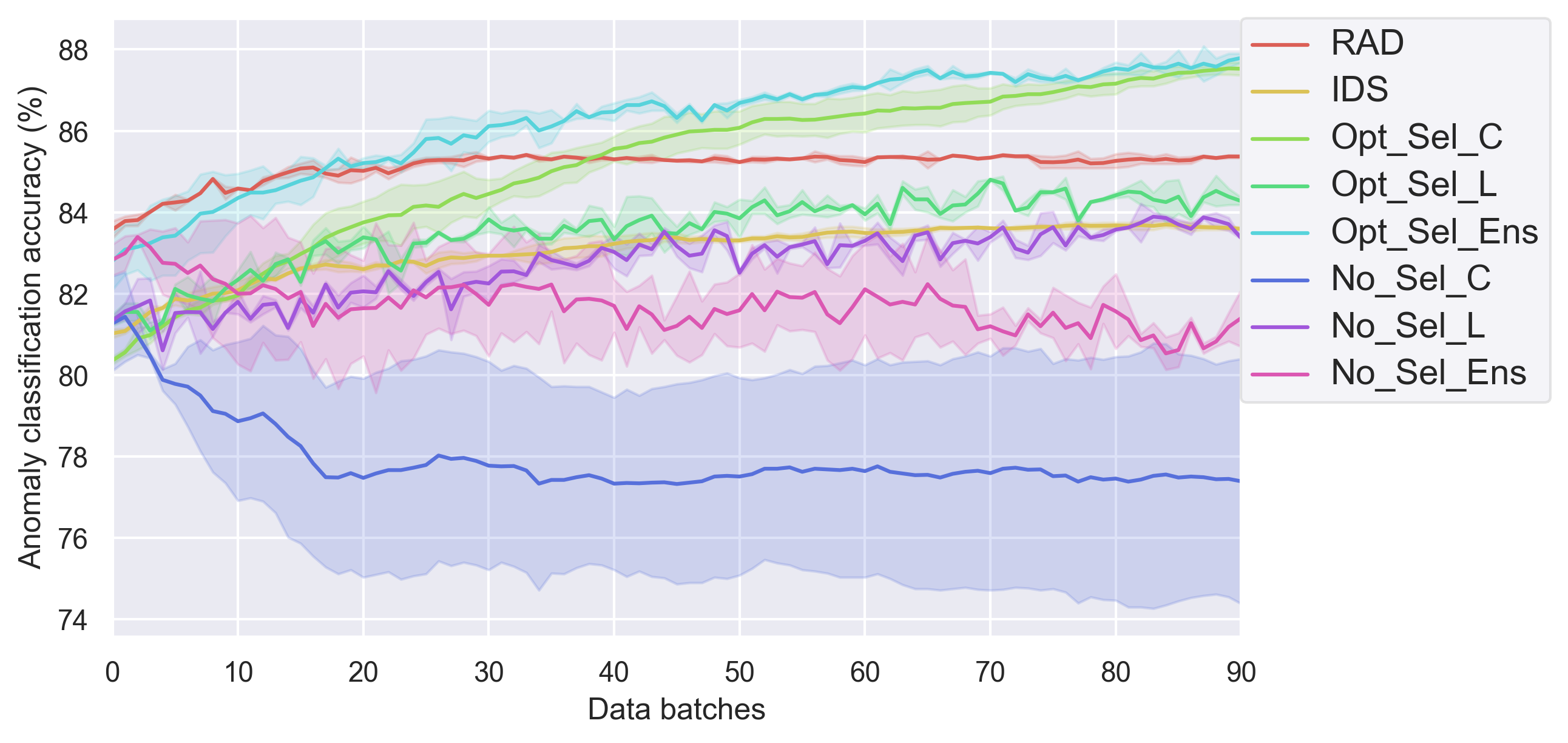}
			\label{fig:EvolOverTime-ClusterTasks-30}
		}
		\hfil
		\subfloat[With data noise level of 40\%]{
			\includegraphics[width=0.99\columnwidth]{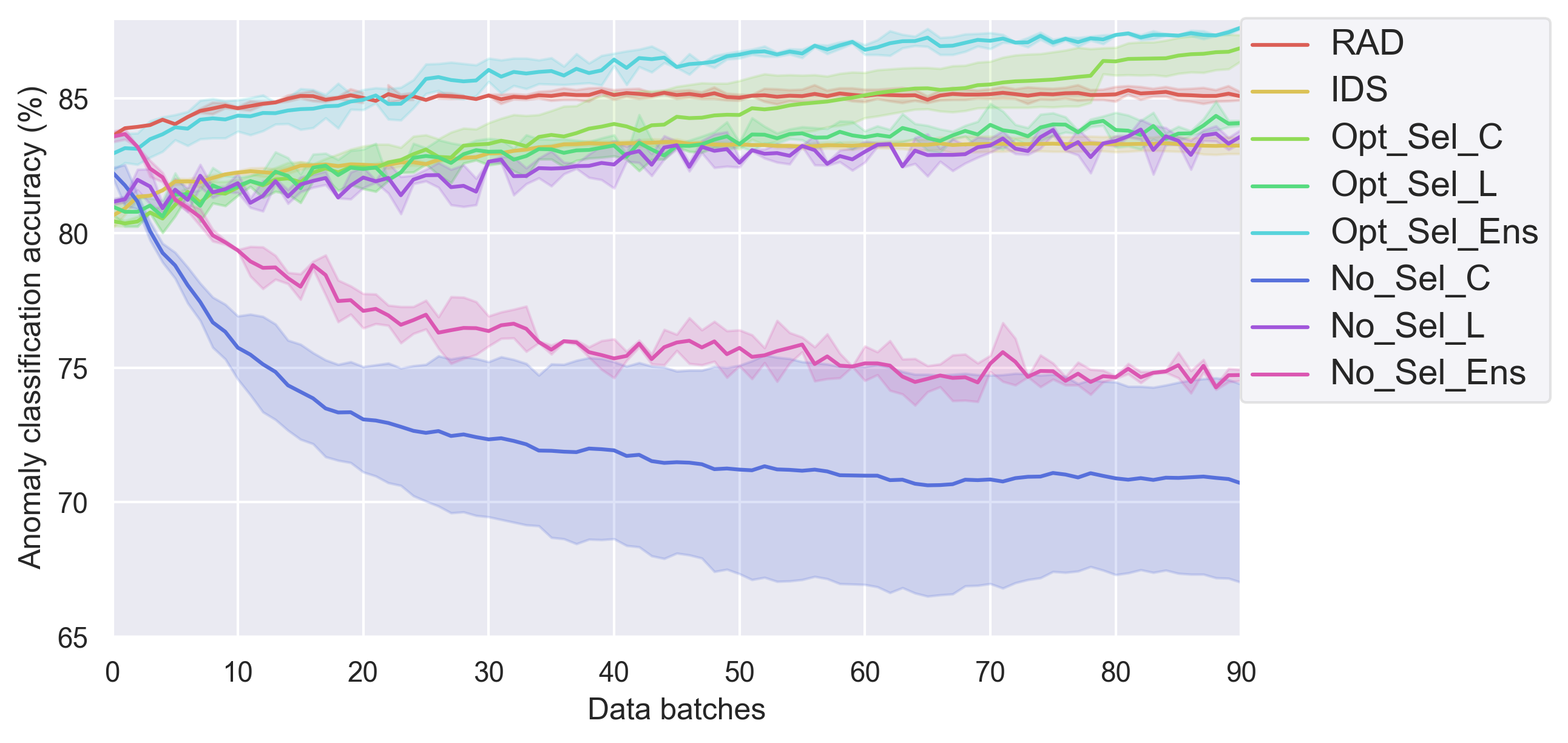}
			\label{fig:EvolOverTime-ClusterTasks-40}
		}
		\caption{Evolution of learning over time -- Use case of Cluster task failures}
		\vspace{-1.5em}
		\label{fig:EvolOverTime-ClusterTasks}
	\end{center}
\end{figure*}

\system is developed in Python using scikit-learn~\cite{scikit-learn}. The main performance evaluation metric is accuracy. All results are averaged across three runs.

{\bf Noise.} We inject noise into the two datasets by exchanging the true label of data instances with erroneous one.
The label noise is symmetric, i.e., following the \textit{noise completely at random} model (NCAR) in~\cite{Frenay:TNNLS14:survey}
where a label is picked with equal probability from all classes except the true one. The noise level $\tilde{Y}$ represents the percentage of data instances with noisy labels. 
We assume that all data is affected by label noise, except $\mathcal{D}_0$ and the testing data. We regenerate the noise at each experiment run.


{\bf Continual learning.} We start with an initial data batch of 6000 data instances for the Cluster task failures and the IoT devices dataset.
Then, data instances arrive continuously in batches of 600 (Cluster) and 300 (IoT) data instances. 
To kick-start the label and classification models in \system we assume first batch contains only clean data, and subsequent data batches are affected by noise.
We select 6000 clean data instances as the test dataset for both use case. Test dataset will be used at the end of each epoch to evaluate the accuracy of the trained classification models. We show the evolution of the model accuracy over data batch arrivals until the performance of \system converges.


{\bf Label model.} We use a multilayer perceptron to mainly assess the quality of each label, it will also be used to join the ensemble prediction. We fix the model initialization across experiment runs. For IoT and Cluster dataset, the neural network consists of two layers with 28 neurons each. The precision and robustness of the label model are critical to filter out the noisy labels and provide a clean training set to the classification model. We considered different models. Neural networks provided the best results in terms of accuracy and stability over time. Adaboost gave excellent accuracy when training from the initial data with ground truth, but it is too sensitive to label noise. Random forest is also known to be robust against label noise~\cite{Frenay:TNNLS14:survey}, however its accuracy was below the neural network one.


{\bf Classification model.} We use KNN to jointly do ensemble prediction with label model. For the extensions \systemVoting and \systemActive, classification model will also play a role as label model to assess the quality of the data. We set the number of neighbours to five in KNN.
Higher values can increase the resilience of the algorithm to residual noise, but also induce extra computational cost. The current choice stems from good results in preliminary experiments.


{\bf Slimmed framework.} For the face recognition task we use \systemSlimmed. In this case we use a 110-layers ResNet~\cite{He2015DeepRL} as classification model. ResNet is a type of CNN architecture which introduces residual functions to alleviate the vanishing gradient problem in training deep neural networks improving the classification performance and model convergence. We use a fixed model initialization across experiment runs.

{\bf Baselines.} The proposed \system is compared against following baseline data selection schemes: 1) {\em No-Sel}, where all data instances of arriving batches are used for training the classification model; 2) {\em Opt-Sel} which emulates an omniscient agent who can perfectly distinguish between clean and noisy labels, and only use clean data to train the models; and, 3) {\em IDS}: the intrusion detection system from~\cite{Agarwal:2016}. The main idea and structure of {\em IDS} are similar to the proposed \system. The differences are: i) IDS only trains label quality model with $\mathcal{D}_0$ once without continuous updated; and, ii) IDS only uses classification model for predictions, instead of combining prediction results of quality and classification models.
In addition, we consider: 4) {\em Full-Clean} which simulates perfectly recovered labels, i.e., all wrong labels have been correctly identified and recovered by, e.g., an oracle. This represents the ideal solution which provides all clean data in each data batch. In the following, model names ending in `\_C' means the predictions are obtained from the anomaly classification model, ending in `\_L' means the predictions are obtained from the label quality model, and ending in '\_Ens' means the predictions are obtained from both anomaly classification and label quality model specified in Algorithm~\ref{alg:prediction_technique}. {\em No-Sel}, {\em Opt-Sel} and {\em Full-Clean} use all the data to independently train the label quality and anomaly classification model. And  Algorithm~\ref{alg:prediction_technique} is used to generate the final prediction. There is no filtering process in these cases.

To compare with \systemSlimmed on image dataset, we introduce two state-of-the-art approaches: 1) Forward~\cite{patrini2017making} estimates the noise transition matrix before training the model, and subsequently uses this transition matrix for loss correction;
and 2) Co-Teaching~\cite{coteaching} trains two deep neural networks simultaneously to let them teach each other. For Forward, we use the same network architecture as for \systemSlimmed, i.e. 110-layers ResnNet. As Co-Teaching trains two models, we use two 56-layers ResNet. To speed up model convergence for \systemSlimmed, \systemSlimmedLimit, and Forward, we implement the E (Exponential)/PD (Proportional-Derivative)-Control~\cite{epd} and Event-Based Control Learning rate~\cite{eblrc} as learning rate schedule based on stochastic gradient descent (SGD) optimizer. Co-Teaching has its own learning rate scheduler.

\subsection{Handling Dynamic Data}
\label{ssec:EvalDynamicData}

\begin{figure}[tb]
	\begin{center}
		\subfloat[IoT thermostat device attacks]{
			\includegraphics[width=0.47\columnwidth]{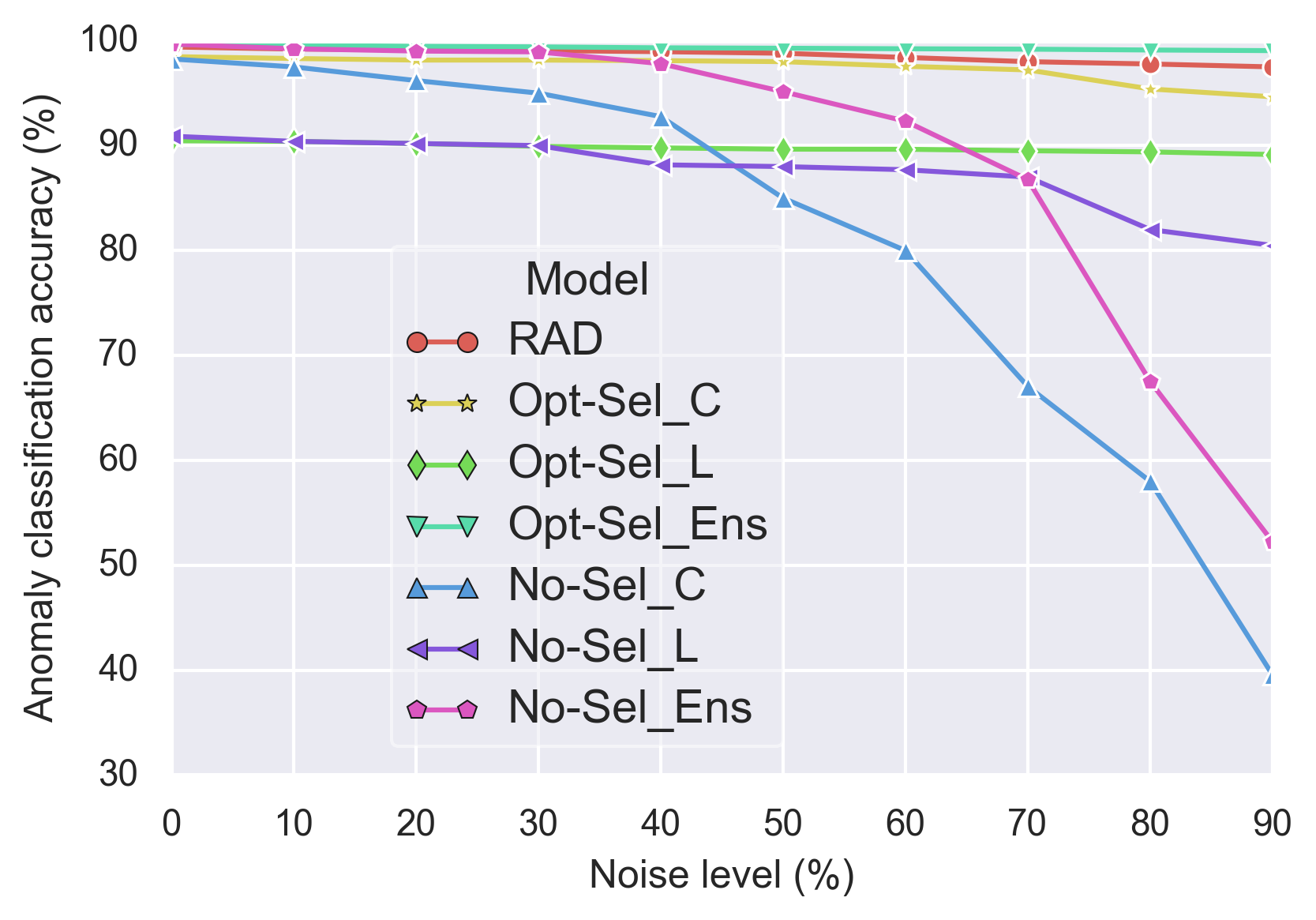}
			\label{fig:eval-noise-IoT}
		}
		\hfil
		\subfloat[Cluster task failures]{
			\includegraphics[width=0.47\columnwidth]{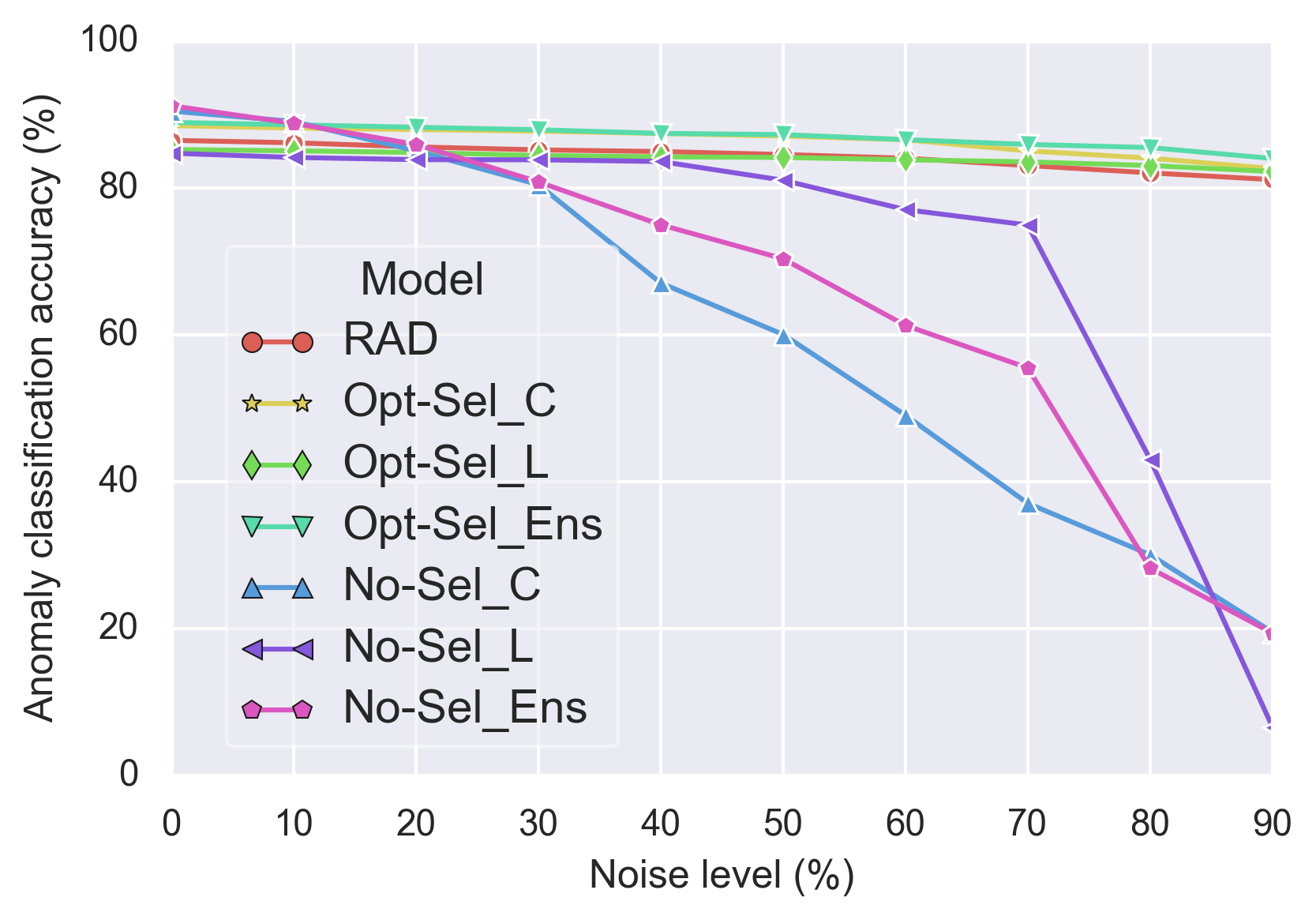}
			\label{fig:eval-noise-ClusterTasks}
		}
		\caption{Impact of data noises on \system accuracy}
		\vspace{-1.5em}
		\label{fig:NoiseLevel}
	\end{center}
\end{figure}

Fig.~\ref{fig:EvolOverTime-IoT} and \ref{fig:EvolOverTime-ClusterTasks} show the evolution of the mean and variance of the classification accuracy achieved by \system  on the thermostat and task failure datasets, respectively.
Each figure moreover presents results under two levels of label noise: 30\% and 40\%. We compare \system against no selection ({\em No-Sel}), optimal selection ({\em Opt-Sel}) and {\em IDS}. 
One can notice that learning from all data instances without cleansing (i.e., {\em No-Sel} curves) gives consistently lower accuracy in all cases. For the task failure dataset, the accuracy even oscillates and diverges. The performance of \system is better: (1) the accuracy does not diverge and (2) the accuracy consistently increases until it converges.
The end accuracies, under 30\% and 40\% noise level, are all around 99\% and 85\% for the IoT attack and cluster tasks datasets, respectively. 
For the first dataset, the accuracy of \system follows closely the ensemble prediction accuracy of {\em Opt-Sel}. 
As for the second dataset, \system follows ensemble prediction of {\em Opt-Sel} at first but then converges after 30 data batch arrivals.
Note that \system gives also more stable results as shown by shorter variance bars which in magnitude are in line with the ones obtained by an ideal data cleansing. For {\em No-Sel} the bars are significantly larger. 

We note that ensemble prediction can greatly enhance the learning outcomes in the presence of noisy data, compared the  prediction of solely the label quality  or classification model. Such an observation holds for the different data selection schemes discussed in the subsequent sections. Due to space limits, we skip the presentation of those results.


In summary: (i) continual learning is advantageous compared to using only the initial dataset; however, (ii) continual learning exposes to possible classification accuracy degradation stemming from noisy labels if proper data selection is lacking, (iii) \system improves the classification accuracy compared to taking all labels, (iv) the data selection of \system is good, and close to being optimal in some cases, and (v) ensemble prediction can greatly enhance the robustness against noisy data.

\subsection{Evaluation of Noise Robustness of \system}
\label{ssec:EvalNoiseImpact}

Next we investigate the impact of different noise levels on the \system performance in terms of classification accuracy.

Fig.~\ref{fig:eval-noise-IoT} and~\ref{fig:eval-noise-ClusterTasks} present the classification accuracy for various levels of noise, ranging from 0\% (all data are clean) up to 90\% for our two main reference datasets: IoT thermostat device attacks and Cluster task failures. All experiment settings remain the same as before, only the noise level of training data batches varies. Once again, the \system performance is compared to learning from all data ({\em No-Sel}) and an omniscient data cleanser ({\em Opt-Sel}).

As illustrated in Section~\ref{sec:ProblemStatement}, for {\em No-Sel} the noisier the data are, the worse the classification accuracy, with ensemble prediction, dropping to 20\% and 52\% for the Cluster and IoT datasets, respectively.
A decreasing trend can also be found for \system and {\em Opt-Sel}, however the drops are significantly smaller: at most 5\%. As there is by definition no noise in {\em Opt-Sel} case, the decrease in classification accuracy is only due to the reduction of the overall amount of clean data to learn from. Since the data cleansing of \system is not perfect, the accuracy reduction is caused by noise pollution and overall clean data reduction. Nevertheless, the impact is small and any huge accuracy pitfall is avoided which results in \system's performance being close to {\em Opt-Sel}. We can conclude that \system can limit the impact of the amount of noise across a wide range of noise levels.

\subsection{Analysis of All Datasets}
\label{ssec:SystemGenericity}

Summary results are reported in Table~\ref{tab:rad_results}. One can see that results of \system are always better than IDS and {\em No-Sel}. For IoT dataset with 40\% noise, RAD is even better than any single model of {\em Opt-Sel}. But there is still room for improvement between \system and {\em Opt-Sel\_Ens}.  F1 scores are consistent with accuracy results with few exceptions.  Under the cluster dataset and 30\% noise, F1 score of {\em No-Sel\_Ens} is better than {\em No-Sel\_L} and {\em No-Sel\_C} even if the accuracy is slightly worse than {\em No-Sel\_L}. Under 40\% noise, the accuracy of {\em No-Sel\_Ens} is $8.43$ points worse than {\em No-Sel\_L}, however the difference in F1 score is only $0.01$. This means that the accuracy difference is mostly due to data unbalance.

\begin{table}[t]
	\begin{center}
		\caption{Final accuracy of all algorithms for the Cluster task failures and IoT device attacks datasets on 30\% and 40\% noise level. Final F1-score is reported in brackets for the Cluster dataset. All results are averaged across 3 runs.}
		\label{tab:rad_results}
		\begin{tabular}{L{1.8cm} C{1.4cm} C{1.1cm} C{1.4cm} C{1.1cm}}
			\toprule
			Algorithm  & Cluster (30\%) & IoT(30\%) & Cluster (40\%) & IoT(40\%) \\
			\midrule
			Full-Clean\_C & 89.35(0.90)  & 98.28 & 89.35(0.90)  & 98.28  \\
			Full-Clean\_L & 85.17(0.84)  & 90.87 & 85.17(0.84)  & 90.87\\
			Full-Clean\_Ens & 91.08(0.91)  & 99.83 & 91.08(0.91)  & 99.83 \\
			Opt-Sel\_C &  87.68(0.87) & 98.08 &  87.16(0.87) & 98.06\\
			Opt-Sel\_L & 84.37(0.82)  & 90.81 &  84.18(0.83) & 89.70\\
			Opt-Sel\_Ens & 87.88(0.87)  & 99.35 &  87.60(0.87) & 99.25 \\
			No-Sel\_C &  77.40(0.79) & 95.47 &  71.02(0.74) & 92.27\\
			No-Sel\_L &  83.54(0.82) & 89.95 &  83.35(0.80) & 89.57\\
			No-Sel\_Ens & 81.53(0.83)  & 98.06 &  74.92(0.79) & 97.51 \\
			\midrule
			\system    & 85.46(0.84)  & 99.01 & 85.03(0.83)  & 98.95\\
 			\systemIDS  &  83.63(0.81) & 97.83 & 83.31(0.81)  & 97.23\\
			\midrule
			\systemVoting    & 86.01(0.85)  & 99.21 & 85.73(0.84)  & 99.07\\
			$\mbox{RAD-AL}^1$   & 90.84(0.90)  & 99.72 & 90.77(0.90)  & 99.58\\
			\midrule
			$\mbox{RAD-AL-L}^2$   &90.00(0.90)& 99.68 & - & -\\
 			$\mbox{PSO}^3$  &87.83(0.87)& 98.85 & - & - \\
			\bottomrule
			\multicolumn{5}{l}{\footnotesize 1. RAD-AL: RAD Active Learning} \\
			\multicolumn{5}{l}{\footnotesize 2. RAD-AL-L: RAD Active Learning Limited } \\
 			\multicolumn{5}{l}{\footnotesize 3. PSO: Pre-Select Oracle } \\
		\end{tabular}
	\end{center}
	\vspace{-1.5em}
\end{table}



The resilience to high levels of noise might be even more important than the benefits of continual learning. Under such levels, the classification accuracy without data cleansing diverges for all datasets. Even if it is rare to have noise levels of 90\% or above, they might still happen for short time periods in case of attacks to the auto-labelling system, e.g., via flooding of malicious labels. Hence this property can be crucial for the dependability of the auto-labelling system.

\subsection{Limitation of RAD Framework}
\label{ssec:Limitation}
Though \system works well for datasets of Cluster task failures and IoT device attacks. We can still see the potential limitations of this framework. For example: 1) the assumption of availability of a small fraction of clean data which may not be possible; 2) if data is coming at high rates, 
training two models simultaneously instead of one can slow down the system;
3) as the anomaly classifier receives only the data selected by the label model, there is a risk that the classifier model overfits to label model. To address these issues we devised the two extensions presented in Sec.~\ref{ssec:extensions}. These are evaluated in the next subsections.

\subsection{\systemVoting and History Extension}
\label{ssec:SystemVoting}
\begin{figure*}[t]
	\centering
	\subfloat[Iot data with noise level of 30\%]{
		\includegraphics[width=0.99\columnwidth]{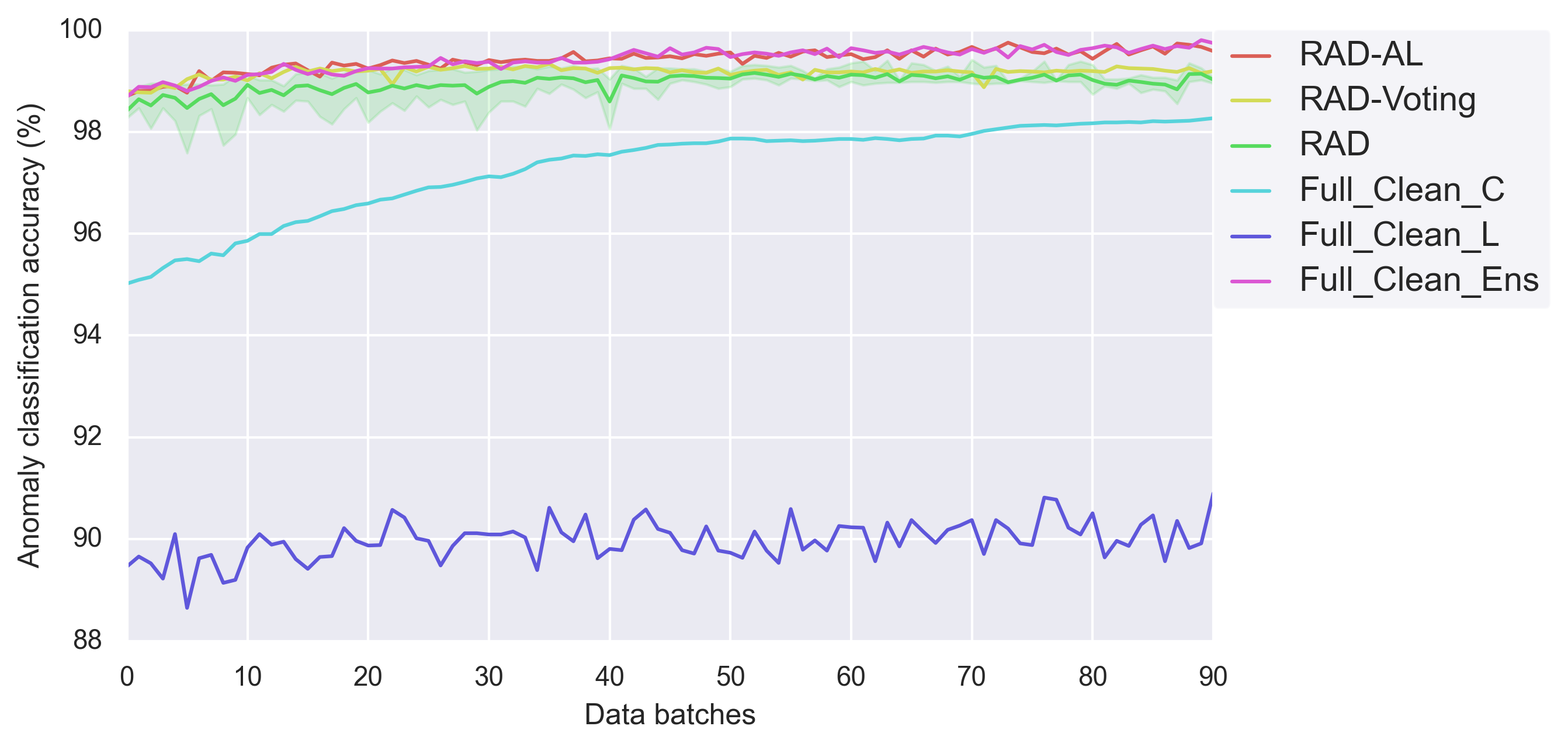}
		\label{fig:EvolOverTime-IoT-Voting-30}
	}
	\hfil
	\subfloat[Iot data with noise level of 40\%]{
		\includegraphics[width=0.99\columnwidth]{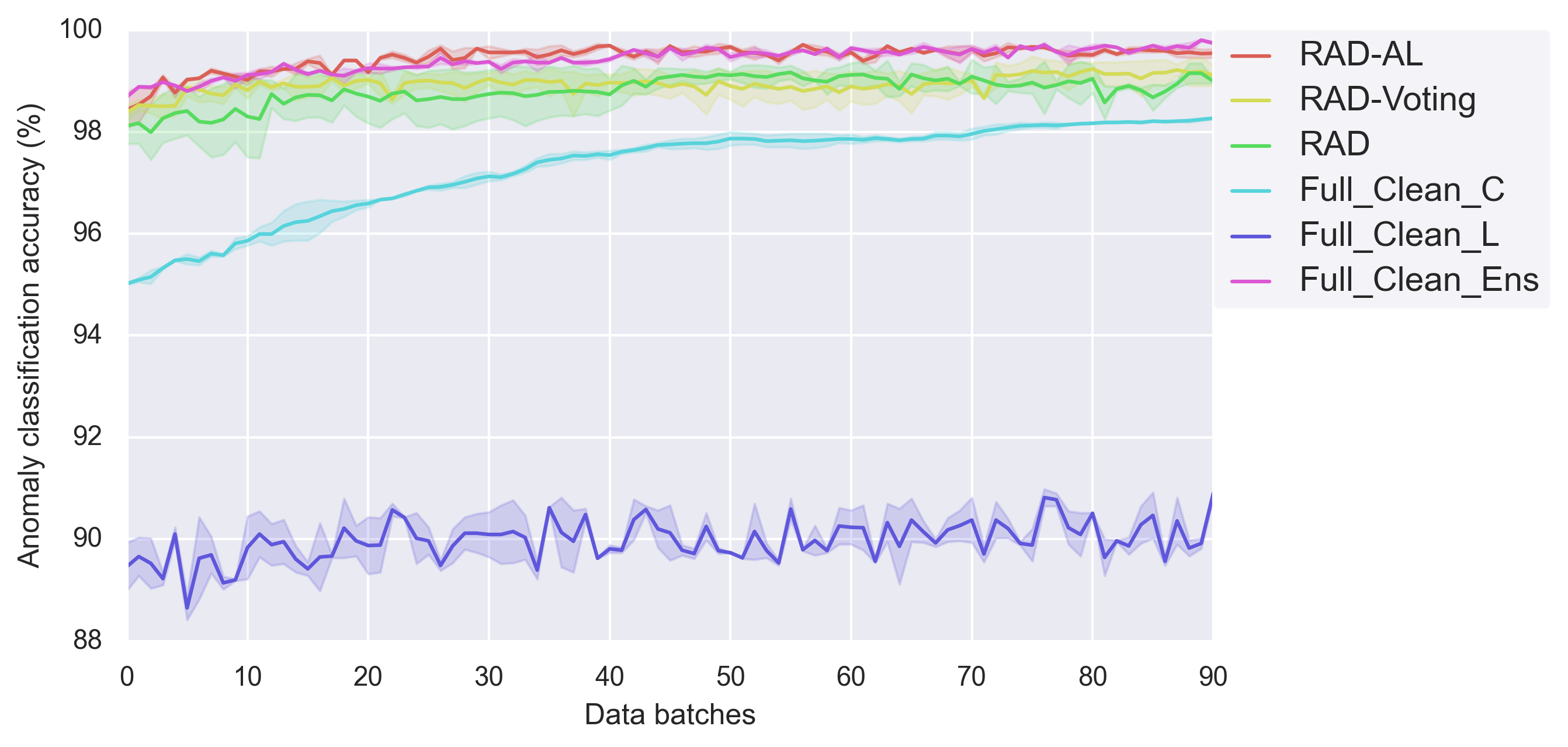}
		\label{fig:EvolOverTime-IoT-Voting-40}
	}
	\caption{Evolution of learning over time -- Use case of IoT thermostat device attacks with \systemVoting and \systemActive (RAD-AL). Full\_clean means that no label noise is injected.}
	\label{fig:EvolOverTime-IoT-voting}
	\vspace{-1.5 em}
\end{figure*}

\begin{figure*}[!t]
	\centering
	\subfloat[Cluster data with noise level of 30\%]{
		\includegraphics[width=0.99\columnwidth]{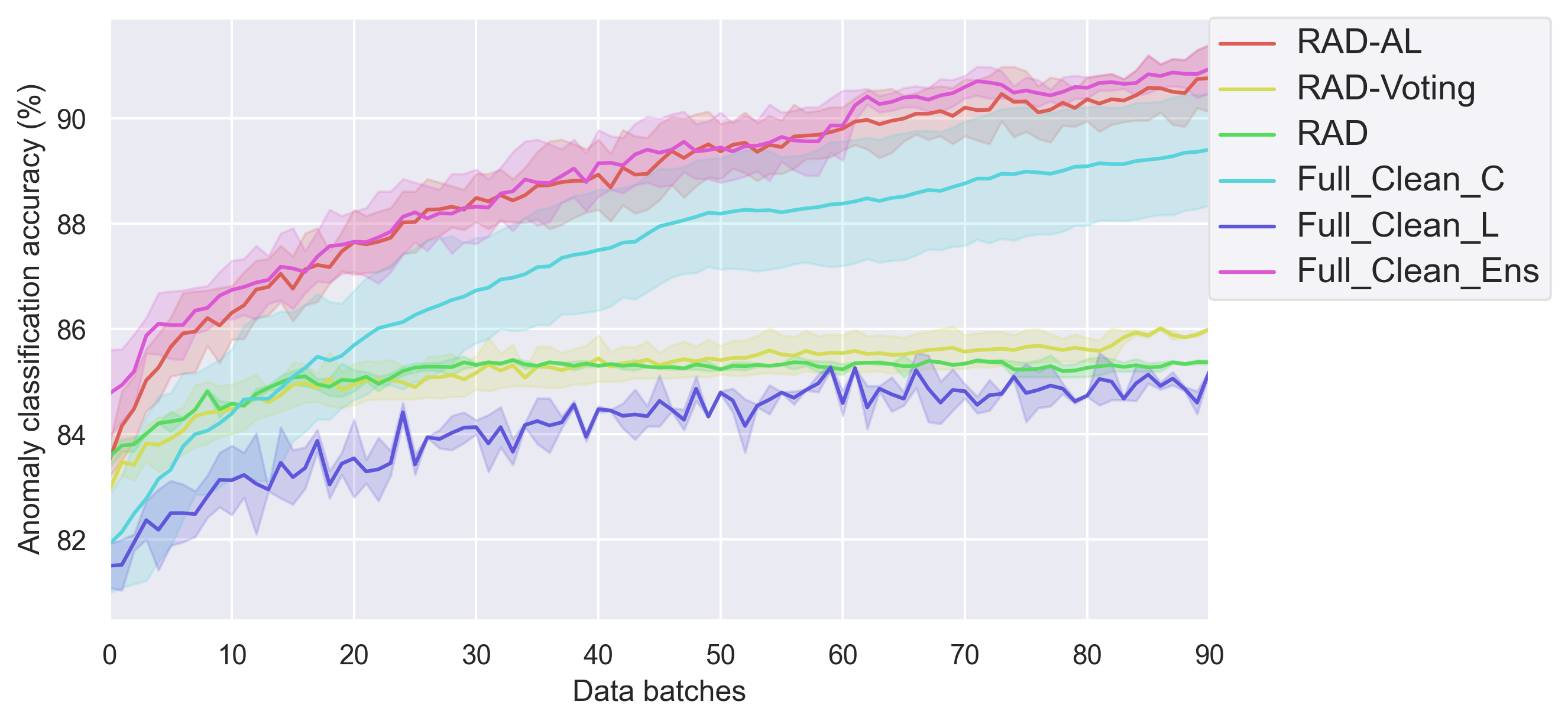}
		\label{fig:EvolOverTime-Cluster-Voting-30}
	}
	\hfil
	\subfloat[Cluster data with noise level of 40\%]{
		\includegraphics[width=0.99\columnwidth]{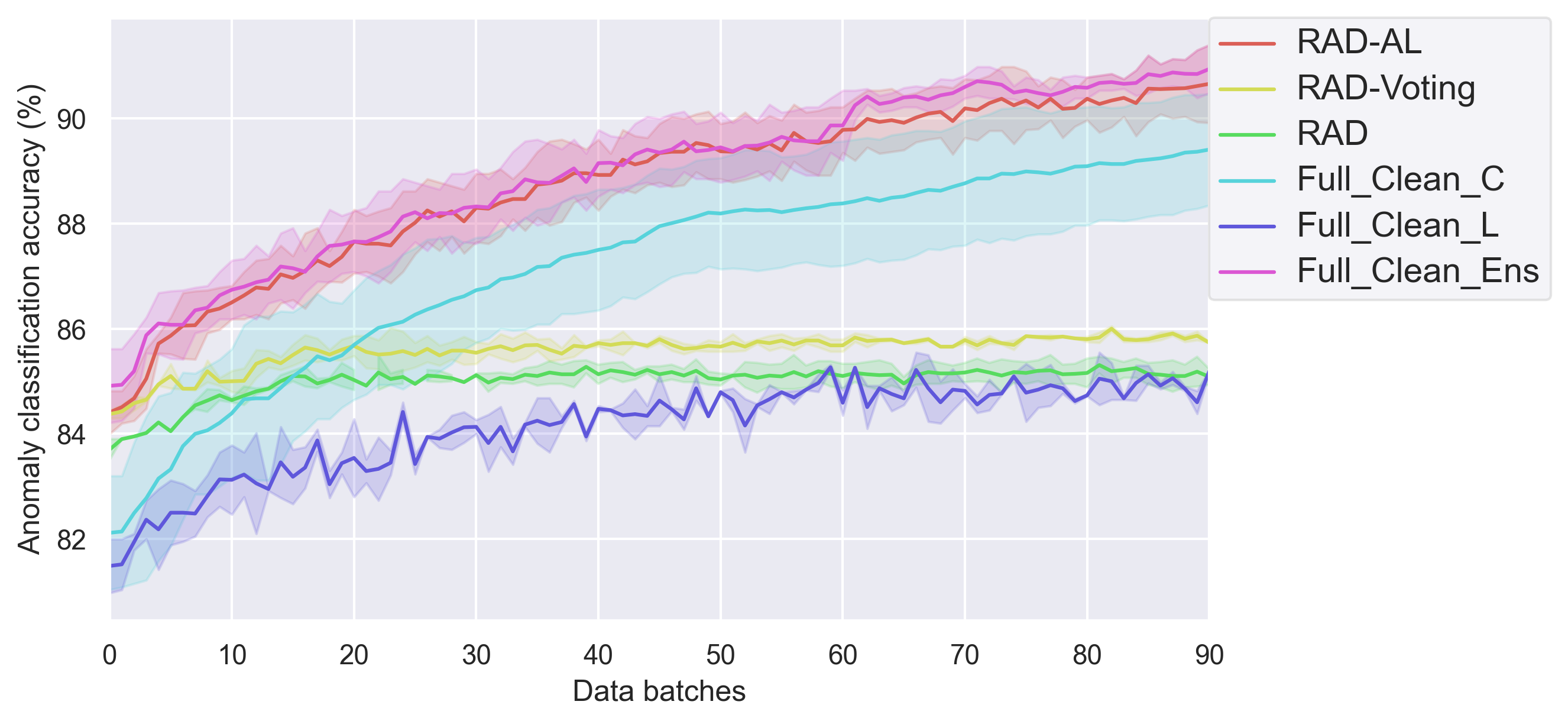}
		\label{fig:EvolOverTime-Cluster-Voting-40}
	}
	\caption{Evolution of learning over time -- Use case of Cluster task failures with \systemVoting and \systemActive}
	\label{fig:EvolOverTime-Cluster-voting}
	\vspace{-1.5em}
\end{figure*}

\begin{figure}[!ht]
	\centering
	\subfloat[IoT data with 30\% noise]{
		\includegraphics[width=0.47\columnwidth]{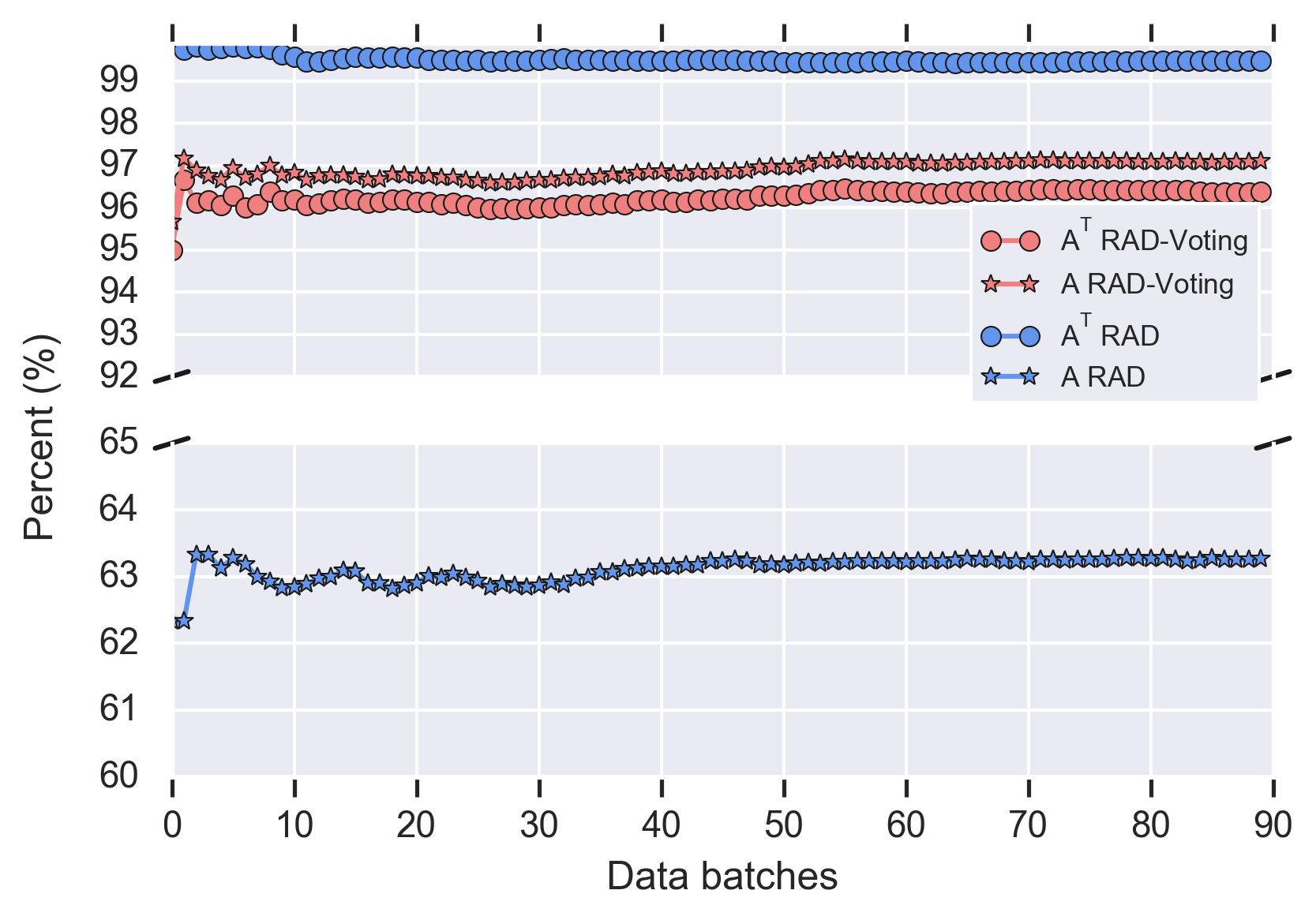}
		\label{fig:EvolOverTime-True-Active-IoT}
	}
	\hfil
	\subfloat[Cluster data with 30\% noise]{
		\includegraphics[width=0.47\columnwidth]{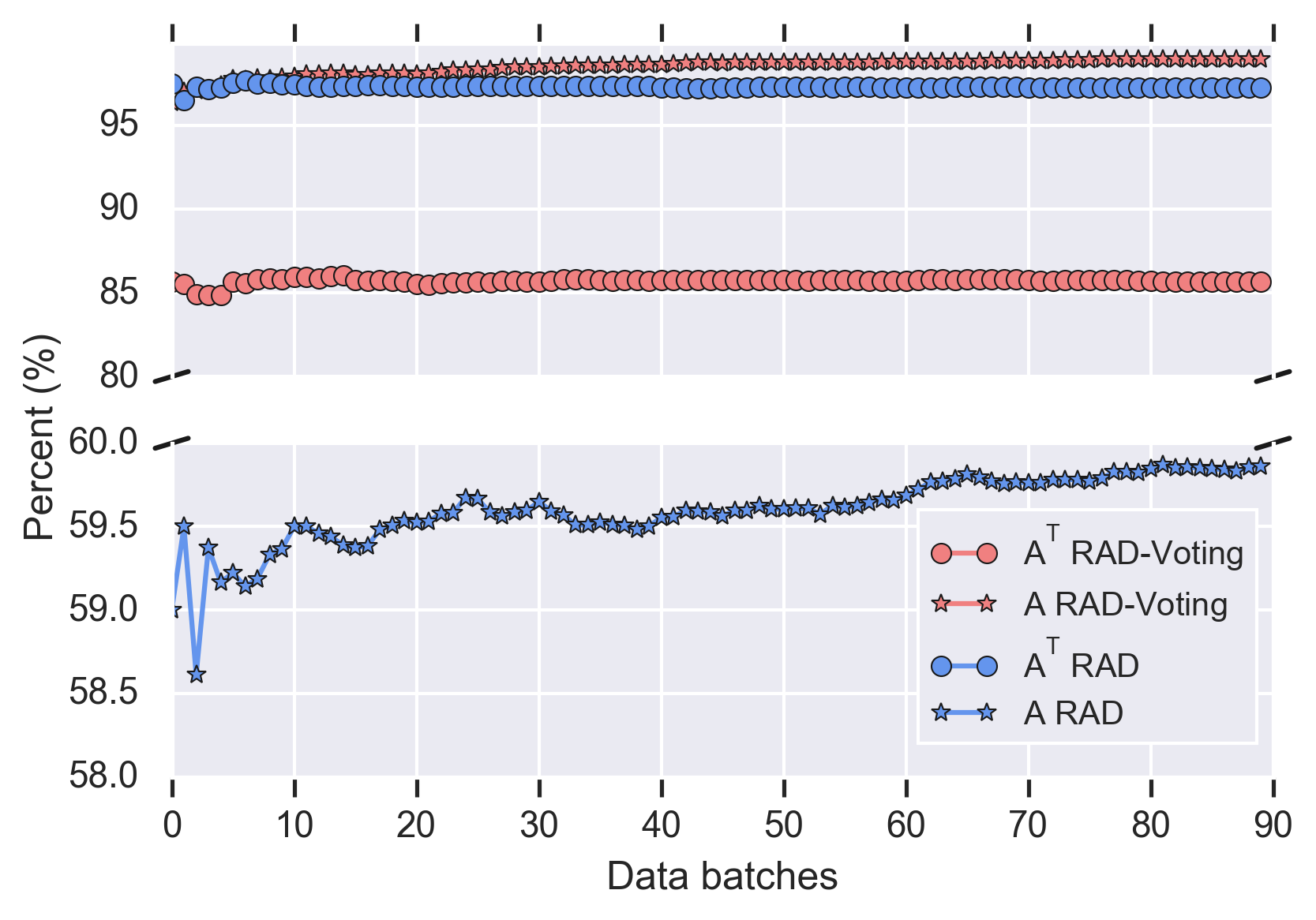}
		\label{fig:EvolOverTime-True-Active-Cluster}
	}
	\caption{\system and \systemVoting: percentage of Hot data and How-Truth it is.}
	\label{fig:active-true-active}
	\vspace{-1.5em}
\end{figure}

\begin{figure*}[!ht]
	\centering
	\subfloat[Iot data with noise level of 30\%]{
		\includegraphics[width=0.83\columnwidth]{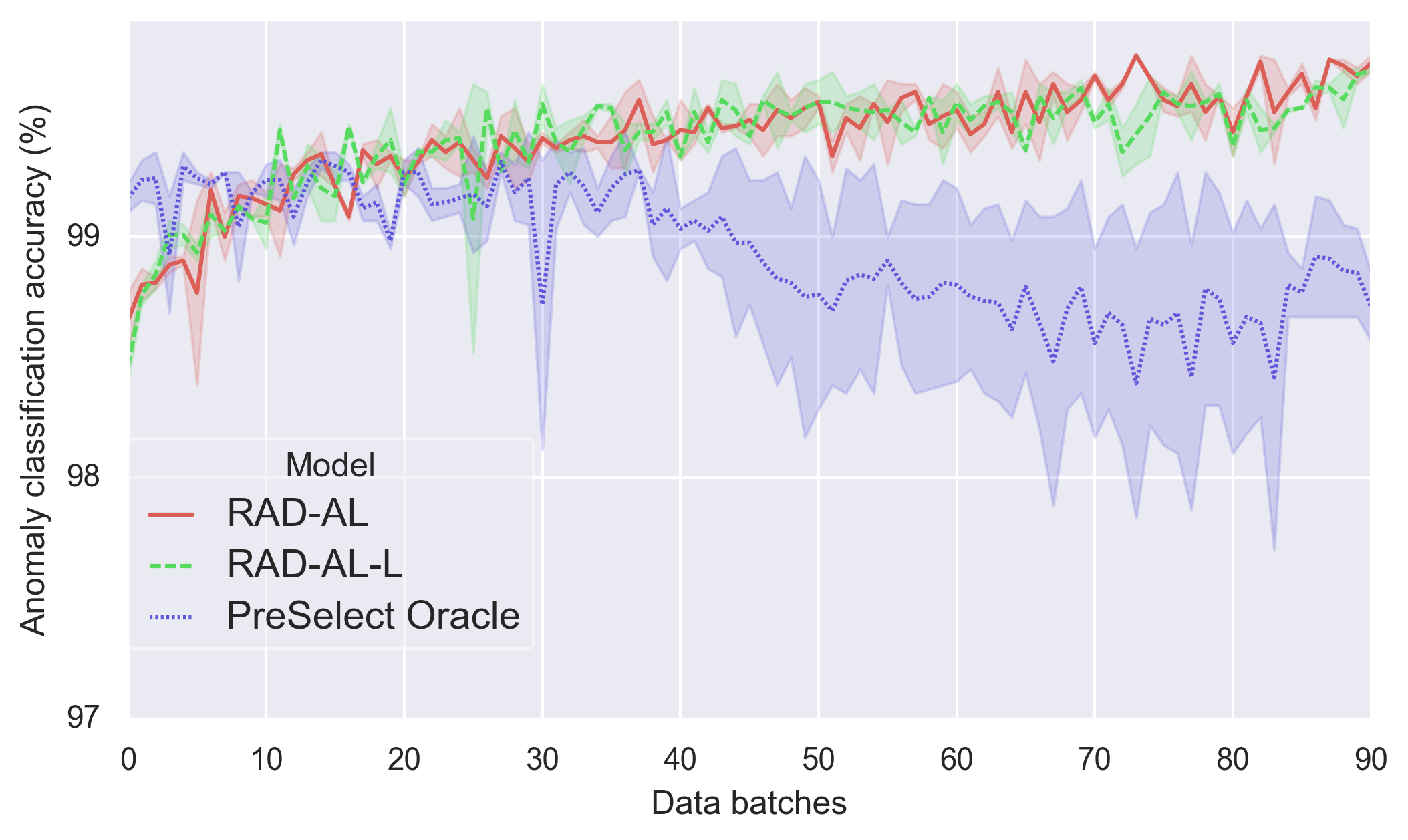}
		\label{fig:preselection-thermo}
	}
	\hfil
	\subfloat[Cluster data with noise level of 30\%]{
		\includegraphics[width=0.83\columnwidth]{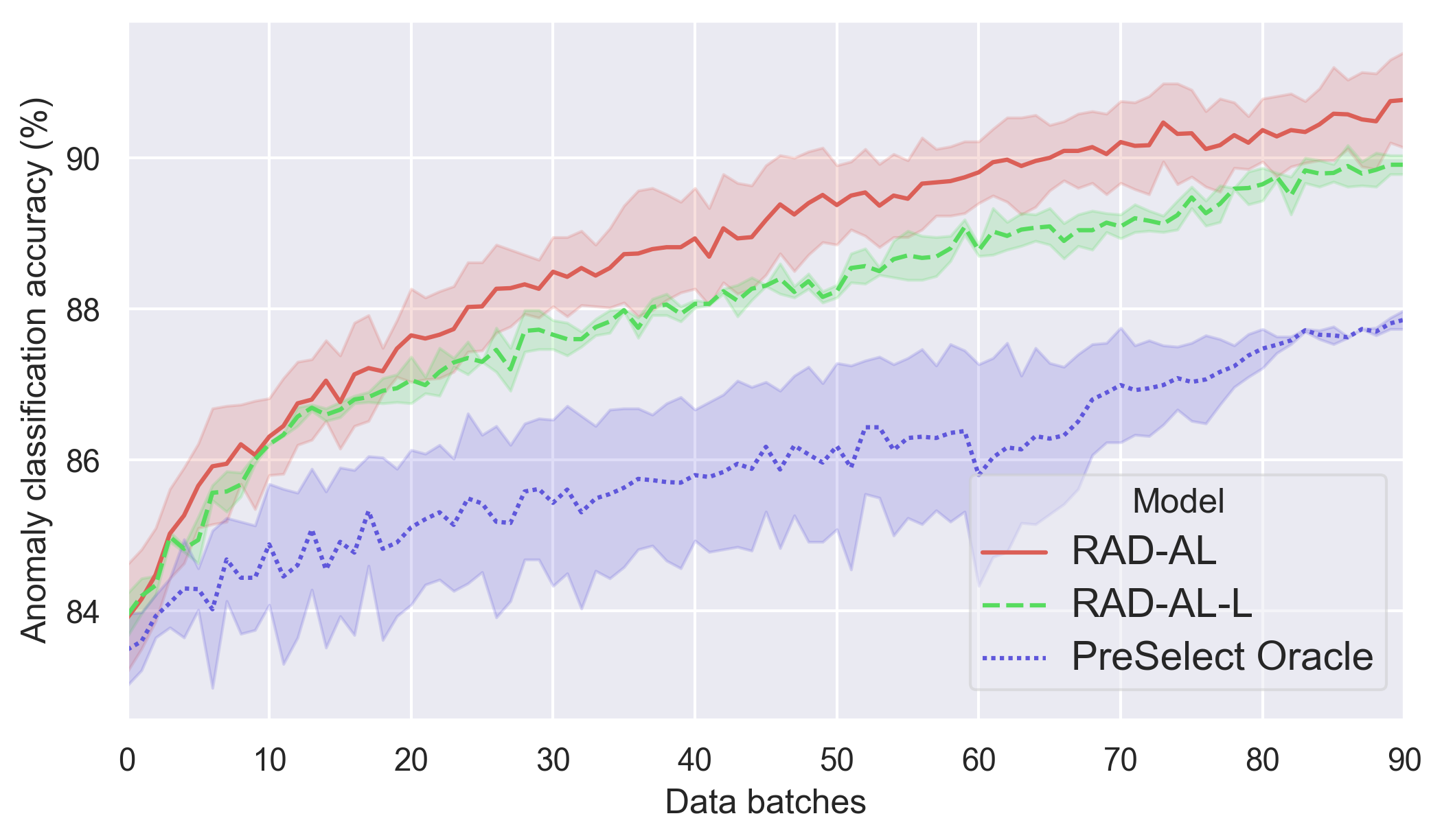}
		\label{fig:preselection-task}
	}
	\caption{Comparison of \systemActiveLimit (RAD-AL-L) and Pre-Select Oracle, showing the power of selection.}
	\label{fig:oracle-preselection}
	\vspace{-1.0em}
\end{figure*}

In the first extension we let both the label and classifier models vote on the label quality and include the possibility to recover instances from history to be evaluated as the model performances improve over time.

We evaluate the accuracy of \systemVoting over time and different noise levels in Fig.~\ref{fig:EvolOverTime-IoT-voting} and Fig.~\ref{fig:EvolOverTime-Cluster-voting} for the IoT thermostat and Cluster task failures, respectively. For the IoT dataset, \systemVoting is better than any single model of {\em Full-Clean}. For the Cluster dataset, \systemVoting does not converge as \system. 
Table.~\ref{tab:rad_results} summarizes and compares the \systemVoting performance with others.
We can see that \systemVoting performance is always better than \system. This is because we correct labels in the \systemVoting algorithm, which increases the number of training instances over \system. 
The F1-score results of the Cluster dataset are in line with the accuracy results.

To better understand the different performance between the two datasets we define $A$ (called Hot) as the percent of data used for training till time $t_i$:
\begin{align}
A = \frac{\sum_{k=1}^{i} (|\mathcal{D}_{k}^{*}| + |\mathcal{U}_k^{*}|)}{\sum_{k=1}^{i}|\mathcal{D}_k|}.
\end{align}
Knowing the number of true clean labels used per batch $C_i^T$, we further define $A^T$ (called Hot-Truth) as the percent of true clean active data.
\begin{align}
A^{T} = \frac{\sum_{k=1}^{i} C_k^T}{\sum_{k=1}^{i}  (|\mathcal{D}_{k}^{*}| + |\mathcal{U}_k^{*}|)}
\end{align}
In both formulas, we exclude the initial clean batch $\mathcal{D}_0$. Intuitively, $A$ tells how much of the incoming data we use for training, and $A^{T}$ how clean the used training data is.

Fig.~\ref{fig:EvolOverTime-True-Active-IoT} and \ref{fig:EvolOverTime-True-Active-Cluster} plot $A$ and $A^{T}$ using RAD and RAD Voting over time for the IoT and Cluster datasets, respectively. 
For the IoT dataset both $A$ and $A^T$ improve over time with RAD Voting, see Fig.~\ref{fig:EvolOverTime-True-Active-IoT}. This means that both the quantity of active data, i.e. $A$, and the quality, i.e. cleanliness, of the active data $A^T$ improve over time. For  Cluster dataset, $A^T$ of RAD Voting does not improve over time even though $A$ increases, see Fig.~\ref{fig:EvolOverTime-True-Active-Cluster}. We attribute this to the fact that both $\mathfrak{C}$ and $\mathcal{L}$ predict the same wrong class and this class is used to replace the original label of the data instance. 
Next we compare the performance of RAD and RAD Voting. 
One can see that RAD Voting includes more data into the training set (higher $A$) than RAD, but the data is less clean (lower $A^T$). Overall since RAD Voting filters out less training data (final $A$ of RAD Voting is 27.74\% and 39.07\% higher than RAD for the IoT and Cluster experiments, respectively), the difference in $A^T$ is relatively small (final $A^T$ of RAD Voting is only 3.11\% and 11.56\% smaller than RAD for the IoT and Cluster experiments, respectively). $A^T \cdot A$ reflects the ratio between clean label data used in training and the total received data. 
Both RAD and RAD-voting receive the same amount of data in every epoch. 
Therefore the higher $A^T \cdot A$ is, the more clean label data is used in training. For both use cases, the final $A^T \cdot A$ of RAD Voting is higher than RAD. Intuitively RAD Voting should be better than RAD and experiment results are in line with this intuition.

\subsection{\systemActive}
\label{subsec:rad_active_learning}

\systemActive extends \system with the ability of asking an oracle to provide the true label for data instances where the two models disagree. First we consider \systemActive with no limits on the number of oracle requests followed by \systemActiveLimit which limits the number of oracle interactions.



Fig.~\ref{fig:EvolOverTime-IoT-voting} and Fig.~\ref{fig:EvolOverTime-Cluster-voting} show the performance of \systemActive (RAD-AL) for the IoT and Cluster datasets under 30\% and 40\% noise, respectively. The figures compare \systemActive to \systemVoting, {\em No-Sel} and {\em Full-Clean}.  We can see that \systemActive is always better than \systemVoting and almost as good as {\em Full-Clean\_Ens} across the two different datasets and different noise levels. From the results in Table~\ref{tab:rad_results}, we can observe that the result of \systemActive is extremely close to {\em Full-Clean\_Ens} who is the best in every column. That shows that our training data selection in \systemActive is very accurate. Almost all noisy data are filtered out for consultations to expert.

Consulting every single uncertain data instance with expert might be too expensive or impossible in practice. Hence, we consider \systemActiveLimit which limits the consultations with experts. Here we limit the number of queries per batch  to 20\% of the batch size. To illustrate the power of our training data selection process, we introduce a new comparison: \systemPreselect. \systemPreselect has the same number of consultation to oracles as \systemActiveLimit, but data instances are selected randomly before training. Fig.~\ref{fig:oracle-preselection} shows the results for IoT and Cluster datasets, one can notice that the curve of \systemActiveLimit (RAD-AL-L) increases along with \systemActive and largely outperforms \systemPreselect. The accuracy difference here is due to the uncertainty ranking used by the \highestActiveLimit. From the result in Table~\ref{tab:rad_results}, one can see that after imposing a query limit of 20\%,  \systemActiveLimit reaches similar accuracy as  \systemActive, and higher accuracy than \system and \systemVoting.


\subsection{Impact of Initialization}
\label{ssec:ImpactInitialization}

Here we study the impact on \system and its extensions of 
the size of the initial dataset $\mathcal{D}_0$.
We vary the number of initial clean data instances from 100 to 6000, and measure the classification accuracy after 90 data batch arrivals. We consider the {\em Opt-Sel} baseline since the {\em No-Sel} baseline is meant for the framework configuration, not its performance evaluation.

\begin{figure}[htb]
   \vspace{-1.0em}
	\begin{center}
		\subfloat[IoT thermostat device attacks]{
			\includegraphics[width=0.47\columnwidth]{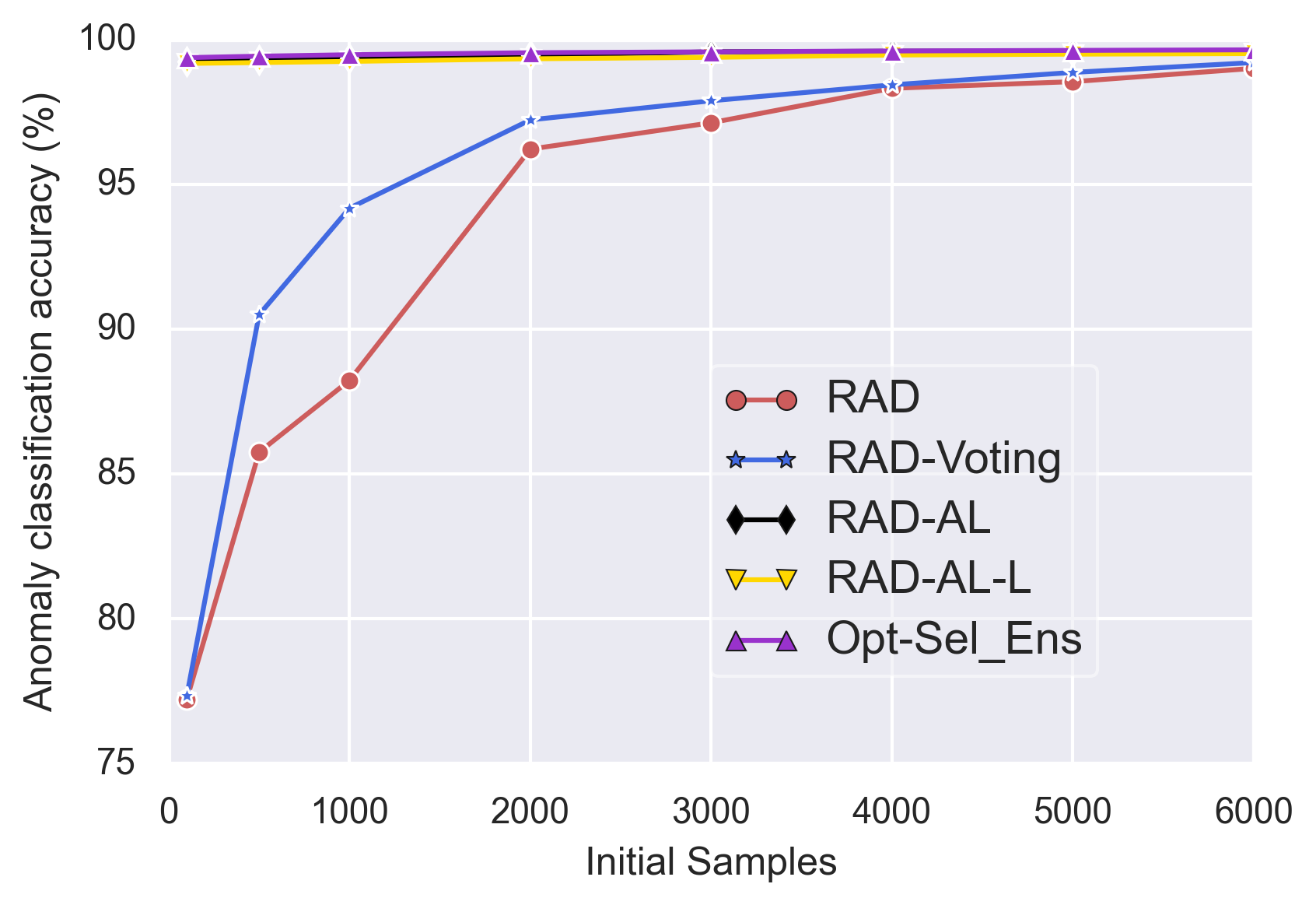}
			\label{fig:eval-init-data-IoT}
		}
		\hfil
		\subfloat[Cluster task failures]{
			\includegraphics[width=0.47\columnwidth]{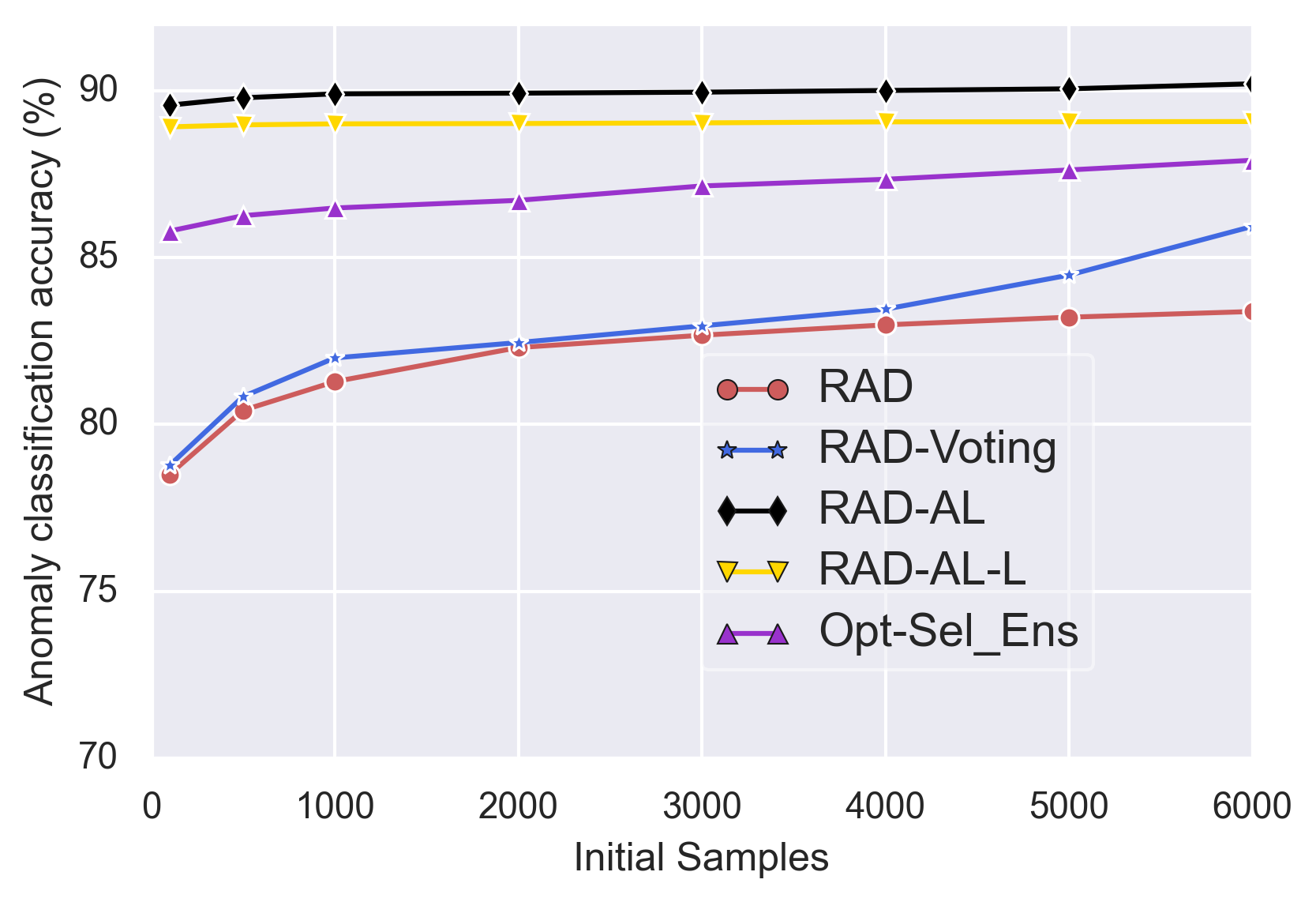}
			\label{fig:eval-init-data-ClusterTasks}
		}
		\caption{Impact of size of initial data batch $\mathcal{D}_0$ on \system accuracy with 30\% noise level.}
		\label{fig:IntialCluster}
	\end{center}
	\vspace{-2.0em}
\end{figure}

Fig.~\ref{fig:eval-init-data-IoT} and Fig.~\ref{fig:eval-init-data-ClusterTasks} show the results for the IoT and Cluster datasets, respectively. 
{\em Opt-Sel\_Ens} seems to perform independent from the number of initial data instances ($|\mathcal{D}_0|$) in Fig.~\ref{fig:eval-init-data-IoT}. This is due to the fact that after 90 batch arrivals the amount of training data is sufficient for the accuracy to converge. In Fig.~\ref{fig:eval-init-data-ClusterTasks} however, we can see that $|D_0|$ influences the accuracy of {\em Opt-Sel\_Ens}. The model is yet to converge at the end of learning, but the influence is clearly smaller than that for \system and \systemVoting. For these two models the size of $\mathcal{D}_0$ matters more: the larger the better. At $|\mathcal{D}_0|=2000$ their performances are similar to {\em Opt-Sel\_Ens} (less than 5\% difference) for IoT dataset, and at $|\mathcal{D}_0|=6000$ they almost overlap. \systemVoting outperforms \system  in both datasets under all sizes of $\mathcal{D}_0$. This is because \systemVoting can correct data labels and thus increase the number of training instances. Finally, \systemActive and \systemActiveLimit (20\% limit) do not depend on the size of $\mathcal{D}_0$, since they can ask the oracle for the label of uncertain data instances.

This justifies our earlier choice of $\mathcal{D}_0$ having 6000 data instances as it enables to achieve the best accuracy. However, all proposed frameworks could also perform well with only half the initial data instances in $\mathcal{D}_0$.

\subsection{\systemSlimmed on Image Data}
\label{ssec:systemSlimmed}

\begin{figure}[t]
    \centering
    \includegraphics[width=0.83\linewidth]{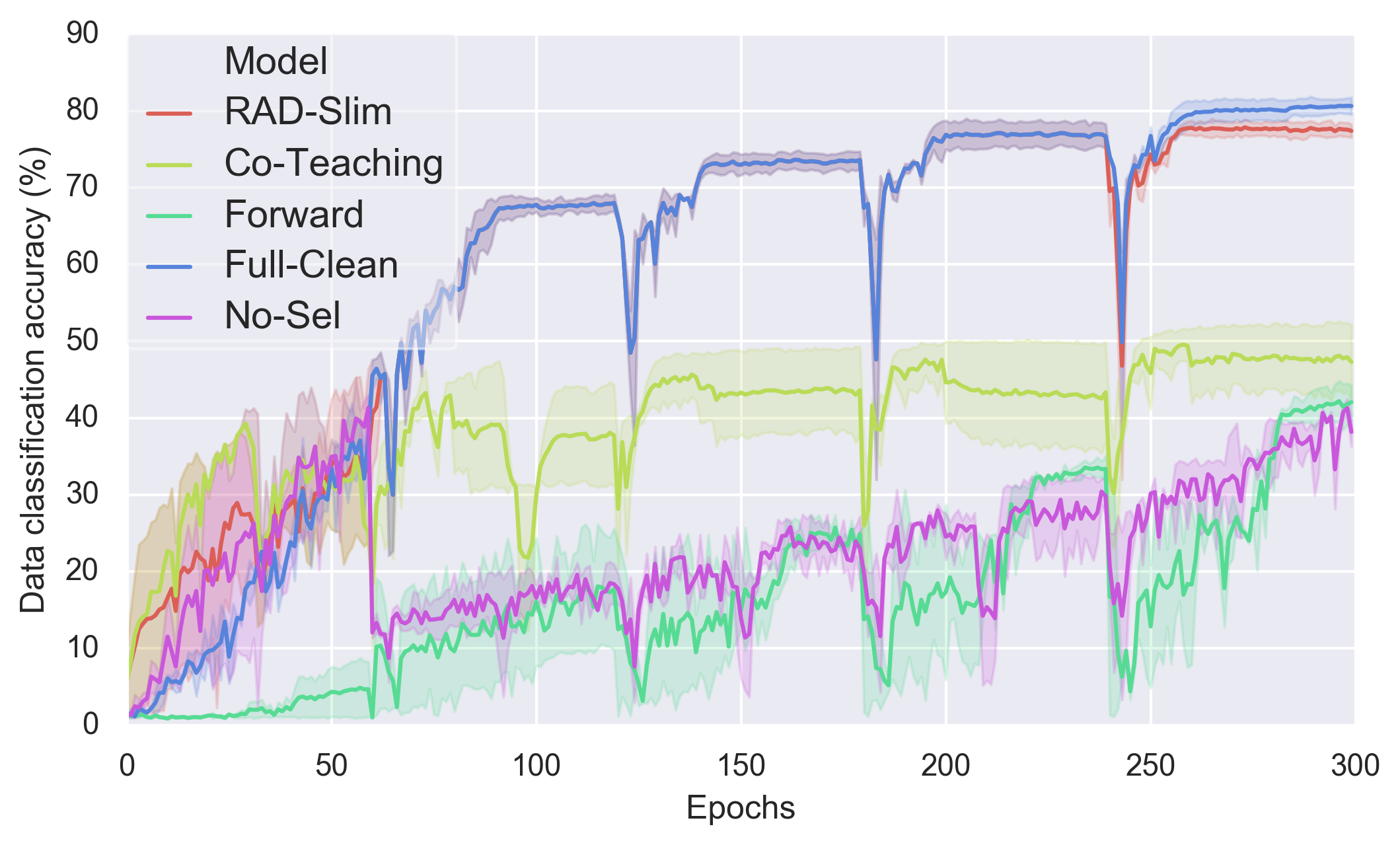}
    \vspace{-0.2em}
    \caption{FaceScrub with noise level of 30\%} 
    \label{fig:curve-oracle-image}
    \vspace{-1.0em}
\end{figure}

\begin{figure}[t]
    \centering
    \includegraphics[width=0.83\linewidth]{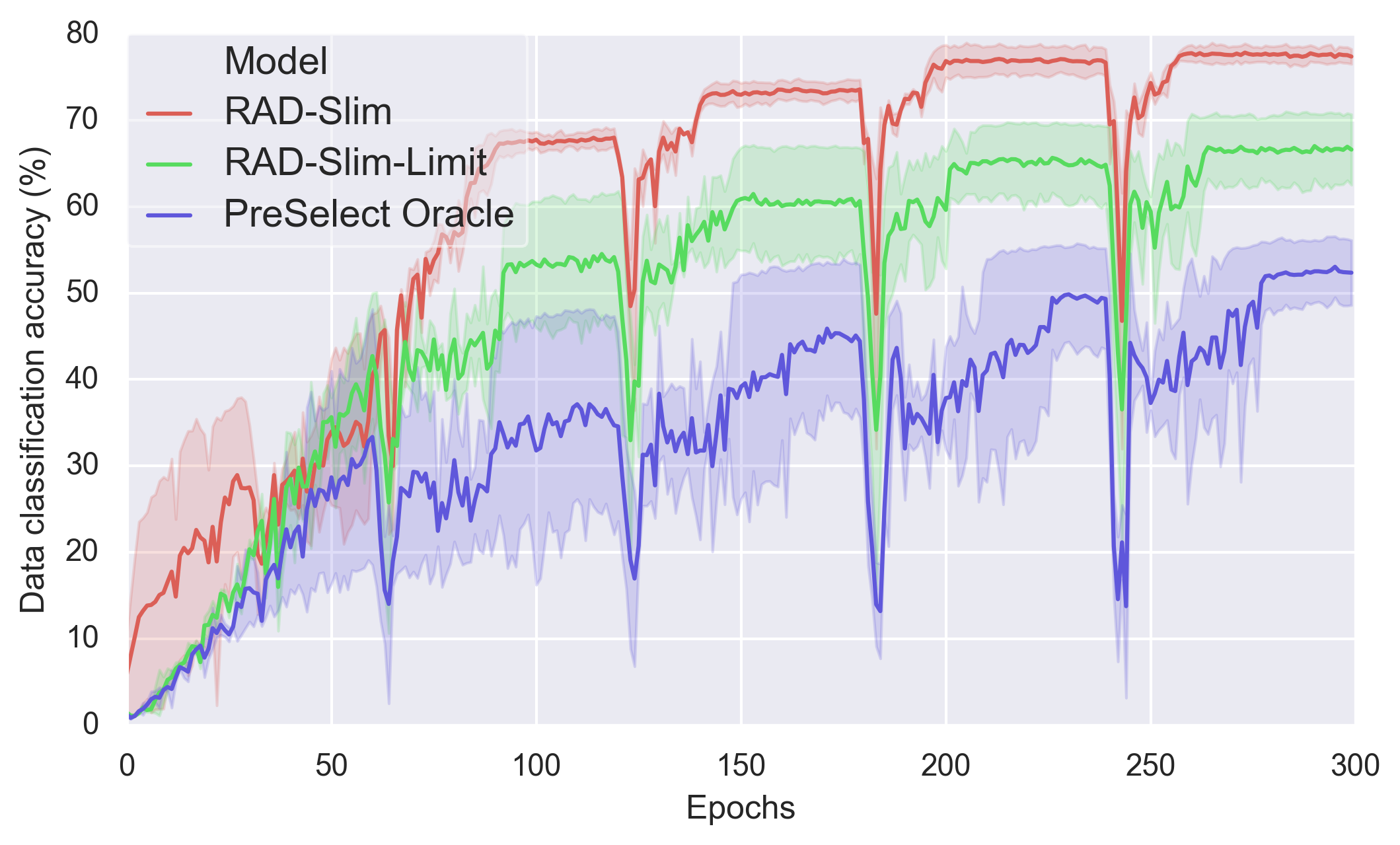}
    \vspace{-0.2em}
    \caption{\systemSlimmedLimit on FaceScrub with 30\% noise} 
    \label{fig:curve-oracle-limit}
    \vspace{-1.0em}
\end{figure}

We evaluate the \system framework on the challenging case of  noisy image classification. Specifically,  we apply \systemSlimmed and \systemSlimmedLimit  (20\% of batch size query limit per batch) to train a classifier that encounters on-line noisy images.  
Fig.~\ref{fig:curve-oracle-image} shows the accuracy results across the batch arrivals. We can observe that \systemSlimmed is close to the {\em Full-Clean} baseline and largely outperforms other baselines. Detailed numbers are summarized in Table~\ref{tab:image_result}. Fig.~\ref{fig:curve-oracle-limit} shows the comparison between \systemSlimmed, \systemSlimmedLimit and \systemPreselect (same design as in Sec.~\ref{subsec:rad_active_learning}). One can see that \systemSlimmedLimit performs significantly better than \systemPreselect when both use the same limit on the number of expert queries.

\begin{figure}[t]
    \centering
    \includegraphics[width=0.83\linewidth]{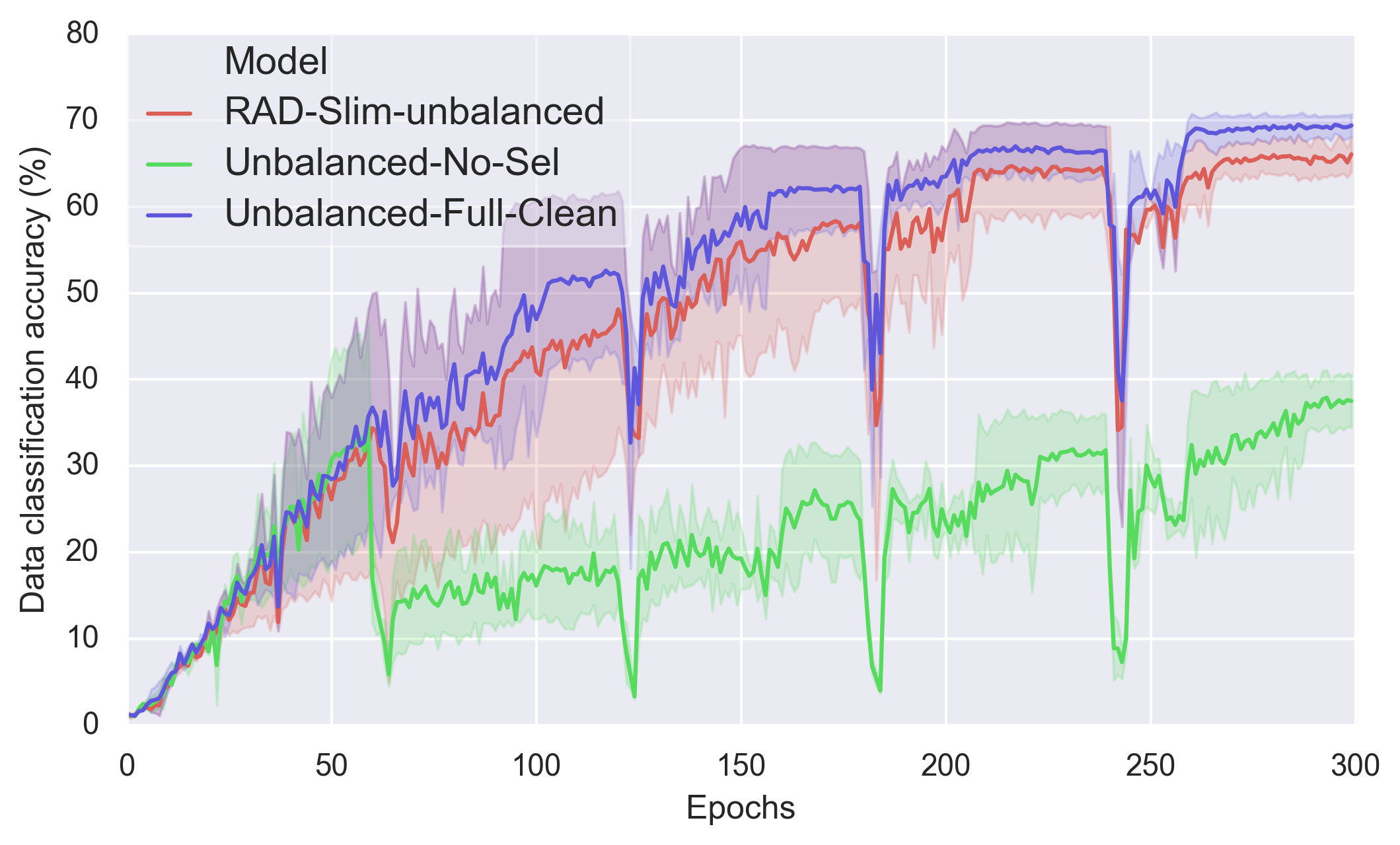}
    \vspace{-0.2em}
    \caption{Unbalanced FaceScrub with 30\% noise} 
    \label{fig:curve-facescrub-unbalanced}
\end{figure}

To further display the effectiveness of \systemSlimmed on different types of attack, we design a series of unbalanced noisy data batches. The original Facescrub dataset comprises a ratio of 55\%:45\% male and female images. For $\mathcal{D}_{0}$ of the unbalanced data batches, image ratio of male and female is 90\%:10\% followed by  45\%:55\% in subsequent batches. Fig.~\ref{fig:curve-facescrub-unbalanced} shows the results, \systemSlimmed performs definitely better than no selection, and very close to full clean scenario, which shows that \systemActive can not only defend the model from different noise levels, but also resist other types of attack.

Another observation is that all curves suffer a periodic up-down pattern. This is because for image dataset, each time a new batch comes, we only use this new batch data as training dataset. As different batches provide different subviews of the data the empirical distribution can be different from the calculated optimum, but the model remains. So for the first epoch of a new batch, we will generate a gradient which is based on new data but applied on an old model. This can influence the accuracy of the model. Moreover when retraining on each new data batch we reset the learning rate which causes a bump in the learning rate. Therefore, even if all batches follow the same distribution, the system could temporarily wander off from the previous optimum.


\begin{table}[t]
	\begin{center}
		\caption{Final accuracy of different algorithms on FaceScrub dataset with 30\% noise, results are averaged on 3 runs.}
		\label{tab:image_result}
		\begin{tabular}{L{3.5cm} C{2cm}}
			\toprule
			 Algorithm & Accuracy\\
			\midrule
			  Full-Clean &  81.72  \\
		      No-Sel &  38.89  \\
		      Forward &41.71\\
		      \systemCoteaching & 47.39\\
		     \midrule
			  \systemSlimmed & 77.51\\
			  \systemSlimmedLimit & 67.18\\
			  \systemPreselect & 52.12\\
			 \midrule
			  \systemSlimmed-Unbalanced & 66.12\\
			  Unbalanced-No-Sel & 37.11\\
			  Unbalanced-Full-Clean & 69.95\\
			  
		\end{tabular}
	\end{center}
	\vspace{-2em}
\end{table}

\section{Concluding Remarks}
\label{sec:Conclusion}

While machine learning classification algorithms are widely applied to detect anomalies, the commonly employed assumption of clean anomaly labels often does not hold for data collected in the wild due to careless annotation and malicious dirty label pollution. The noisy labels can significantly degrade the accuracy of anomaly detection 
and are challenging to tackle due to the lack of ground truth of label quality. In this paper, we present a on-line framework for robust anomaly detection, \system, which can continuously learn the system dynamics and anomaly behaviours from streams of arriving data after filtering out suspicious noisy data. 

\system is a general framework that composes of label quality predictor and classification model, where the former mainly captures the label dynamics and the latter focuses on increasing the diversity of prediction. Predictions from both contribute to the final decision on detecting anomaly. To adapt to the on-line nature of anomaly detection, we extend \system with additional features of conflicting opinions of classifiers, repetitively cleaning, and oracle knowledge, corresponding to \systemVoting, \systemActive, and \systemActiveLimit. We demonstrate the effectiveness of \system and its extensions on three uses cases, i.e., detecting IoT device attacks, predicting task failures at Google clusters and recognising celebrity faces from FaceScrub. The evaluation results on three use cases show remarkable accuracy that are close to the case without encountering anomaly input. In short, RAD is a general robust learning framework that can be applied on different classification models and enhances their robustness against noisy inputs during on-line training. 


\section*{Acknowledgement}
This work has been partly funded by the French LabEx PERSYVAL-Lab (ANR-11-LABX-0025-01), the Swiss National Science Foundation NRP75 project 407540\_167266 and the National Natural Science Foundation of China (Grant No. 61872337).  

\bibliographystyle{abbrvname}
\bibliography{biblio}

\begin{thebibliography}{10}

\bibitem{Agarwal:2016}
M.~Agarwal, D.~Pasumarthi, S.~Biswas, and S.~Nandi.
\newblock Machine learning approach for detection of flooding dos attacks in
  802.11 networks and attacker localization.
\newblock {\em Int. J. Machine Learning {\&} Cybernetics}, 7(6):1035--1051,
  2016.

\bibitem{An:Neurocomput.13:SVM}
W.~An and M.~Liang.
\newblock Fuzzy support vector machine based on within-class scatter for
  classification problems with outliers or noises.
\newblock {\em Neurocomput.}, 110:101--110, June 2013.

\bibitem{Anbar:2018}
M.~Anbar, R.~Abdullah, B.~N. Al{-}Tamimi, and A.~Hussain.
\newblock A machine learning approach to detect router advertisement flooding
  attacks in next-generation ipv6 networks.
\newblock {\em Cognitive Computation}, 10(2):201--214, 2018.

\bibitem{Banescu:2017}
S.~Banescu, C.~S. Collberg, and A.~Pretschner.
\newblock Predicting the resilience of obfuscated code against symbolic
  execution attacks via machine learning.
\newblock In {\em 26th {USENIX} Security Symposium, Vancouver, BC, Canada.},
  pages 661--678. {USENIX} Association, 2017.

\bibitem{biggio2011support}
B.~Biggio, B.~Nelson, and P.~Laskov.
\newblock Support vector machines under adversarial label noise.
\newblock In {\em Asian Conference on Machine Learning}, pages 97--112, 2011.

\bibitem{BirkeDSN2014}
R.~Birke, G.~Ioana, \myname{Lydia Y. Chen}, D.~Wiesmann, and T.~Engbersen.
\newblock Failure analysis of virtual and physical machines: patterns, causes
  and characteristics.
\newblock In {\em {IEEE/IFIP DSN}}, pages 1--12, 2014.

\bibitem{Birke:TNSM16:Cloud}
R.~Birke, A.~Podzimek, \myname{Lydia Y. Chen}, and E.~Smirni.
\newblock Virtualization in the private cloud: State of the practice.
\newblock {\em {IEEE} Trans. Network and Service Management}, 13(3):608--621,
  2016.

\bibitem{brodley1996}
C.~E. {Brodley} and M.~A. {Friedl}.
\newblock Improving automated land cover mapping by identifying and eliminating
  mislabeled observations from training data.
\newblock In {\em IGARSS '96. 1996 International Geoscience and Remote Sensing
  Symposium}, volume~2, pages 1379--1381 vol.2, 1996.

\bibitem{Campos:2018}
J.~R. Campos, M.~Vieira, and E.~Costa.
\newblock Exploratory study of machine learning techniques for supporting
  failure prediction.
\newblock In {\em 14th European Dependable Computing Conference, Ia{\c{s}}i,
  Romania}, pages 9--16. {IEEE} Computer Society, 2018.

\bibitem{Cerf:NIPsWS18:Duao}
S.~Cerf, R.~Birke, and \myname{Lydia Y. Chen}.
\newblock Duo learning for classifications with noisy labels.
\newblock In {\em Continual Learning Workshop, in conjunction with Neural
  Information Processing Systems (NIPS)}, 2018.

\bibitem{Fan:2018}
Y.~Fan, J.~Li, and D.~Zhang.
\newblock A method for identifying critical elements of a cyber-physical system
  under data attack.
\newblock {\em {IEEE} Access}, 6:16972--16984, 2018.

\bibitem{Fang:2010}
Z.~Fang, J.~Tzeng, C.~C. Chen, and T.~Chou.
\newblock A study of machine learning models in epidemic surveillance: Using
  the query logs of search engines.
\newblock In {\em Pacific Asia Conference on Information Systems, {PACIS} 2010,
  Taipei, Taiwan, 9-12 July 2010}, page 137. AISeL, 2010.

\bibitem{Frenay:TNNLS14:survey}
B.~Fr{\'{e}}nay and M.~Verleysen.
\newblock Classification in the presence of label noise: {A} survey.
\newblock {\em {IEEE} Trans. Neural Netw. Learning Syst.}, 25(5):845--869,
  2014.

\bibitem{tps2019}
A.~{Ghiassi}, T.~{Younesian}, Z.~{Zhao}, R.~{Birke}, V.~{Schiavoni}, and L.~Y.
  {Chen}.
\newblock Robust (deep) learning framework against dirty labels and beyond.
\newblock In {\em IEEE International Conference on Trust, Privacy and Security
  in Intelligent Systems and Applications (TPS-ISA)}, pages 236--244, 2019.

\bibitem{Giantamidis:2016}
G.~Giantamidis and S.~Tripakis.
\newblock Learning moore machines from input-output traces.
\newblock In {\em {FM} 2016: Formal Methods - 21st International Symposium,
  Limassol, Cyprus, Proceedings}, volume 9995 of {\em Lecture Notes in Computer
  Science}, pages 291--309, 2016.

\bibitem{guyon1994}
I.~Guyon, N.~Mati\'{c}, and V.~Vapnik.
\newblock Discovering informative patterns and data cleaning.
\newblock In {\em Proceedings of International Conference on Knowledge
  Discovery and Data Mining}, AAAIWS'94, page 145–156, 1994.

\bibitem{coteaching}
B.~Han, Q.~Yao, X.~Yu, G.~Niu, M.~Xu, W.~Hu, I.~W. Tsang, and M.~Sugiyama.
\newblock Co-teaching: Robust training of deep neural networks with extremely
  noisy labels.
\newblock In {\em Advances in Neural Information Processing Systems 31: Annual
  Conference on Neural Information Processing Systems, Montr{\'{e}}al, Canada},
  pages 8536--8546, 2018.

\bibitem{He2015DeepRL}
K.~He, X.~Zhang, S.~Ren, and J.~Sun.
\newblock Deep residual learning for image recognition.
\newblock {\em 2016 IEEE Conference on Computer Vision and Pattern Recognition
  (CVPR)}, pages 770--778, 2015.

\bibitem{He:2017}
Y.~He, G.~J. Mendis, and J.~Wei.
\newblock Real-time detection of false data injection attacks in smart grid:
  {A} deep learning-based intelligent mechanism.
\newblock {\em {IEEE} Trans. Smart Grid}, 8(5):2505--2516, 2017.

\bibitem{Dan:nips2018:Using}
D.~Hendrycks, M.~Mazeika, D.~Wilson, and K.~Gimpel.
\newblock Using trusted data to train deep networks on labels corrupted by
  severe noise.
\newblock In {\em Conf. on Neural Information Processing Systems, Montr\'eal,
  Canada.}, 2018.

\bibitem{Huang:2017}
T.~H. {Huang}, C.~{Yu}, and H.~{Kao}.
\newblock Data-driven and deep learning methodology for deceptive advertising
  and phone scams detection.
\newblock In {\em 2017 Conference on Technologies and Applications of
  Artificial Intelligence (TAAI)}, pages 166--171, Dec 2017.

\bibitem{piyasak2010}
P.~Jeatrakul, K.~Wong, and C.~Fung.
\newblock Data cleaning for classification using misclassification analysis.
\newblock {\em JACIII}, 14:297--302, 04 2010.

\bibitem{Kang:2018}
J.~Kang, I.~Joo, and D.~Choi.
\newblock False data injection attacks on contingency analysis: Attack
  strategies and impact assessment.
\newblock {\em {IEEE} Access}, 6:8841--8851, 2018.

\bibitem{Karagiannis:2018}
D.~Karagiannis and A.~Argyriou.
\newblock Jamming attack detection in a pair of {RF} communicating vehicles
  using unsupervised machine learning.
\newblock {\em Vehicular Communications}, 13:56--63, 2018.

\bibitem{Kozik:2018}
R.~Kozik, M.~Choras, M.~Ficco, and F.~Palmieri.
\newblock A scalable distributed machine learning approach for attack detection
  in edge computing environments.
\newblock {\em J. Parallel Distrib. Comput.}, 119:18--26, 2018.

\bibitem{larsen1998design}
J.~Larsen, L.~Nonboe, M.~Hintz-Madsen, and L.~K. Hansen.
\newblock Design of robust neural network classifiers.
\newblock In {\em Acoustics, Speech and Signal Processing, 1998. Proceedings of
  the 1998 IEEE International Conference on}, volume~2, pages 1205--1208. IEEE,
  1998.

\bibitem{li2017learning}
Y.~Li, J.~Yang, Y.~Song, L.~Cao, J.~Luo, and L.-J. Li.
\newblock Learning from noisy labels with distillation.
\newblock In {\em ICCV}, pages 1928--1936, 2017.

\bibitem{meidan2018n}
Y.~Meidan, M.~Bohadana, Y.~Mathov, Y.~Mirsky, A.~Shabtai, D.~Breitenbacher, and
  Y.~Elovici.
\newblock N-baiot—network-based detection of iot botnet attacks using deep
  autoencoders.
\newblock {\em IEEE Pervasive Computing}, 17(3):12--22, 2018.

\bibitem{DBLP:conf/hais/MirandaGCL09}
A.~L.~B. Miranda, L.~P.~F. Garcia, A.~C. P. L.~F. de~Carvalho, and A.~C.
  Lorena.
\newblock Use of classification algorithms in noise detection and elimination.
\newblock In {\em Hybrid Artificial Intelligence Systems, 4th International
  Conference, {HAIS} 2009, Salamanca, Spain, June 10-12, 2009. Proceedings},
  pages 417--424, 2009.

\bibitem{natarajan2013learning}
N.~Natarajan, I.~S. Dhillon, P.~K. Ravikumar, and A.~Tewari.
\newblock Learning with noisy labels.
\newblock In {\em Advances in neural information processing systems}, pages
  1196--1204, 2013.

\bibitem{facescrub:2014}
H.-W. Ng and S.~Winkler.
\newblock A data-driven approach to cleaning large face datasets.
\newblock {\em 2014 IEEE International Conference on Image Processing, ICIP
  2014}, pages 343--347, 01 2015.

\bibitem{patrini2017making}
G.~Patrini, A.~Rozza, A.~K. Menon, R.~Nock, and L.~Qu.
\newblock Making deep neural networks robust to label noise: A loss correction
  approach.
\newblock In {\em IEEE CVPR}, pages 2233--2241, 2017.

\bibitem{scikit-learn}
F.~Pedregosa, G.~Varoquaux, A.~Gramfort, V.~Michel, B.~Thirion, O.~Grisel,
  M.~Blondel, P.~Prettenhofer, R.~Weiss, V.~Dubourg, J.~Vanderplas, A.~Passos,
  D.~Cournapeau, M.~Brucher, M.~Perrot, and E.~Duchesnay.
\newblock Scikit-learn: Machine learning in {P}ython.
\newblock {\em Journal of Machine Learning Research}, 12:2825--2830, 2011.

\bibitem{Pellegrini:2015}
A.~Pellegrini, P.~di~Sanzo, and D.~R. Avresky.
\newblock A machine learning-based framework for building application failure
  prediction models.
\newblock In {\em {IEEE} International Parallel and Distributed Processing
  Symposium Workshop, Hyderabad, India}, pages 1072--1081. {IEEE} Computer
  Society, 2015.

\bibitem{DBLP:journals/tpds/PhamWTBTKI17}
C.~Pham, L.~Wang, B.~Tak, S.~Baset, C.~Tang, Z.~T. Kalbarczyk, and R.~K. Iyer.
\newblock Failure diagnosis for distributed systems using targeted fault
  injection.
\newblock {\em {IEEE} Trans. Parallel Distrib. Syst.}, 28(2):503--516, 2017.

\bibitem{Pitakrat:2013}
T.~Pitakrat, A.~van Hoorn, and L.~Grunske.
\newblock A comparison of machine learning algorithms for proactive hard disk
  drive failure detection.
\newblock In {\em Proceedings of the 4th international {ACM} Sigsoft symposium
  on Architecting critical systems, Vancouver, BC, Canada}, pages 1--10, 2013.

\bibitem{reiss2011google}
C.~Reiss, J.~Wilkes, and J.~L. Hellerstein.
\newblock Google cluster-usage traces: format+ schema.
\newblock {\em Google Inc., White Paper}, pages 1--14, 2011.

\bibitem{Reuter:2018}
U.~Reuter, A.~Sultan, and D.~S. Reischl.
\newblock A comparative study of machine learning approaches for modeling
  concrete failure surfaces.
\newblock {\em Advances in Engineering Software}, 116:67--79, 2018.

\bibitem{Rosa:TSC17:failurePrediction}
A.~Ros{\`{a}}, L.~Y. Chen, and W.~Binder.
\newblock Failure analysis and prediction for big-data systems.
\newblock {\em {IEEE} Trans. Services Computing}, 10(6):984--998, 2017.

\bibitem{Rosa-DSN15}
A.~Ros{\`{a}}, \myname{Lydia Y. Chen}, and W.~Binder.
\newblock Understanding the dark side of big data clusters: An analysis beyond
  failures.
\newblock In {\em {IEEE/IFIP DSN}}, pages 207--218, 2015.

\bibitem{DBLP:conf/cvpr/SchroffKP15}
F.~Schroff, D.~Kalenichenko, and J.~Philbin.
\newblock Facenet: {A} unified embedding for face recognition and clustering.
\newblock In {\em {IEEE} {CVPR} 2015 , Boston, MA, USA}, pages 815--823. {IEEE}
  Computer Society, 2015.

\bibitem{Simonyan15}
K.~Simonyan and A.~Zisserman.
\newblock Very deep convolutional networks for large-scale image recognition.
\newblock In {\em International Conference on Learning Representations}, San
  Diego, CA, USA, 2015.

\bibitem{sukhbaatar2014training}
S.~Sukhbaatar, J.~Bruna, M.~Paluri, L.~Bourdev, and R.~Fergus.
\newblock Training convolutional networks with noisy labels.
\newblock {\em arXiv preprint arXiv:1406.2080}, 2014.

\bibitem{DBLP:conf/cvpr/TaigmanYRW14}
Y.~Taigman, M.~Yang, M.~Ranzato, and L.~Wolf.
\newblock Deepface: Closing the gap to human-level performance in face
  verification.
\newblock In {\em {IEEE} Conference on Computer Vision and Pattern Recognition,
  Columbus, OH, USA}, pages 1701--1708. {IEEE} Computer Society, 2014.

\bibitem{DBLP:conf/apweb/ThongkamXZH08}
J.~Thongkam, G.~Xu, Y.~Zhang, and F.~Huang.
\newblock Support vector machine for outlier detection in breast cancer
  survivability prediction.
\newblock In {\em Advanced Web and Network Technologies, and Applications,
  International Workshops: BIDM, IWHDM, and DeWeb Shenyang, China}, pages
  99--109, 2008.

\bibitem{Vagin:2011}
V.~N. Vagin and M.~V. Fomina.
\newblock Problem of knowledge discovery in noisy databases.
\newblock {\em Int. J. Machine Learning {\&} Cybernetics}, 2(3):135--145, 2011.

\bibitem{Vahdat:NIPS17:NoisyLabelDNN}
A.~Vahdat.
\newblock Toward robustness against label noise in training deep discriminative
  neural networks.
\newblock In {\em NIPS}, pages 5601--5610, 2017.

\bibitem{veit2017learning}
A.~Veit, N.~Alldrin, G.~Chechik, I.~Krasin, A.~Gupta, and S.~J. Belongie.
\newblock Learning from noisy large-scale datasets with minimal supervision.
\newblock In {\em CVPR}, pages 6575--6583, 2017.

\bibitem{DBLP:conf/cvpr/WangWZJGZL018}
H.~Wang, Y.~Wang, Z.~Zhou, X.~Ji, D.~Gong, J.~Zhou, Z.~Li, and W.~Liu.
\newblock Cosface: Large margin cosine loss for deep face recognition.
\newblock In {\em {IEEE} CVPR, Salt Lake City, UT, USA}, pages 5265--5274,
  2018.

\bibitem{Wilson:ML2000:MLselection}
D.~R. Wilson and T.~R. Martinez.
\newblock Reduction techniques for instance-basedlearning algorithms.
\newblock {\em Mach. Learn.}, 38(3):257--286, Mar. 2000.

\bibitem{Xue:TNSM18:Ticket}
J.~Xue, R.~Birke, \myname{Lydia Y. Chen}, and E.~Smirni.
\newblock Spatial-temporal prediction models for active ticket managing in data
  centers.
\newblock {\em {IEEE} Trans. Network and Service Management}, 15(1):39--52,
  2018.

\bibitem{DBLP:conf/iclr/ZhangBHRV17}
C.~Zhang, S.~Bengio, M.~Hardt, B.~Recht, and O.~Vinyals.
\newblock Understanding deep learning requires rethinking generalization.
\newblock In {\em 5th International Conference on Learning Representations,
  Toulon, France.}, 2017.

\bibitem{DBLP:conf/ijcai/ZhangLSLH19}
M.~Zhang, T.~Li, H.~Shi, Y.~Li, and P.~Hui.
\newblock A decomposition approach for urban anomaly detection across
  spatiotemporal data.
\newblock In {\em Proceedings of the Twenty-Eighth International Joint
  Conference on Artificial Intelligence, Macao, China}, pages 6043--6049, 2019.

\bibitem{Zhao:DSN19}
Z.~Zhao, S.~Cerf, R.~Birke, B.~Robu, S.~Bouchenak, S.~B. Mokhtar, and L.~Y.
  Chen.
\newblock Robust anomaly detection on unreliable data.
\newblock In {\em 49th Annual {IEEE/IFIP} International Conference on
  Dependable Systems and Networks, {DSN} 2019, Portland, OR, USA}, pages
  630--637, 2019.

\bibitem{epd}
Z.~Zhao, S.~Cerf, B.~Robu, and N.~Marchand.
\newblock Feedback control for online training of neural networks.
\newblock In {\em {IEEE} Conference on Control Technology and Applications,
  Hong Kong, China}, pages 136--141, 2019.

\bibitem{eblrc}
Z.~{Zhao}, S.~{Cerf}, B.~{Robu}, and N.~{Marchand}.
\newblock Event-based control for online training of neural networks.
\newblock {\em IEEE Control Systems Letters}, pages 1--6, 2020.

\bibitem{Zhou:2018}
B.~Zhou, J.~Li, J.~Wu, S.~Guo, Y.~Gu, and Z.~Li.
\newblock Machine-learning-based online distributed denial-of-service attack
  detection using spark streaming.
\newblock In {\em {IEEE} International Conference on Communications, Kansas
  City, USA}, pages 1--6, 2018.

\end{thebibliography}

\begin{IEEEbiography}[{\includegraphics[width=1in, clip,keepaspectratio]{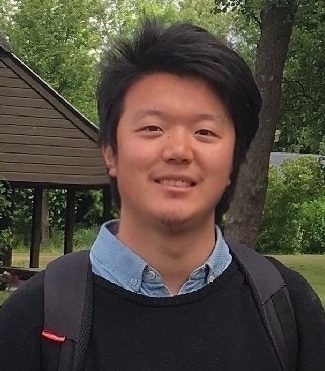}}]
{Zilong Zhao} is a Postdoctoral researcher in the Delft University of Technology in The Netherlands since 2020. He received his PhD in Automation and Information from Université Grenoble Alpes (UGA). He was the lead AI and data engineer at the Miro Health. His research focuses on applying control theory into machine learning algorithms, building robust and dependable machine learning systems, AI application optimization and generative adversarial networks (GANs).  
\end{IEEEbiography}

\begin{IEEEbiography}[{\includegraphics[width=1in, clip,keepaspectratio]{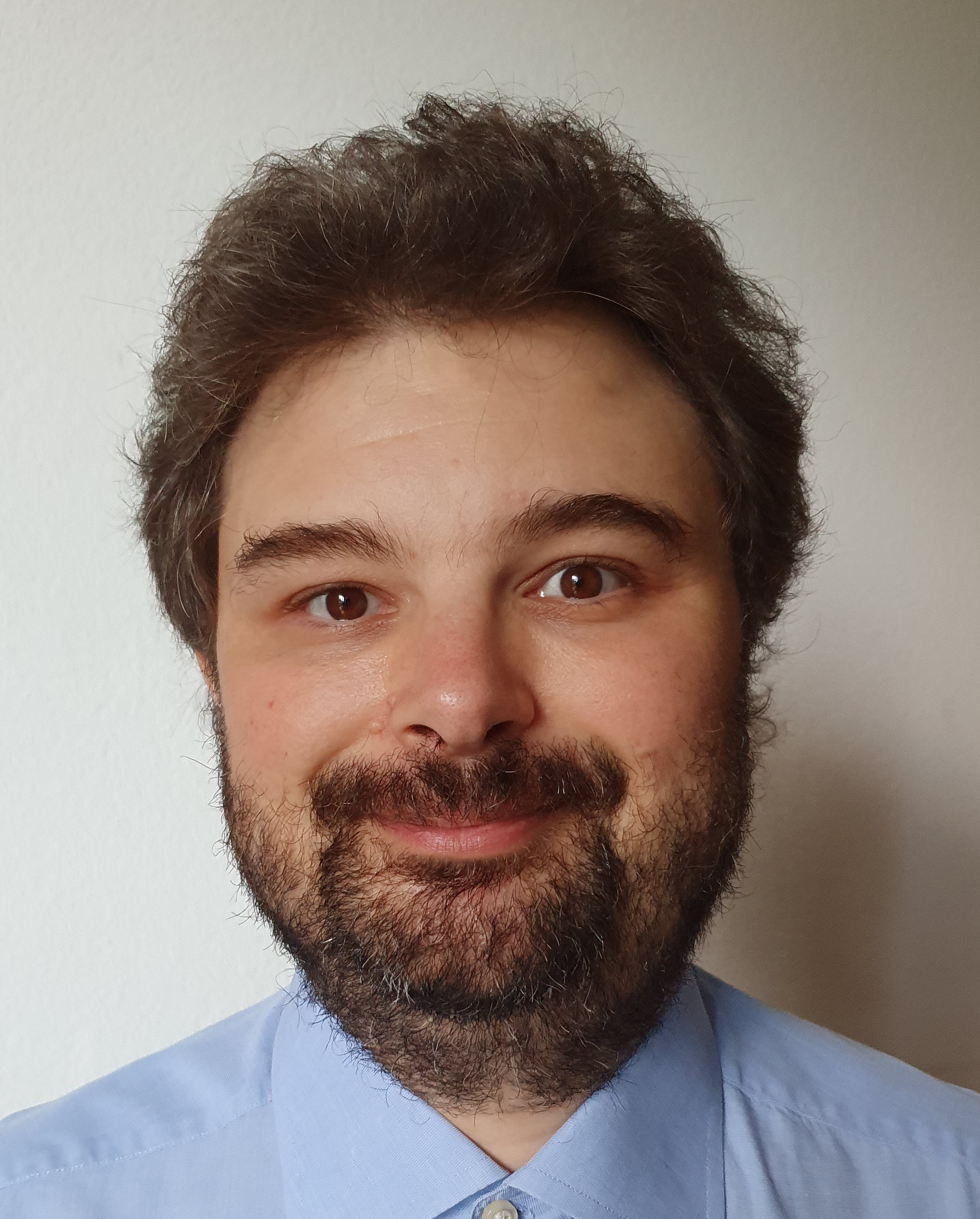}}]
{Robert Birke} is a Principal Scientist at the ABB Research Lab, Switzerland. He received his Ph.D. in Electronics and Communications Engineering from the Politecnico di Torino, Italy. His research interests are in the broad area of virtual resource management including network design, workload characterization, and AI and big-data application optimization. He has published more than 80 papers at venues related to communication and system performance, e.g., SIGCOMM, SIGMETRICS, FAST, INFOCOM, and JSAC.
He is a senior member of IEEE.
\end{IEEEbiography}

\begin{IEEEbiography}[{\includegraphics[width=1in,height=1.25in,clip,keepaspectratio]{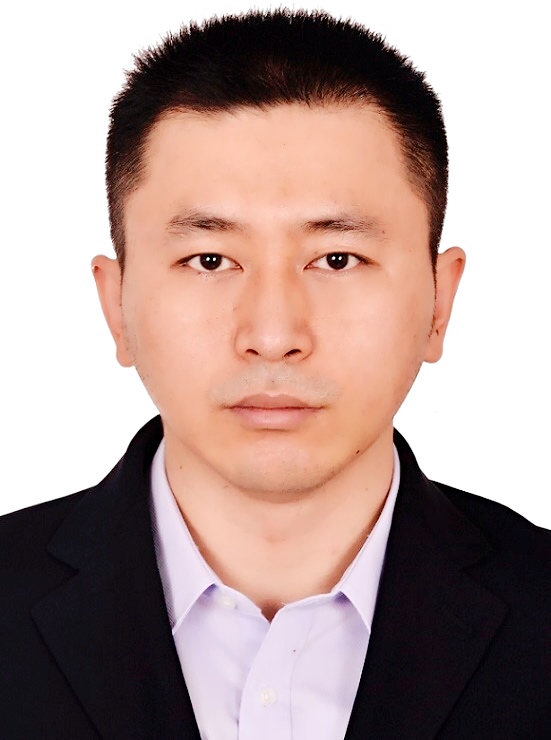}}]{Rui Han}
is an Associate Professor at the School of Computer Science and Technology, Beijing Institute of Technology, China. Before joining BIT, He received MSc with honor in 2010 from Tsinghua University, China, and obtained his PhD degree in 2014 from the Department of Computing, Imperial College London, UK. His research interests are system optimization for deep learning workloads. He has over 40 publications in these areas, including papers at TPDS, TKDE, TC, INFOCOM, and ICDCS.
\end{IEEEbiography}

\begin{IEEEbiography}[{\includegraphics[width=1in, clip,keepaspectratio]{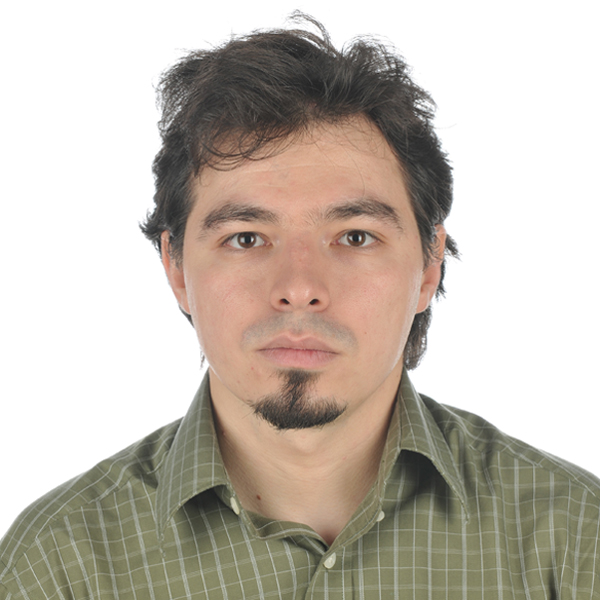}}]
{Bogdan Robu} is associate professor at the Université Grenoble
Alpes (UGA) and a researcher in GIPSA-lab
laboratory, Grenoble, France since September
2011. He received his PhD in 2010 from the University
of Toulouse. His research focus is on applying
control theory techniques to machine learning algorithms, Cloud/Fog software
and parallel computing systems in order to achieve dependable, trustworthy and highly available systems. He has co-authored around 40 publications in peer-reviewed conferences and journals.
\end{IEEEbiography}

\begin{IEEEbiography}[{\includegraphics[width=1in, clip,keepaspectratio]{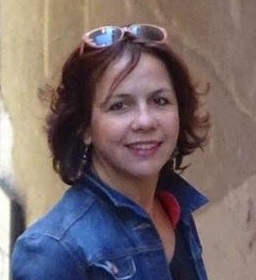}}]
{Sara Bouchenak} is Professor at INSA Lyon – LIRIS laboratory. She conducts research on highly available, dependable, privacy preserving and efficient distributed computer systems. She serves as Women in Computer Science Committee Chair for INSA Lyon – Department of Computer Science. She has co-authored more than 70 publications. She has been Chair and member of the PC of several conferences (ATC, DSN, ICDCS, SRDS, The WebConf, etc.). She has coordinated and participated in several national and international projects. She serves as scientific expert for the European Commission, ANR 5france), FNS (Switzerland), Vinnova (Sweden).
\end{IEEEbiography}

\begin{IEEEbiography}[{\includegraphics[width=1in, clip,keepaspectratio]{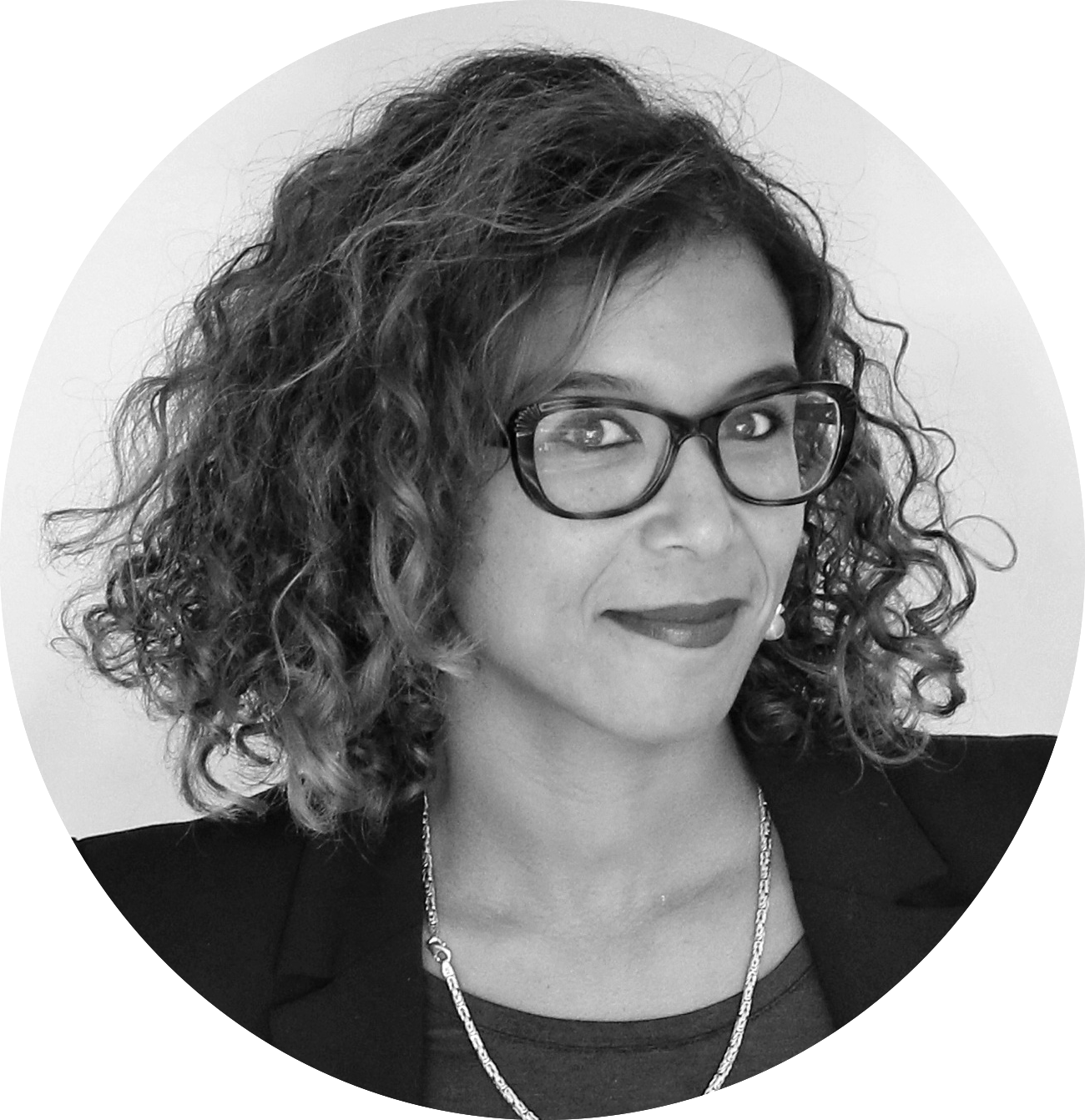}}]
{Sonia Ben Mokhtar} is a CNRS research director at the LIRIS lab and the head of the distributed systems and information retrieval group (DRIM). She received her PhD in 2007 from Université Pierre et Marie Curie before spending two years at University College London (UK). Her research focuses on the design of resilient and privacy-preserving distributed systems. Sonia has co-authored 70+ papers in peer-reviewed conferences and journals and has served on the editorial board of IEEE Transactions on Dependable and Secure Computing. 
\end{IEEEbiography}

\begin{IEEEbiography}[{\includegraphics[width=1in, clip,keepaspectratio]{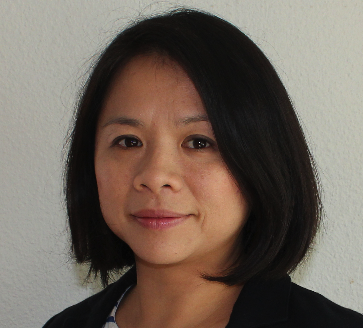}}]
{Lydia Y. Chen} is an Associate Professor in the Department of Computer Science at the Delft University of Technology in The Netherlands. Prior to joining TU Delft, she was a research staff member at the IBM Research Zurich Lab from 2007 to 2018. She holds a PhD from Pennsylvania State University and a BA from National Taiwan University. Her research interests are distributed machine learning,  dependability management, resource allocation for large-scale data processing systems and services. More specifically, her work focuses on developing stochastic and machine learning models, and applying these techniques to application domains, such as data centers and AI systems.  She has published more than 100 papers in peer-reviewed journals. She has served on the editorial boards of IEEE Transactions on Dependable and Secure Computing, IEEE Transactions on Parallel and Distributed Systems, IEEE Transactions on Service Computing and IEEE Transactions on Network and Service Management. She is a Senior Member of IEEE.
\end{IEEEbiography}
\end{document}